\documentclass[sigconf]{acmart}
\AtBeginDocument{%
  }

\usepackage{amsmath}
\usepackage{amsfonts}
\usepackage{bm}
\usepackage{graphicx}
\usepackage{caption}
\usepackage{multirow}
\usepackage{adjustbox}
\usepackage{siunitx}
\usepackage{pgfplots}
\usepackage{subcaption}

\pgfplotsset{compat=1.18}
\usepgfplotslibrary{groupplots, colorbrewer, colormaps, statistics, fillbetween}
\usetikzlibrary{positioning, arrows.meta, calc, fit, backgrounds, patterns, decorations.pathmorphing}

\definecolor{darkred}{RGB}{160,0,0}

\newlength{\imggap}
\newlength{\vgap}

\definecolor{red50}{HTML}{FFEBEE}
\definecolor{red100}{HTML}{FFCDD2}
\definecolor{red200}{HTML}{EF9A9A}
\definecolor{red300}{HTML}{E57373}
\definecolor{red400}{HTML}{EF5350}
\definecolor{red500}{HTML}{F44336}
\definecolor{red600}{HTML}{E53935}
\definecolor{red700}{HTML}{D32F2F}
\definecolor{red800}{HTML}{C62828}
\definecolor{red900}{HTML}{B71C1C}

\definecolor{pink50}{HTML}{FCE4EC}
\definecolor{pink100}{HTML}{F8BBD0}
\definecolor{pink200}{HTML}{F48FB1}
\definecolor{pink300}{HTML}{F06292}
\definecolor{pink400}{HTML}{EC407A}
\definecolor{pink500}{HTML}{E91E63}
\definecolor{pink600}{HTML}{D81B60}
\definecolor{pink700}{HTML}{C2185B}
\definecolor{pink800}{HTML}{AD1457}
\definecolor{pink900}{HTML}{880E4F}

\definecolor{purple50}{HTML}{F3E5F5}
\definecolor{purple100}{HTML}{E1BEE7}
\definecolor{purple200}{HTML}{CE93D8}
\definecolor{purple300}{HTML}{BA68C8}
\definecolor{purple400}{HTML}{AB47BC}
\definecolor{purple500}{HTML}{9C27B0}
\definecolor{purple600}{HTML}{8E24AA}
\definecolor{purple700}{HTML}{7B1FA2}
\definecolor{purple800}{HTML}{6A1B9A}
\definecolor{purple900}{HTML}{4A148C}

\definecolor{deeppurple50}{HTML}{EDE7F6}
\definecolor{deeppurple100}{HTML}{D1C4E9}
\definecolor{deeppurple200}{HTML}{B39DDB}
\definecolor{deeppurple300}{HTML}{9575CD}
\definecolor{deeppurple400}{HTML}{7E57C2}
\definecolor{deeppurple500}{HTML}{673AB7}
\definecolor{deeppurple600}{HTML}{5E35B1}
\definecolor{deeppurple700}{HTML}{512DA8}
\definecolor{deeppurple800}{HTML}{4527A0}
\definecolor{deeppurple900}{HTML}{311B92}

\definecolor{indigo50}{HTML}{E8EAF6}
\definecolor{indigo100}{HTML}{C5CAE9}
\definecolor{indigo200}{HTML}{9FA8DA}
\definecolor{indigo300}{HTML}{7986CB}
\definecolor{indigo400}{HTML}{5C6BC0}
\definecolor{indigo500}{HTML}{3F51B5}
\definecolor{indigo600}{HTML}{3949AB}
\definecolor{indigo700}{HTML}{303F9F}
\definecolor{indigo800}{HTML}{283593}
\definecolor{indigo900}{HTML}{1A237E}

\definecolor{blue50}{HTML}{E3F2FD}
\definecolor{blue100}{HTML}{BBDEFB}
\definecolor{blue200}{HTML}{90CAF9}
\definecolor{blue300}{HTML}{64B5F6}
\definecolor{blue400}{HTML}{42A5F5}
\definecolor{blue500}{HTML}{2196F3}
\definecolor{blue600}{HTML}{1E88E5}
\definecolor{blue700}{HTML}{1976D2}
\definecolor{blue800}{HTML}{1565C0}
\definecolor{blue900}{HTML}{0D47A1}

\definecolor{lightblue50}{HTML}{E1F5FE}
\definecolor{lightblue100}{HTML}{B3E5FC}
\definecolor{lightblue200}{HTML}{81D4FA}
\definecolor{lightblue300}{HTML}{4FC3F7}
\definecolor{lightblue400}{HTML}{29B6F6}
\definecolor{lightblue500}{HTML}{03A9F4}
\definecolor{lightblue600}{HTML}{039BE5}
\definecolor{lightblue700}{HTML}{0288D1}
\definecolor{lightblue800}{HTML}{0277BD}
\definecolor{lightblue900}{HTML}{01579B}

\definecolor{cyan50}{HTML}{E0F7FA}
\definecolor{cyan100}{HTML}{B2EBF2}
\definecolor{cyan200}{HTML}{80DEEA}
\definecolor{cyan300}{HTML}{4DD0E1}
\definecolor{cyan400}{HTML}{26C6DA}
\definecolor{cyan500}{HTML}{00BCD4}
\definecolor{cyan600}{HTML}{00ACC1}
\definecolor{cyan700}{HTML}{0097A7}
\definecolor{cyan800}{HTML}{00838F}
\definecolor{cyan900}{HTML}{006064}

\definecolor{teal50}{HTML}{E0F2F1}
\definecolor{teal100}{HTML}{B2DFDB}
\definecolor{teal200}{HTML}{80CBC4}
\definecolor{teal300}{HTML}{4DB6AC}
\definecolor{teal400}{HTML}{26A69A}
\definecolor{teal500}{HTML}{009688}
\definecolor{teal600}{HTML}{00897B}
\definecolor{teal700}{HTML}{00796B}
\definecolor{teal800}{HTML}{00695C}
\definecolor{teal900}{HTML}{004D40}

\definecolor{green50}{HTML}{E8F5E9}
\definecolor{green100}{HTML}{C8E6C9}
\definecolor{green200}{HTML}{A5D6A7}
\definecolor{green300}{HTML}{81C784}
\definecolor{green400}{HTML}{66BB6A}
\definecolor{green500}{HTML}{4CAF50}
\definecolor{green600}{HTML}{43A047}
\definecolor{green700}{HTML}{388E3C}
\definecolor{green800}{HTML}{2E7D32}
\definecolor{green900}{HTML}{1B5E20}

\definecolor{lightgreen50}{HTML}{F1F8E9}
\definecolor{lightgreen100}{HTML}{DCEDC8}
\definecolor{lightgreen200}{HTML}{C5E1A5}
\definecolor{lightgreen300}{HTML}{AED581}
\definecolor{lightgreen400}{HTML}{9CCC65}
\definecolor{lightgreen500}{HTML}{8BC34A}
\definecolor{lightgreen600}{HTML}{7CB342}
\definecolor{lightgreen700}{HTML}{689F38}
\definecolor{lightgreen800}{HTML}{558B2F}
\definecolor{lightgreen900}{HTML}{33691E}

\definecolor{lime50}{HTML}{F9FBE7}
\definecolor{lime100}{HTML}{F0F4C3}
\definecolor{lime200}{HTML}{E6EE9C}
\definecolor{lime300}{HTML}{DCE775}
\definecolor{lime400}{HTML}{D4E157}
\definecolor{lime500}{HTML}{CDDC39}
\definecolor{lime600}{HTML}{C0CA33}
\definecolor{lime700}{HTML}{AFB42B}
\definecolor{lime800}{HTML}{9E9D24}
\definecolor{lime900}{HTML}{827717}

\definecolor{yellow50}{HTML}{FFFDE7}
\definecolor{yellow100}{HTML}{FFF9C4}
\definecolor{yellow200}{HTML}{FFF59D}
\definecolor{yellow300}{HTML}{FFF176}
\definecolor{yellow400}{HTML}{FFEE58}
\definecolor{yellow500}{HTML}{FFEB3B}
\definecolor{yellow600}{HTML}{FDD835}
\definecolor{yellow700}{HTML}{FBC02D}
\definecolor{yellow800}{HTML}{F9A825}
\definecolor{yellow900}{HTML}{F57F17}

\definecolor{amber50}{HTML}{FFF8E1}
\definecolor{amber100}{HTML}{FFECB3}
\definecolor{amber200}{HTML}{FFE082}
\definecolor{amber300}{HTML}{FFD54F}
\definecolor{amber400}{HTML}{FFCA28}
\definecolor{amber500}{HTML}{FFC107}
\definecolor{amber600}{HTML}{FFB300}
\definecolor{amber700}{HTML}{FFA000}
\definecolor{amber800}{HTML}{FF8F00}
\definecolor{amber900}{HTML}{FF6F00}

\definecolor{orange50}{HTML}{FFF3E0}
\definecolor{orange100}{HTML}{FFE0B2}
\definecolor{orange200}{HTML}{FFCC80}
\definecolor{orange300}{HTML}{FFB74D}
\definecolor{orange400}{HTML}{FFA726}
\definecolor{orange500}{HTML}{FF9800}
\definecolor{orange600}{HTML}{FB8C00}
\definecolor{orange700}{HTML}{F57C00}
\definecolor{orange800}{HTML}{EF6C00}
\definecolor{orange900}{HTML}{E65100}

\definecolor{deeporange50}{HTML}{FBE9E7}
\definecolor{deeporange100}{HTML}{FFCCBC}
\definecolor{deeporange200}{HTML}{FFAB91}
\definecolor{deeporange300}{HTML}{FF8A65}
\definecolor{deeporange400}{HTML}{FF7043}
\definecolor{deeporange500}{HTML}{FF5722}
\definecolor{deeporange600}{HTML}{F4511E}
\definecolor{deeporange700}{HTML}{E64A19}
\definecolor{deeporange800}{HTML}{D84315}
\definecolor{deeporange900}{HTML}{BF360C}

\definecolor{brown50}{HTML}{EFEBE9}
\definecolor{brown100}{HTML}{D7CCC8}
\definecolor{brown200}{HTML}{BCAAA4}
\definecolor{brown300}{HTML}{A1887F}
\definecolor{brown400}{HTML}{8D6E63}
\definecolor{brown500}{HTML}{795548}
\definecolor{brown600}{HTML}{6D4C41}
\definecolor{brown700}{HTML}{5D4037}
\definecolor{brown800}{HTML}{4E342E}
\definecolor{brown900}{HTML}{3E2723}

\definecolor{grey50}{HTML}{FAFAFA}
\definecolor{grey100}{HTML}{F5F5F5}
\definecolor{grey200}{HTML}{EEEEEE}
\definecolor{grey300}{HTML}{E0E0E0}
\definecolor{grey400}{HTML}{BDBDBD}
\definecolor{grey500}{HTML}{9E9E9E}
\definecolor{grey600}{HTML}{757575}
\definecolor{grey700}{HTML}{616161}
\definecolor{grey800}{HTML}{424242}
\definecolor{grey900}{HTML}{212121}

\definecolor{bluegrey50}{HTML}{ECEFF1}
\definecolor{bluegrey100}{HTML}{CFD8DC}
\definecolor{bluegrey200}{HTML}{B0BEC5}
\definecolor{bluegrey300}{HTML}{90A4AE}
\definecolor{bluegrey400}{HTML}{78909C}
\definecolor{bluegrey500}{HTML}{607D8B}
\definecolor{bluegrey600}{HTML}{546E7A}
\definecolor{bluegrey700}{HTML}{455A64}
\definecolor{bluegrey800}{HTML}{37474F}
\definecolor{bluegrey900}{HTML}{263238}

\definecolor{bluegrey900}{HTML}{263238}
\definecolor{darkest_blue}{HTML}{08306B}

\colorlet{neutral_color}{bluegrey900}
\colorlet{tensor_color}{bluegrey900}
\colorlet{weight_color}{bluegrey500}
\colorlet{shape_color}{bluegrey300}

\colorlet{encoder_color}{green300}
\colorlet{decoder_color}{red300}
\colorlet{mha_color}{cyan300}
\colorlet{mha_color_faded}{cyan200}
\colorlet{other_color}{bluegrey300}
\colorlet{special_color}{deeppurple300}

\tikzset{
    every picture/.append style={
        draw=neutral_color, 
        rounded corners=3pt
    },
    every node/.append style={
        text=neutral_color
    },
}

\pgfmathsetmacro{\ssp}{0.35}           %
\pgfmathsetmacro{\msp}{0.6}            %
\pgfmathsetmacro{\bsp}{1}              %
\pgfmathsetmacro{\mhaoffset}{0.6}
\pgfmathsetmacro{\boxpadding}{0.2}
\newcommand{\textoffset}{-0.1cm}

\tikzset{
    block/.style={
        draw, thick, rounded corners,
        minimum width=2.5cm, minimum height=0.6cm,
        fill=other_color,
    },
    mha block/.style={block, fill=mha_color},
    ffn block/.style={block, fill=indigo300},
    norm block/.style={block, fill=orange300},
    add/.style={draw, circle, minimum size=0.5cm, fill=grey300},
    main/.style={thick},
    main arrow/.style={thick, -{Latex[scale=1]}},
    bgbox/.style={draw, thick, inner sep=\boxpadding cm},
    groupbox/.style={draw, dashed, inner sep=\boxpadding cm},
    tensor/.style={color=tensor_color, font=\Large\bfseries},
    weight/.style={color=weight_color, font=\Large\bfseries},
    line/.style={color=neutral_color, thick},
    shape_annotation/.style={color=shape_color, font=\small},
    shape_annotation_box/.style={
        draw=shape_color, dashed, thick,
        fill=white, text=shape_color, font=\footnotesize,
    },
}

\pgfplotsset{
    every axis/.append style={
        grid style={thin, dashed, bluegrey300},
        grid=major,
        legend cell align=left,
        line width=1.25pt,
        mark size=1.8pt,
        legend style={font=\small},
        xlabel style={font=\small},
        ylabel style={font=\small},
        tick label style={font=\small},
        tick align=inside,
        xtick pos=left,
        ytick pos=bottom,
    },
}

\pgfplotsset{
    /pgfplots/colormap={ScatterDiverging}{
        rgb255=(57,73,171) rgb255=(156,164,213) rgb255=(255,255,255)
        rgb255=(249,168,142) rgb255=(244,81,30)
    },
    /pgfplots/colormap={ScatterMagma}{
        rgb255=(0,0,3) rgb255=(28,16,70) rgb255=(80,18,123)
        rgb255=(130,37,129) rgb255=(182,54,121) rgb255=(230,81,98)
        rgb255=(251,136,97) rgb255=(254,196,136) rgb255=(251,252,191)
    },
    /pgfplots/colormap={Blues}{
        rgb255=(247,251,255) rgb255=(222,235,247) rgb255=(198,219,239)
        rgb255=(158,202,225) rgb255=(107,174,214) rgb255=( 66,146,198)
        rgb255=( 33,113,181) rgb255=(  8, 81,156) rgb255=(  8, 48,107)
    },
    /pgfplots/colormap={BluesRev}{
        rgb255=(  8, 48,107) rgb255=(  8, 81,156) rgb255=( 33,113,181)
        rgb255=( 66,146,198) rgb255=(107,174,214) rgb255=(158,202,225)
        rgb255=(198,219,239) rgb255=(222,235,247) rgb255=(247,251,255)
    },
    /pgfplots/colormap={RdWtGn}{
        rgb255=(215, 48, 39) rgb255=(252,141, 89)
        rgb255=(255,255,255)
        rgb255=(145,207, 96) rgb255=( 26,152, 80)
    },
    /pgfplots/colormap={WtGn}{
        rgb255=(255,255,255) rgb255=(166,217,106) rgb255=( 26,152, 80)
    },
    /pgfplots/colormap={RdWt}{
        rgb255=(215, 48, 39) rgb255=(252,141, 89) rgb255=(255,255,255)
    },
}

\pgfplotsset{
  heatmap axis/.style={
    name=ax,
    /tikz/rounded corners=0pt,
    width=5.6cm, height=5.4cm, view={0}{90}, grid=none,
    xlabel={}, ylabel={},
    xtick={1,2,3,4}, xticklabels={\texttt{1},\texttt{2},\texttt{4},\texttt{8}},
    xticklabel pos=upper, tick style={draw=none}, xticklabel style={font=\large},
    ytick={1,2,3,4,5,6},
    yticklabels={\texttt{384},\texttt{192},\texttt{96},\texttt{48},\texttt{24},\texttt{12}},
    yticklabel style={font=\large},
    y dir=reverse, enlargelimits=false, axis on top,
    title style={yshift=1.5mm, font=\Large},
  },
  hm plot/.style={matrix plot*, point meta=explicit, mesh/rows=6, mesh/cols=4},
}

\pgfplotsset{
  attn axis/.style={
    /tikz/rounded corners=0pt,
    width=5.6cm, height=5.4cm, view={0}{90}, grid=none,
    xlabel={Decoder}, ylabel={Encoder},
    xlabel style={at={(0.5,1)}, anchor=south, yshift=4mm, font=\large},
    ylabel style={at={(0,0.5)}, anchor=south, yshift=4mm, font=\large},
    xtick={1,2,3}, xticklabels={\texttt{eff},\texttt{flash},\texttt{swin}},
    xticklabel style={font=\large}, xticklabel pos=upper, tick style={draw=none},
    ytick={1,2,3}, yticklabels={\texttt{eff},\texttt{flash},\texttt{swin}},
    yticklabel style={font=\large, rotate=90},
    y dir=reverse, enlargelimits=false, axis on top,
    title style={yshift=6mm, font=\Large},
  },
  hm3 plot/.style={matrix plot*, point meta=explicit, mesh/rows=3, mesh/cols=3},
}

\def\dimension{d}
\def\dmodel{\dimension_m}
\def\dk{\dimension_k}
\def\dv{\dimension_v}

\def\height{H}
\def\width{W}

\def\patchsize{P}
\def\windowsize{M}

\def\seqlen{N}

\DeclareMathOperator{\softmax}{Softmax}

\DeclareMathOperator{\gelu}{GELU}
\def\concat{\texttt{Concat}}

\def\upsample{\texttt{Up}}
\DeclareMathOperator{\attention}{Attention}

\def\mha{\texttt{MHA}}

\def\ffn{\texttt{FFN}}

\def\layernorm{\texttt{LN}}

\def\linear{\texttt{Linear}}
\def\conv{\texttt{Conv}}
\def\dwconv{\texttt{DW-Conv}}
\def\avgpool{\texttt{AvgPool}}
\def\patchembed{\texttt{PatchEmbed}}

\def\patchexpand{\texttt{PatchExpand}}
\def\mixffn{\texttt{Mix-FFN}}

\def\swin{\texttt{swin}}

\def\eff{\texttt{eff}}
\def\flash{\texttt{flash}}
\def\shiftedwindow{\texttt{swin}}

\def\1{\bm{1}}

\newcommand{\R}{\mathbb{R}}

\def\mA{{\bm{A}}}    \def\mC{{\bm{C}}}

  \def\mK{{\bm{K}}}  
    \def\mO{{\bm{O}}}
  \def\mQ{{\bm{Q}}}  
\def\mS{{\bm{S}}}    
\def\mV{{\bm{V}}}  \def\mW{{\bm{W}}}  \def\mX{{\bm{X}}}
\def\mY{{\bm{Y}}}

\DeclareMathAlphabet{\mathsfit}{\encodingdefault}{\sfdefault}{m}{sl}
\SetMathAlphabet{\mathsfit}{bold}{\encodingdefault}{\sfdefault}{bx}{n}

\def\eqref#1{equation~\ref{#1}}

\copyrightyear{2026}
\acmYear{2026}
\setcopyright{cc}
\setcctype{by}
\acmConference[SIGSPATIAL '26]{The 34th ACM International Conference on Advances in Geographic Information Systems}{November 03--06, 2026}{Riverside, CA, USA}
\acmBooktitle{The 34th ACM International Conference on Advances in Geographic Information Systems (SIGSPATIAL '26), November 03--06, 2026, Riverside, CA, USA}
\acmDOI{10.1145/3841645.3842978}
\acmISBN{979-8-4007-2950-8/2026/11}

\begin{document}

\title[Pixel-Level Transformers]{Pixel-Level Transformers in Remote Sensing:\protect\\A Canopy Height Case Study}

\author{Sven Ligensa}
\email{sven.ligensa@uni-muenster.de}
\orcid{0009-0007-4653-2394}
\affiliation{%
  \institution{University of M\"unster}
  \city{Münster}
  \country{Germany}
}

\author{Jan Pauls}
\email{jan.pauls@uni-muenster.de}
\orcid{0009-0001-8367-2501}
\affiliation{%
  \institution{University of M\"unster}
  \city{Münster}
  \country{Germany}
}

\author{Karsten Schr\"odter}
\email{karsten.schroedter@uni-muenster.de}
\orcid{0009-0005-8010-7063}
\affiliation{%
  \institution{University of M\"unster}
  \city{Münster}
  \country{Germany}
}

\author{Ibrahim Fayad}
\email{ibrahim.fayad@lsce.ipsl.fr}
\orcid{0000-0001-7504-5623}
\affiliation{%
  \institution{Laboratoire des Sciences du Climat et de l’Environnement}
  \city{Paris}
  \country{France}
}

\author{Fabian Gieseke}
\email{fabian.gieseke@uni-muenster.de}
\orcid{0000-0001-7093-5803}
\affiliation{%
  \institution{University of M\"unster}
  \city{Münster}
  \country{Germany}
}

\renewcommand{\shortauthors}{Ligensa et al.}

\begin{abstract}
    Predicting canopy height from medium-resolution satellite imagery is a common and scalable approach for assessing the condition of the world's forests, which play a crucial role in climate change mitigation.
    While Transformer-based architectures have shown strong performance in many domains, their straightforward application to dense (i.e., pixel-level) regression tasks often yields suboptimal results.
    In particular, the patch size has a crucial impact on the model performance.
    In this work, we consider pixel-level attention schemes and show that the resulting models generally outperform those relying on larger patch sizes.
    However, pixel-level attention can be a prohibitively resource-intensive operation.
    For this reason, we conduct an extensive experimental study using efficient attention variants to identify favorable trade-offs between prediction quality and resource requirements, facilitating the practical deployment of the proposed models.
    In addition, we perform a comprehensive comparison with several well-established models in the field and show that, with suitable hyperparameter choices, Transformer-based architectures can outperform competing approaches.
    Our findings provide practical guidance for designing models for pixel-level regression tasks on medium-resolution satellite imagery, including canopy height and biomass estimation, soil moisture mapping, and yield forecasting.
\end{abstract}



\maketitle

\begin{figure*}[t!]
    \centering
    \input{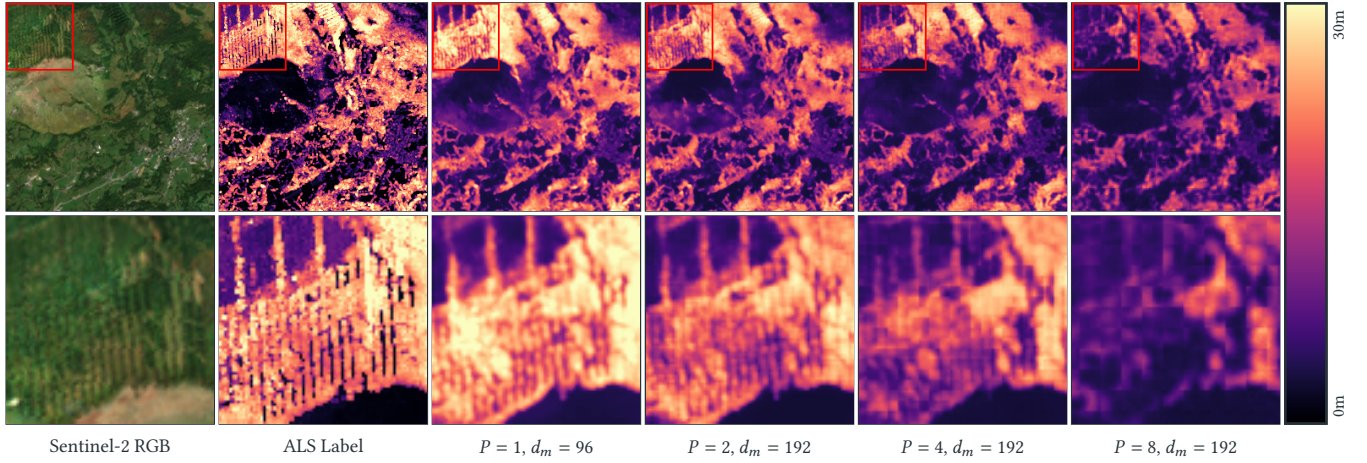}
    \vspace{-\baselineskip}
    \caption{Qualitative comparison of pixel-level regression by Transformer-based models with varying patch size ($\patchsize$) and model dimension ($\dmodel$). The first column shows the input image (only RGB channels shown, while the models are trained on twelve channels), the second column shows Airborne Laser Scanning (ALS) canopy height labels, and the following columns show the predictions of models with increasingly larger patch sizes. Especially for areas with fine details, models with smaller patch sizes produce predictions of higher perceptual quality. Further details are provided in Section~\ref{sec:experiments} and additional examples are given in the appendix.}
    \Description{This example patch features trees of varying heights, sometimes with fine details, which are only well identified by models with a patch size of one or two.}
	\label{fig:qualitative_patchsizes}
\end{figure*}

\section{Introduction}

Forests play a key role in the carbon cycle and in regulating climate change~\cite{Bonan2008ForestsAndClimateChange}. One important component of forest carbon stocks is above-ground biomass~(AGB), which refers to the mass of living vegetation located above the soil surface, including stems, branches, bark, foliage, and shrubs~\cite{IPCC2006}.
While collecting field estimates of AGB is time-consuming and expensive, satellite imagery is widely available and permits the estimation of, e.g., tree height~\cite{Drusch2012Sentinel2}, which can in turn be used to estimate AGB via the approximate power-law relationship between canopy height and AGB~\cite{Schwartz2023FORMS}.
For that reason, machine learning and in particular deep learning approaches have been employed extensively in the past to generate global-scale canopy height maps~\cite{Pauls2026ECHOSAT, Tolan2024VeryHighRes, Pauls2024EstimatingCanopyHeight, Lang2023HighresolutionCanopyHeight, Potapov2021MappingGlobalForest}.

Generating a canopy height map of global scale is a significant engineering challenge, requiring a considerable amount of resources for the inference~\cite{Pauls2024EstimatingCanopyHeight}. 
The goal is to train a model, which achieves high predictive performance, while being efficient in terms of having a short training and inference time. While for high-resolution images, the semantic usually does not change significantly among a few neighbored pixels, it is different for dense prediction on medium-resolution satellite imagery, like Sentinel-2 data, where a single pixel typically corresponds to an area of $10 \times 10$ meters~\cite{Drusch2012Sentinel2}.

Recently, the importance of pixel-level attention was stressed for general Vision Transformer (ViT) models~\cite{Nguyen2025PixelLevelViT,Wang2025ScalingLawsInPatchification}. In this work, we investigate the importance and effectiveness of pixel-level ViT-based architectures for dense prediction tasks in remote sensing, using canopy height estimation as a representative large-scale regression task.
The patch size~$\patchsize \in \mathbb{N}$ is a critical hyperparameter for ViT-based architectures, as it affects their memory and runtime complexity as well as the prediction quality~\cite{Wang2025ScalingLawsInPatchification}. Figure~\ref{fig:qualitative_patchsizes} shows exemplary pixel-level tree-height predictions on satellite images for models with varying patch size.
Choosing a smaller $\patchsize$ leads to a larger number of tokens $\seqlen = \frac {\height \width} {\patchsize^2}$ to be processed by ViT-based models, where $\width,\height \in \mathbb{N}$ are the image's width and height. With vanilla attention, memory and runtime complexity increase quadratically with the sequence length~\cite{Vaswani17,Dosovitskiy21}. For the ViT introduced by \citet{Dosovitskiy21}, selecting a comparably large patch size of $\patchsize=16$ was necessary to be able to process reasonably-sized images (e.g., $224 \times 224$ pixels). They also observed that smaller patches sizes generally improved performance; however the quadratic scaling behavior of the attention mechanism with regard to the sequence length made a smaller patch size prohibitive in practice~\cite{Dosovitskiy21}.
With the introduction of more efficient attention mechanisms, such as shifted-window attention, the patch size was reduced to $\patchsize=4$, which made the ViT-architectures more suitable for dense prediction tasks~\cite{Liu2021Swin, Xie2021SegFormer}.

To investigate the role of pixel-level ViTs for dense prediction in remote sensing, we conduct an extensive empirical study on canopy height estimation.\footnote{To facilitate reproducibility and future research, the code of our experimental evaluation is publicly available at \url{https://github.com/SvenLigensa/pixel-level-transformers}.} Our contributions are as follows:
\begin{enumerate}
    \item We give practical guidance on choosing crucial hyperparameters of ViTs (like $\patchsize$ or the model dimension $\dmodel$) for dense regression tasks on medium-resolution satellite images.
    \item We unveil shortcomings of traditional quantitative evaluation of canopy height predictions, based on noisy labels.
    \item We propose a new model configuration that outperforms other well-known models on canopy height prediction.
\end{enumerate}

\section{Background}

In this section, we first give an overview of related work on ViTs, recent work on pixel-level ViTs, and proposed variants for reducing the attention mechanism's complexity. Then, specifics of the canopy height prediction task are revised.

\subsection{Vision Transformers}

The original ViT~\cite{Dosovitskiy21} adapted the Transformer architecture~\cite{Vaswani17} to the task of image classification via minimal changes, giving rise to a whole new family of Vision Transformers. Hierarchical Transformers improved on the ViT's two main weaknesses (1) by introducing hierarchical processing to account for visual elements of varying scale and (2) by reducing $\patchsize$ from $16$ to $4$ pixels, which was facilitated by improving the attention computation.

The two most well-known, concurrently developed, architectures are the so-called Swin Transformer~\cite{Liu2021Swin} and the Pyramid Vision Transformer (PVT)~\cite{Wang2021PVT}.
Both architectures comprise four stages---each consisting of multiple Transformer layers---operating on increasingly coarser spatial resolution.
They differ, however, with regard to the attention mechanism used, and the way adjacent tokens are combined in between stages.
The Swin Transformer uses shifted-window attention, which restricts attention to local windows applied with two alternating positions~\cite{Liu2021Swin}, while the PVT reduces the number of keys and values and thereby the number of computations of the attention by a constant factor~\cite{Wang2021PVT}.
To reduce the resolution between stages, the Swin Transformer merges the four patches inside a $2 \times 2$ window by concatenating them along the embedding dimension, and then linearly projecting them to twice the previous embedding dimension~\cite{Liu2021Swin}.
The PVT reduces the resolution by applying an overlapping patch embedding layer with a patch size of two at the beginning of every stage~\cite{Wang2021PVT}.

For Transformers performing dense prediction tasks like semantic segmentation (pixel-level classification) and canopy height prediction (pixel-level regression), the output needs to have the same spatial dimension as the input.
To achieve that, the architectures generally follow the encoder-decoder paradigm, where the decoder constitutes the counterpart of the encoder, and increases the spatial dimensionality~\cite{Zheng2021SETR, Xie2021SegFormer, Cao2022SwinUnet}. Note that as most models choose a patch size of $P=4$ up to $P=16$ pixels, the decoder's output resolution is still smaller than the input resolution. To obtain a prediction for every pixel of the input image, it is common practice to perform non-parametric upsampling (e.g. bilinear interpolation)~\cite{Zheng2021SETR} or parametric upsampling (e.g. transposed convolution or patch expand)~\cite{Cao2022SwinUnet}. The architectures mainly differ in how they combine the hierarchical feature maps produced by the encoder.
Popular Segmentation Transformers are the SETR~\cite{Zheng2021SETR}, SegFormer~\cite{Xie2021SegFormer}, and the Swin-Unet~\cite{Cao2022SwinUnet}. A recently proposed variant is the U-MixFormer~\cite{Yeom2025UMixFormer}, fusing encoder and decoder features by using mix-attention, which allows the encoder feature to query the decoder feature hierarchy.

\subsection{Pixel-Level Vision Transformers}

Recent work targeting pixel-level ViTs include \citet{Nguyen2025PixelLevelViT} and \citet{Wang2025ScalingLawsInPatchification}. \citet{Nguyen2025PixelLevelViT} first demonstrated the effectiveness of treating each pixel as a token. Due to the computational complexity, they focused on small-scale datasets like CIFAR-100. They primarily analyzed the results from the perspective of removing the locality inductive bias~\cite{Nguyen2025PixelLevelViT}.
\citet{Wang2025ScalingLawsInPatchification} performed additional experiments, thoroughly testing the impact of the patch size hyperparameter across various vision tasks, input scales, and architectures. The emerging patterns are called patchification scaling laws, stating that decreasing patch sizes generally result in lower test loss.\footnote{Similar to \citet{Kaplan2020ScalingLaws}, who showed that the test loss scales according to a power-law w.r.t. both model size and dataset size.}
In ViT-based architectures, a patch is the ``basic unit of operation after the first projection layer''~\cite{Nguyen2025PixelLevelViT}, i.e., all pixels of the same patch are considered jointly throughout the whole model.
The patchification step can be viewed as a compression step, resulting in possibly irreversible information loss; operating on a pixel-level can avoid this bottleneck~\cite{Wang2025ScalingLawsInPatchification}. This compression is stronger the larger the patch size and the smaller the model dimension.

Still, training ViT-based models on tens to hundreds of thousands of tokens per image\footnote{Images of size $256 \times 256$ ($512 \times 512$) pixels already give rise to $65\,536$ ($262\,144$) tokens.} is challenging, as also noted by~\cite{Nguyen2025PixelLevelViT}.
The following section gives a short overview of attention variants, which reduce the memory and runtime complexity of the operation, making pixel-level ViTs more efficient in practice.

\subsection{Attention Mechanism Variants}

Dozens of attention mechanism variants have been proposed to alleviate the memory (and runtime) complexity of the vanilla attention by \citet{Vaswani17}, which is quadratic in the number of tokens~$N$.
Sparse attention reduces the number of tokens that can interact with each other. The pattern of interacting tokens can either be fixed or dynamic. Examples for fixed sparse attention mechanisms are local window attention~\cite{Yuan2021HRFormer}, shifted window attention~\cite{Liu2021Swin}, and attention along axes~\cite{Ho2019Axial,Dong2022CSWin}.
Dynamic (i.e., data-dependent) patterns include clustering the tokens via locality-sensitive hashing~\cite{Kitaev2020Reformer,Vyas2020ClusteredAttention} or $k$-means~\cite{Chen2020iGPT,Roy2021RoutingTransformer,Vyas2020ClusteredAttention}, and only calculating attention among tokens of the same cluster.
Low-rank approximations are based on the empirical observation that the self-attention matrix can often be approximated via a low-rank matrix, meaning that it can be efficiently represented as the product of smaller low-dimensional matrices~\cite{Wang2020Linformer}. Examples are the Linformer~\cite{Wang2020Linformer}, Linear Transformer~\cite{Katharopoulos2020LinearTransformer}, Performer~\cite{Choromanski2021Performer}, random feature attention~\cite{Peng2021RandomFeatureAttention}, and efficient attention~\cite{Shen2021Efficient}. A more detailed description of the attention mechanisms considered in this work is given in Section~\ref{sec:attention_mechanisms}.

\subsection{Canopy Height Prediction}

To analyze pixel-level ViTs for dense prediction in remote sensing, we consider canopy height prediction, a key task in Earth observation applications, where the goal is to predict the height of the (usually) tallest tree for each pixel of a remote sensing image, e.g., a medium-resolution satellite image. Creating such canopy height maps at a large scale relies on satellites providing spatially and temporally continuous and consistent images. Two common data sources are the Sentinel-2 mission operated by the ESA~\cite{Drusch2012Sentinel2,sentinel2_l2a_c1} and the Landsat mission operated by the NASA~\cite{Landsat20}. The corresponding satellites provide, among other things, optical, i.e. passive, multi-spectral images at a resolution of up to \SI{10}{m} and \SI{30}{m}, respectively.
The Sentinel-2 mission comprises two active satellites in the same sun-synchronous orbit, phased half an orbit apart~\cite{Drusch2012Sentinel2}. The revisit time is 5 days at the equator and resolutions range from \SI{10}{m} to \SI{60}{m} depending on the band. The imaging instrument of \mbox{Sentinel-2} captures 13 bands between \SI{433}{nm} and \SI{2280}{nm}, including RGB, near infrared and others~\cite{Drusch2012Sentinel2}.

Commonly used canopy height labels are either derived from space-borne instruments, such as Global Ecosystem Dynamics Investigation (GEDI)~\cite{Dubayah2020} or ICESat~\cite{Markus2017ICESat}, or measurements from Airborne Laser Scanning (ALS) campaigns~\cite{Fogel2025OpenCanopy}. While the latter ones exhibit a much higher vertical accuracy and spatial resolution, both their spatial and temporal coverage is heavily limited. For large-scale applications, models are typically trained on data from space-borne instruments, such as the GEDI mission, as it provides data from 2019 onwards and exhibits a global coverage between $51.6^\circ N$ and $51.6^\circ S$. The GEDI system~\cite{Dubayah2020} is a set of three full-waveform LiDAR sensors mounted on the ISS that capture eight parallel tracks (two power beams measuring on two tracks each and one coverage beam measuring on four tracks) with a \SI{600}{m} inter-track and \SI{60}{m} intra-track  spacing. Each measurement has an approximate diameter of \SI{25}{m} (slightly varying depending on the ISS height) and starting with the second release of the GEDI product, the geolocation accuracy has met its target of being unbiased (i.e., having a mean of zero) and having a standard deviation of less than \SI{10}{m}. To convert the waveform of each measurement into a scalar label, two commonly used metrics are the relative height~(rh) at which $95\%$~(rh95) and $98\%$~(rh98) of the cumulative returned energy was received.

ALS data on the other hand does not suffer from low geolocation accuracy or low vertical measurement quality. Typically captured by LiDAR sensors mounted on airplanes, they exhibit a resolution of \SI{1}{m} up to \SI{10}{cm}. Also, in contrast to GEDI's sparse measurements, LiDAR sensors generally provide dense measurements.\footnote{For the task of canopy height estimation, OpenCanopy~\cite{Fogel2025OpenCanopy} provides a dataset of combined optical satellite data and unified ALS campaign over France, which will be used for evaluation in our experiments.}

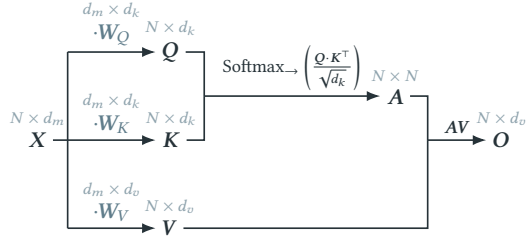
\begin{figure}[t]
    \centering
    \resizebox{!}{3.25cm}{%
        \begin{tikzpicture}[every path/.append style={rounded corners=0pt}]
    \node[tensor] (X) at (0,0) {$\mX$};
    \node[shape_annotation, above=\textoffset of X] {$\seqlen \times \dmodel$};

    \path (X.east) ++(0.25cm,0) coordinate (helper2);
    \path (helper2.north) ++(0,1.5cm) coordinate (helper1);
    \path (helper2.south) ++(0,-1.5cm) coordinate (helper3);

    \node[tensor, right=1.5cm of helper1] (Q) {$\mQ$};
    \node[shape_annotation, above=\textoffset of Q] {$\seqlen \times \dk$};

    \node[tensor, right=1.5cm of helper2] (K) {$\mK$};
    \node[shape_annotation, above=\textoffset of K] {$\seqlen \times \dk$};

    \node[tensor, right=1.5cm of helper3] (V) {$\mV$};
    \node[shape_annotation, above=\textoffset of V] {$\seqlen \times \dv$};

    \node[weight, anchor=south] (WQ) at ($ (helper1.east) !0.5! (Q.west) $) {$\cdot \mW_Q$};
    \node[shape_annotation, above=\textoffset of WQ] (WQ_shape) {$\dmodel \times \dk$};

    \node[weight, anchor=south] (WK) at ($ (helper2.east) !0.5! (K.west) $) {$\cdot \mW_K$};
    \node[shape_annotation, above=\textoffset of WK] {$\dmodel \times \dk$};

    \node[weight, anchor=south] (WV) at ($ (helper3.east) !0.5! (V.west) $) {$\cdot \mW_V$};
    \node[shape_annotation, above=\textoffset of WV] {$\dmodel \times \dv$};

    \path let \p1 = (Q), \p2 = (K) in coordinate (midQK) at (\x2, {(\y1+\y2)/2});
    \path (K.east) ++(0.3cm,0) coordinate (Kshift);
    \path let \p1 = (Kshift), \p2 = (midQK) in coordinate (helper4) at (\x1, \y2);

    \node[tensor, right=3cm of helper4] (A) {$\mA$};
    \node[shape_annotation, above=\textoffset of A] {$\seqlen \times \seqlen$};

    \node[anchor=south] (A_Calc) at ($ (helper4.east) !0.5! (A.west) $) {$\softmax_\rightarrow\left(\frac{\textcolor{tensor_color}{\mQ \cdot \mK^\top}}{\sqrt{\dk}}\right)$};

    \path let \p1 = (X), \p2 = (V) in coordinate (midAV) at (\x2, \y1);    
    \path (A.east) ++(0.3cm,0) coordinate (Ashift);
    \path let \p1 = (Ashift), \p2 = (midAV) in coordinate (helper5) at (\x1, \y2);

    \node[tensor, right=1cm of helper5] (O) {$\mO$};
    \node[shape_annotation, above=\textoffset of O] {$\seqlen \times \dv$};

    \node[text=tensor_color, anchor=south] (O_Calc) at ($ (helper5.east) !0.5! (O.west) $) {$\mA\mV$};

    \draw[line] (X) -- (helper2);
    \draw[line] (helper2) -- (helper1);
    \draw[line] (helper2) -- (helper3);
    \draw[line, -Latex] (helper1) -- (Q);
    \draw[line, -Latex] (helper2) -- (K);
    \draw[line, -Latex] (helper3) -- (V);

    \draw[line] (Q) -| (helper4);
    \draw[line] (K) -| (helper4);
    \draw[line, -Latex] (helper4) -- (A);

    \draw[line] (A) -| (helper5);
    \draw[line] (V) -| (helper5);
    \draw[line, -Latex] (helper5) -- (O);
\end{tikzpicture}
    }
    \caption{Computations of a vanilla attention head~\cite{Vaswani17}.}
    \Description{The computations are shown in form of a computation graph with the input X, intermediary variables Q (query), K (key), and V (value), and A (attention matrix), and the output O.}
    \label{fig:vanilla_attention}
\end{figure}

Estimating canopy height from remote sensing data has been established as a reliable source for large-scale forest monitoring, as it can be applied across vast areas more easily than field-based inventories. Such canopy height maps have been created at regional~\cite{SCHWARTZ2024103711}, national~\cite{Fayad24HyTec,Schwartz2023FORMS,Kacic23}, continental~\cite{Liu2023} and global~\cite{Pauls2024EstimatingCanopyHeight,Lang2023HighresolutionCanopyHeight,Pauls2026ECHOSAT} scale and resort to either classical machine learning methods~\cite{Potapov2021MappingGlobalForest,Kacic23}, convolutional neural networks~\cite{SCHWARTZ2024103711,Schwartz2023FORMS,Pauls2024EstimatingCanopyHeight,wagner2025highresolutiontreeheight} or attention-based architectures~\cite{Fayad24HyTec,Pauls2026ECHOSAT}. Although not perfectly matching ALS measurements, these maps have been shown to be reliable estimators for various downstream tasks~\cite{Lukevs2026}.

\section{Attention Mechanisms}\label{sec:attention_mechanisms}

In this section, we present and compare popular attention variants with the vanilla attention mechanism of the Transformer architecture~\cite{Vaswani17}. To simplify notation, we focus on a single attention head. In practice, Transformers use Multi-Head Attention (\mha) layers, which include multiple attention heads performing attention independently and in parallel. We refer to~\citet{Vaswani17} for further details.

\subsection{Notation}
\label{sec:notation}

Let $\mX \in \R^{\seqlen \times \dmodel}$ be an input sequence of $\seqlen \in \mathbb{N}$ token embeddings, each of dimension $\dmodel \in \mathbb{N}$. Let $\dk \in \mathbb{N}$ and $\dv \in \mathbb{N}$ denote the embedding dimensions for the keys and queries, and the values, respectively. The query-, key- and value-transformation matrices are denoted by   
$\mW_Q \in \R^{\dmodel \times \dk}$, 
$\mW_K \in \R^{\dmodel \times \dk}$, and 
$\mW_V \in \R^{\dmodel \times \dv}$. Accordingly, the query, key and value features are defined by $\mQ \coloneq \mX \cdot \mW_Q \in \R^{\seqlen \times \dk}$, $\mK \coloneq \mX \cdot \mW_K \in \R^{\seqlen \times \dk}$ and $\mV \coloneq \mX \cdot \mW_V \in \R^{\seqlen \times \dv}$. 

The row-wise softmax function $\softmax_\rightarrow: \R^{\seqlen \times \seqlen} \to \R^{\seqlen \times \seqlen}$ normalizes the element-wise exponential of the matrix in a row-wise manner, i.e.,
$\softmax_\rightarrow(\mS)_{i,j} = \exp(\mS_{i,j}) / \sum_{l=1}^{\seqlen} \exp(\mS_{i,l})$
for a score matrix $\mS \in \R^{\seqlen \times \seqlen}$ and indices $1 \le i, j \le \seqlen$. Similarly, the column-wise softmax $\softmax_\downarrow: \R^{\seqlen \times \seqlen} \to \R^{\seqlen \times \seqlen}$ normalizes the exponential column-wise.

\subsection{Vanilla Attention}

We shortly revisit the original formulation of scaled-dot product attention as introduced by \citet{Vaswani17}, which we will refer to as vanilla attention in this work. In vanilla attention, the query features $\mQ \in \R^{\seqlen \times \dk}$ and key features $\mK  \in \R^{\seqlen \times \dk}$ are used to
calculate 
the attention matrix 
\begin{equation}
    \mA \coloneq \softmax_\rightarrow \left(\frac{\mQ \cdot \mK^\top}{\sqrt{\dk}}\right) \in \R^{\seqlen \times \seqlen}.
\end{equation}
The output $\mO$ of the attention head is then calculated by multiplying the attention matrix with the value features, i.e. 
\begin{equation}
    \attention_{\texttt{vanilla}}(\mQ,\mK,\mV) \coloneq \mO \coloneq \mA \cdot \mV 
    \in \R^{\seqlen \times \dv}.
\end{equation}

The computation flow of vanilla attention is visualized in Figure~\ref{fig:vanilla_attention}.
Vanilla attention has a runtime complexity of $\mathcal O(\seqlen^2 \dmodel)$ and a memory complexity of $\mathcal O(\seqlen^2)$, both of which are dominated by computing/storing the attention matrix.\footnote{We assume that $\seqlen > \dmodel$, the typical case in practice. Then, the memory required for the projections is dominated by that of the attention matrix.}

The underlying intuition stems from the observation that inside $\mQ \cdot \mK^\top$ each of the $\seqlen$ query vectors is compared to all $\seqlen$ key vectors via a dot-product and, after scaling, is normalized using the softmax operation. Therefore, the rows of the attention matrix $\mA \in \R^{\seqlen \times \seqlen}$ can be interpreted as indicating how much each key contributes to solving the row's query. The corresponding attention weights are then used to obtain, for each query, a weighted sum of the corresponding values.

\begin{figure}[t]
    \centering
    \resizebox{!}{3.25cm}{%
        \begin{tikzpicture}[every path/.append style={rounded corners=0pt}]
    \node[tensor] (X) at (0,0) {$\mX$};
    \node[shape_annotation, above=\textoffset of X] {$\seqlen \times \dmodel$};

    \path (X.east) ++(0.25cm,0) coordinate (helper2);
    \path (helper2.north) ++(0,1.5cm) coordinate (helper1);
    \path (helper2.south) ++(0,-1.5cm) coordinate (helper3);

    \node[tensor, right=1.5cm of helper1] (Q) {$\mQ$};
    \node[shape_annotation, above=\textoffset of Q] {$\seqlen \times \dk$};

    \node[tensor, right=1.5cm of helper2] (K) {$\mK$};
    \node[shape_annotation, above=\textoffset of K] {$\seqlen \times \dk$};

    \node[tensor, right=1.5cm of helper3] (V) {$\mV$};
    \node[shape_annotation, above=\textoffset of V] {$\seqlen \times \dv$};

    \node[weight, anchor=south] (WQ) at ($ (helper1.east) !0.5! (Q.west) $) {$\cdot \mW_Q$};
    \node[shape_annotation, above=\textoffset of WQ] (WQ_shape) {$\dmodel \times \dk$};

    \node[weight, anchor=south] (WK) at ($ (helper2.east) !0.5! (K.west) $) {$\cdot \mW_K$};
    \node[shape_annotation, above=\textoffset of WK] {$\dmodel \times \dk$};

    \node[weight, anchor=south] (WV) at ($ (helper3.east) !0.5! (V.west) $) {$\cdot \mW_V$};
    \node[shape_annotation, above=\textoffset of WV] {$\dmodel \times \dv$};

    \node[tensor, right=3cm of Q] (Q_norm) {$\mQ$};
    \node[shape_annotation, above=\textoffset of Q_norm] {$\seqlen \times \dk$};
    \node[anchor=south] (softmaxrow) at ($ (Q.east) !0.5! (Q_norm.west) $) {$\softmax_\rightarrow (\textcolor{tensor_color}{\mQ})$};

    \node[tensor, right=3cm of K] (K_norm) {${\mK}^\top$};
    \node[shape_annotation, above=\textoffset of K_norm] {$\dk \times \seqlen$};
    \node[anchor=south] (softmaxcol) at ($ (K.east) !0.5! (K_norm.west) $) {$\softmax_\downarrow(\textcolor{tensor_color}{\mK})^\top$};

    \path let \p1 = (K_norm), \p2 = (V) in coordinate (midKV) at (\x2, {(\y1+\y2)/2});
    \path (K_norm.east) ++(0.3cm,0) coordinate (Kshift);
    \path let \p1 = (Kshift), \p2 = (midKV) in coordinate (helper4) at (\x1, \y2);

    \node[tensor, right=1.2cm of helper4] (Context) {$\mC$};
    \node[shape_annotation, above=\textoffset of Context] {$\dk \times \dv$};

    \node[color=tensor_color, anchor=south] (Context_Calc) at ($ (helper4.east) !0.5! (Context.west) $) {${\mK}^\top \mV$};

    \path let \p1 = (X), \p2 = (Q_norm) in coordinate (midQContext) at (\x2, \y1);
    \path (Context.east) ++(0.3cm,0) coordinate (Contextshift);
    \path let \p1 = (Contextshift), \p2 = (midQContext) in coordinate (helper5) at (\x1, \y2);

    \node[tensor, right=1.2cm of helper5] (O) {$\mO$};
    \node[shape_annotation, above=\textoffset of O] {$\seqlen \times \dv$};

    \node[text=tensor_color, anchor=south] (O_Calc) at ($ (helper5.east) !0.5! (O.west) $) {$\mQ\mC$};

    \draw[line] (X) -- (helper2);
    \draw[line] (helper2) -- (helper1);
    \draw[line] (helper2) -- (helper3);
    \draw[line, -Latex] (helper1) -- (Q);
    \draw[line, -Latex] (helper2) -- (K);
    \draw[line, -Latex] (helper3) -- (V);
    
    \draw[line, -Latex] (Q) -- (Q_norm);
    \draw[line, -Latex] (K) -- (K_norm);

    \draw[line] (K_norm) -| (helper4);
    \draw[line] (V) -| (helper4);
    \draw[line, -Latex] (helper4) -- (Context);

    \draw[line] (Q_norm) -| (helper5);
    \draw[line] (Context) -| (helper5);
    \draw[line, -Latex] (helper5) -- (O);

\end{tikzpicture}
    }
    \caption{Computations of an efficient attention head~\cite{Shen2021Efficient}.}
    \Description{The computations are shown in form of a computation graph with the input X, intermediary variables Q (query), K (key), and V (value), and C (context matrix), and the output O.}
    \label{fig:efficient_attention}
\end{figure}
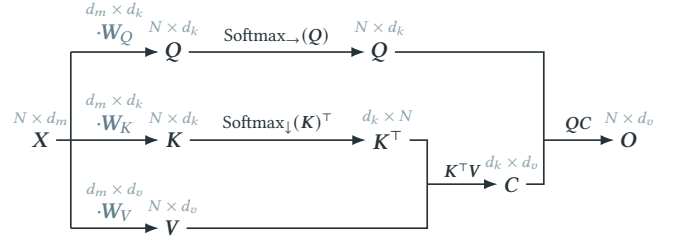

\subsection{Efficient Attention}

Efficient attention~\cite{Shen2021Efficient} reduces the runtime complexity to $\mathcal O(\seqlen \dmodel^2)$ and the memory complexity to $\mathcal O(\seqlen \dmodel + \dmodel^2)$ by slightly rearranging the computations and at the cost of only providing an approximation of vanilla attention.\footnote{This small yet crucial change results in an immense practical performance gain, as usually $\seqlen \gg \dmodel$.} The underlying observation is that matrix multiplication is associative and that, up to the softmax operation, vanilla attention is just a scaled multiplication of $\mQ$, $\mK^\top$ and $\mV$, where the computational bottleneck lies in the materialization of $\mQ \cdot \mK^\top \in \R^{\seqlen \times \seqlen}$. The basic idea is to change the order of computation by first calculating $\mK^\top \cdot \mV \in \R^{\dk \times \dv}$. On the way, the normalization via the softmax function needs to be adjusted. More precisely, efficient attention first calculates a column-wise softmax of the key features $\mK$ and then computes
\begin{equation}
    \mC \coloneq \softmax_\downarrow(\mK)^\top \cdot \mV \in \R^{\dk \times \dv},
\end{equation}
which is multiplied with a row-wise softmax-normalized version of the query features, so that the output $\mO$ is given by 
\begin{equation}
    \attention_{\eff}(\mQ,\mK,\mV) \coloneq \mO \coloneq \softmax_\rightarrow(\mQ) \cdot \mC \in \R^{\seqlen \times \dv}.
\end{equation}
Figure~\ref{fig:efficient_attention} illustrates the underlying computations in form of a graph.
\citet{Shen2021Efficient} provide the following intuition for the modified attention computations: 
The modified keys, i.e., $\softmax_\downarrow(\mK) \in \R^{\seqlen \times \dk}$, can be seen as $\dk$ feature-maps, which indicate how a given feature is distributed across the image. They guide where to aggregate information from. The learned features could be of lower level (e.g., edge orientation) or higher level (e.g., human-made structures), depending on their location in the network.
The values, $\mV \in \R^{\seqlen \times \dv}$, containing the content of the patches, are passed on through the network (after being weighted).
The product of the keys and values is the ``global context'' $\mC \in \R^{\dk \times \dv}$.
It aggregates the content (values) of the patches by the different features (key dimensions), thus producing 
a value (column) vector $\mC_{i, \cdot} \in \R^{\dv}$ for each feature map $1 \le i \le \dk$.
For the query, every row of $\softmax_\rightarrow (\mQ) \in \R^{\seqlen \times \dk}$ defines what information is important for the corresponding patch, i.e., how much each feature map contributes to this patch. Finally, the output for a patch is given by the corresponding weighted sum of the values per feature map from the global context~$\mC$.

\subsection{Windowed Attention}

\begin{figure}[t]
    \centering
    \resizebox{0.95\columnwidth}{!}{%
        \input{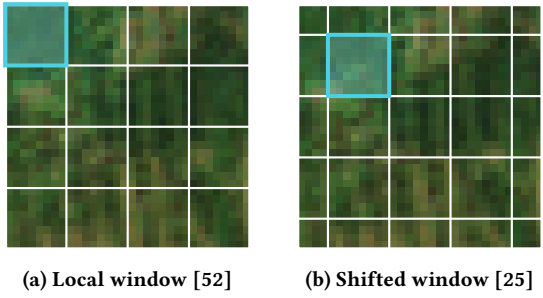}
    }
    \caption{Windowed attention with windows of size $8 \times 8$ pixels. Only pixels inside the same window can attend to each other. An example window is highlighted in cyan.}
    \Description{Two images showing the same 32 times 32 pixels satellite image, once with the default window configuration, and once with the shifted window configuration.}
\end{figure}

Window-based attention mechanisms reduce the number of tokens a token attends to by splitting the input into parts.
The simplest variant is the local window attention (see Figure~\ref{fig:local_window}), where attention is performed within non-overlapping windows of size $\windowsize \times \windowsize$ tokens~\cite{Yuan2021HRFormer}, $\windowsize \in \mathbb{N}$. In the context of ViTs, the input $\mX \in \R^{\seqlen \times \dmodel}$ originates from an image of height $\height'$ and width $\width'$ of embedding vectors with dimension $\dmodel$, i.e. $\mX \in \R^{\height' \times \width' \times \dmodel}$, where $\seqlen = \height ' \cdot \width'$. $\mX$ is then---after zero-padding if necessary---divided into windows of size $\windowsize \times \windowsize$, formally by reshaping $\mX$ to $\mX_\text{reshape} \in \R^{(\height' \cdot \width') / \windowsize^2 \times \windowsize^2 \times \dmodel}$. Viewing the input as a batch of sequences of length $\windowsize^2$, attention is performed on each window individually. This approach achieves linear memory and computational complexity in the number of patches $\seqlen$.
Note that in windowed attention each pixel can only interact with pixels in the same window. \citet{Liu2021Swin} address this by using shifted windows, giving the Swin Transformer its name. The windows alternate between a regular partitioning (see Figure~\ref{fig:local_window}) and one shifted by half the window size (see Figure~\ref{fig:shifted_window}), ensuring that information can flow across window boundaries.

\subsection{Flash Attention}

Efficient attention and window-based attention achieve a runtime linear in the sequence length by approximating the attention mechanism. In contrast, FlashAttention~\cite{Dao2022FlashAttention}, optimizes the exact calculation of attention on the hardware-side by tiling, which reduces the need of copying data between GPU high bandwidth memory and GPU on-chip SRAM.
In an extension, FlashAttention-2~\cite{Dao2023FlashAttention2}, the work partitioning between different thread blocks and warps on the GPU is identified as a remaining bottleneck, which achieves a $2 \times$ speedup compared to FlashAttention.
Later extensions improve performance specifically for Hopper GPUs~\cite{Shah2024FlashAttention3} and Blackwell GPUs~\cite{Zadouri2026FlashAttention4}.

\section{Model Architecture}\label{sec:model_arch}
\begin{figure*}
    \centering
    \resizebox{0.85\textwidth}{!}{
        \tikzset{shape_annotation_box/.append style={fill=none}}
\begin{tikzpicture}

    \node[block, fill=decoder_color] (dec_4) at (0,0) {Block 4 (dec$_4$)};
    \node[shape_annotation_box, below=\ssp of dec_4] (dec_4_input) {$\{ \textcolor{encoder_color}{\mY_{\text{enc}_1}}, \textcolor{decoder_color}{\mY_{\text{dec}_3}}, \textcolor{decoder_color}{\mY_{\text{dec}_2}}, \textcolor{decoder_color}{\mY_{\text{dec}_1}} \} $};
    \node[shape_annotation_box, right=\ssp of dec_4] (dec_4_output) {$\textcolor{decoder_color}{\mY_{\text{dec}_4}}$};
    \draw[main arrow] (dec_4_input) -- (dec_4);
    \draw[main] (dec_4) -- (dec_4_output);

    \node[block, fill=decoder_color, below=\msp of dec_4_input] (dec_3) {Block 3 (dec$_3$)};
    \node[shape_annotation_box, below=\ssp of dec_3] (dec_3_input) {$\{ \textcolor{encoder_color}{\mY_{\text{enc}_1}}, \textcolor{encoder_color}{\mY_{\text{enc}_2}}, \textcolor{decoder_color}{\mY_{\text{dec}_2}}, \textcolor{decoder_color}{\mY_{\text{dec}_1}} \} $};
    \node[shape_annotation_box, right=\ssp of dec_3] (dec_3_output) {$\textcolor{decoder_color}{\mY_{\text{dec}_3}}$};
    \draw[main arrow] (dec_3_input) -- (dec_3);
    \draw[main] (dec_3) -- (dec_3_output);

    \node[block, rotate=90, right=\ssp of dec_3_output, minimum width=1.6cm, anchor=north] (upsample_3) {$\upsample$ $2 \times \uparrow$};
    \draw[main arrow] (dec_3_output) -- (upsample_3);

    \node[block, fill=decoder_color, below=\ssp of dec_3_input] (dec_2) {Block 2 (dec$_2$)};
    \node[shape_annotation_box, below=\ssp of dec_2] (dec_2_input) {$\{ \textcolor{encoder_color}{\mY_{\text{enc}_1}}, \textcolor{encoder_color}{\mY_{\text{enc}_2}}, \textcolor{encoder_color}{\mY_{\text{enc}_3}}, \textcolor{decoder_color}{\mY_{\text{dec}_1}} \} $};
    \node[shape_annotation_box, right=\ssp of dec_2] (dec_2_output) {$\textcolor{decoder_color}{\mY_{\text{dec}_2}}$};
    \draw[main arrow] (dec_2_input) -- (dec_2);
    \draw[main] (dec_2) -- (dec_2_output);

    \node[block, rotate=90, right=\ssp of dec_2_output, minimum width=1.6cm, anchor=north] (upsample_2) {$\upsample$ $4 \times \uparrow$};
    \draw[main arrow] (dec_2_output) -- (upsample_2);

    \node[block, fill=decoder_color, below=\ssp of dec_2_input] (dec_1) {Block 1 (dec$_1$)};
    \node[shape_annotation_box, below=\ssp of dec_1] (dec_1_input) {$\{ \textcolor{encoder_color}{\mY_{\text{enc}_1}}, \textcolor{encoder_color}{\mY_{\text{enc}_2}}, \textcolor{encoder_color}{\mY_{\text{enc}_3}}, \textcolor{encoder_color}{\mY_{\text{enc}_4}} \} $};
    \node[shape_annotation_box, right=\ssp of dec_1] (dec_1_output) {$\textcolor{decoder_color}{\mY_{\text{dec}_1}}$};
    \draw[main arrow] (dec_1_input) -- (dec_1);
    \draw[main] (dec_1) -- (dec_1_output);

    \node[block, rotate=90, right=\ssp of dec_1_output, minimum width=1.6cm, anchor=north] (upsample_1) {$\upsample$ $8 \times \uparrow$};
    \draw[main arrow] (dec_1_output) -- (upsample_1);

    \path let \p1=(dec_4.north), \p2=(dec_1_input.south), \p3=(upsample_1.south) in
      node[block, rotate=90, minimum width={\y1-\y2}, anchor=north] 
      at (\x3+\ssp cm, {(\y1+\y2)/2}) 
      (concat) {$\concat$};

    \node[ffn block, rotate=90, right=\ssp of concat.south, anchor=north] (ffn) {$\ffn$};

    \coordinate[right=\ssp of ffn.south] (dec_out);

    \begin{scope}[on background layer]
        \node[bgbox, draw=decoder_color, fill=decoder_color!20, fit={(dec_4) (dec_1_input) (ffn)}, label=below:{\textcolor{decoder_color}{Decoder}}] (block_dec) {};
    \end{scope}

    \draw[main arrow] (dec_4_output.east) -- (concat.north |- dec_4_output.east);
    \draw[main arrow] (upsample_3.south) -- (concat.north |- upsample_3.south);
    \draw[main arrow] (upsample_2.south) -- (concat.north |- upsample_2.south);
    \draw[main arrow] (upsample_1.south) -- (concat.north |- upsample_1.south);
    \draw[main arrow] (concat.south) -- (ffn.north);
    \draw[main arrow] (ffn.south) -- (dec_out);

    \node[block, fill=encoder_color, left=2*\bsp of dec_4] (enc_1) {Stage 1 (enc$_1$)};
    \node[block, fill=encoder_color, left=2*\bsp of dec_3] (enc_2) {Stage 2 (enc$_2$)};
    \node[block, fill=encoder_color, left=2*\bsp of dec_2] (enc_3) {Stage 3 (enc$_3$)};
    \node[block, fill=encoder_color, left=2*\bsp of dec_1] (enc_4) {Stage 4 (enc$_3$)};

    \begin{scope}[on background layer]
        \node[bgbox, draw=encoder_color, fill=encoder_color!20, fit={(enc_1) (enc_4)}, label=below:{\textcolor{encoder_color}{Encoder}}] (block_enc) {};
    \end{scope}

    \node[shape_annotation_box] at ($ (enc_1 -| block_enc.east)!0.5!(dec_4 -| block_dec.west) $) (enc_1_output) {$\textcolor{encoder_color}{\mY_{\text{enc}_1}}$};
    \node[shape_annotation_box] at ($ (enc_2 -| block_enc.east)!0.5!(dec_3 -| block_dec.west) $) (enc_2_output) {$\textcolor{encoder_color}{\mY_{\text{enc}_2}}$};
    \node[shape_annotation_box] at ($ (enc_3 -| block_enc.east)!0.5!(dec_2 -| block_dec.west) $) (enc_3_output) {$\textcolor{encoder_color}{\mY_{\text{enc}_3}}$};
    \node[shape_annotation_box] at ($ (enc_4 -| block_enc.east)!0.5!(dec_1 -| block_dec.west) $) (enc_4_output) {$\textcolor{encoder_color}{\mY_{\text{enc}_4}}$};

    \coordinate[above=\ssp of enc_1] (input_enc_1);

    \draw[main arrow] (input_enc_1) -- (enc_1.north);
    \draw[main arrow] (enc_1.south) -- (enc_2.north);
    \draw[main arrow] (enc_2.south) -- (enc_3.north);
    \draw[main arrow] (enc_3.south) -- (enc_4.north);
    \draw[main] (enc_1.east) -- (enc_1_output.west);
    \draw[main] (enc_2.east) -- (enc_2_output.west);
    \draw[main] (enc_3.east) -- (enc_3_output.west);
    \draw[main] (enc_4.east) -- (enc_4_output.west);
    \draw[main arrow] (enc_1_output.east) -- (dec_4.west);
    \draw[main arrow] (enc_2_output.east) -- (dec_3.west);
    \draw[main arrow] (enc_3_output.east) -- (dec_2.west);
    \draw[main arrow] (enc_4_output.east) -- (dec_1.west);

    \path let \p1=(block_enc.north), \p2=(block_enc.west) in
      coordinate (stage_origin) at (\x2 - 2.5*\bsp cm, \y1);
    \coordinate (input_stage) at (stage_origin |- input_enc_1);
    \node[block, below=\ssp of input_stage] (patch_embed) {$\patchembed$};
    \node[norm block, below=2*\ssp of patch_embed] (ln_1) {$\layernorm$};
    \node[mha block, below=\msp of ln_1] (msa) {$\mha$ ($\eff$)};
    \node[add, below=\ssp of msa] (add_1) {$+$};
    \node[norm block, below=\msp of add_1] (ln_2) {$\layernorm$};
    \node[ffn block, below=\ssp of ln_2] (ffn_stage) {$\mixffn$};
    \node[add, below=\ssp of ffn_stage] (add_2) {$+$};
    \coordinate[below=\ssp of add_2] (output_stage);

    \draw[main arrow] (input_stage) -- (patch_embed);
    \draw[main arrow] (patch_embed) -- (ln_1);
    \draw[main arrow] (ln_1) -- (msa);
    \draw[main arrow] (msa) -- (add_1);
    \draw[main arrow] (add_1) -- (ln_2);
    \draw[main arrow] (ln_2) -- (ffn_stage);
    \draw[main arrow] (ffn_stage) -- (add_2);
    \draw[main arrow] (add_2) -- (output_stage);

    \coordinate[above=\ssp of ln_1] (help_residual_1);
    \path let 
        \p1 = ($(ln_1.south)!0.5!(add_1)$),
        \p2 = ($(msa.west)-(0.5*\ssp,0)$)
            in coordinate (help_residual_2) at (\x2,\y1);
    \draw[main] (help_residual_1) -| (help_residual_2);
    \draw[main arrow] (help_residual_2) |- (add_1.west);

    \coordinate (help_residual_3) at ($(add_1.south)!0.4!(ln_2.north)$);
    \path let 
        \p1 = ($(help_residual_2)!0.5!(add_2)$),
        \p2 = ($(ffn_stage.west)-(0.5*\ssp,0)$)
            in coordinate (help_residual_4) at (\x2,\y1);
    \draw[main] (help_residual_3) -| (help_residual_4);
    \draw[main arrow] (help_residual_4) |- (add_2.west);

    \coordinate (help_5) at ($(ln_1.south)!0.4!(msa.north)$);
    \draw[main arrow] (help_5) -| ($(msa.north)+(\mhaoffset,0)$);
    \draw[main arrow] (help_5) -| ($(msa.north)+(-\mhaoffset,0)$);

    \begin{scope}[on background layer]
        \node[bgbox, draw=encoder_color, fill=encoder_color!20, fit={(help_residual_1) (help_residual_2) (add_2) (ln_1)}] (block_stage) {};
    \end{scope}

    \node[text=encoder_color, left=0.1cm of block_stage.west, anchor=east, inner sep=0pt, outer sep=0pt] (enc_block_annotation) {$2\times$};

    \node[align=center, text=encoder_color, below=0cm of output_stage, anchor=north] (stage_i_label) {\textcolor{encoder_color}{Stage $i$}};

    \path let 
      \p1 = (block_stage.east),
      \p2 = (block_enc.west),
      \p3 = (block_enc.north),
      \p4 = (block_stage.south)
      in coordinate (midline_top) at ({(\x1+\x2)/2},\y3)
         coordinate (midline_bottom) at ({(\x1+\x2)/2},\y4);

    \draw[dashed, line width=0.7pt, color=gray] (midline_top) -- (midline_bottom);

\end{tikzpicture}
    }
    \caption{Model architecture: The encoder consists of four stages, each of which starts with a $\patchembed$ layer followed by two Transformer blocks, which use efficient attention and $\mixffn$. The decoder consists of four blocks, which perform efficient mix-attention. The encoder and decoder features' shapes are $\mY_{\text{enc}_i} = \mY_{\text{dec}_{4-i+1}} = \frac \height {\patchsize \cdot 2^{i-1}} \times \frac \width {\patchsize \cdot 2^{i-1}} \times 2^{i-1}\dmodel$ (e.g., for $\patchsize=1$ after encoder stage 1 and decoder block 4 we have $\height \times \width \times \dmodel$). After the decoder features' spatial dimension is aligned via bilinear upsampling, the features are concatenated and post-processed via a simple $\ffn$ to generate the prediction.}
    \label{fig:model_architecture}
    \Description{All relevant information is given in the caption and text.}
\end{figure*}
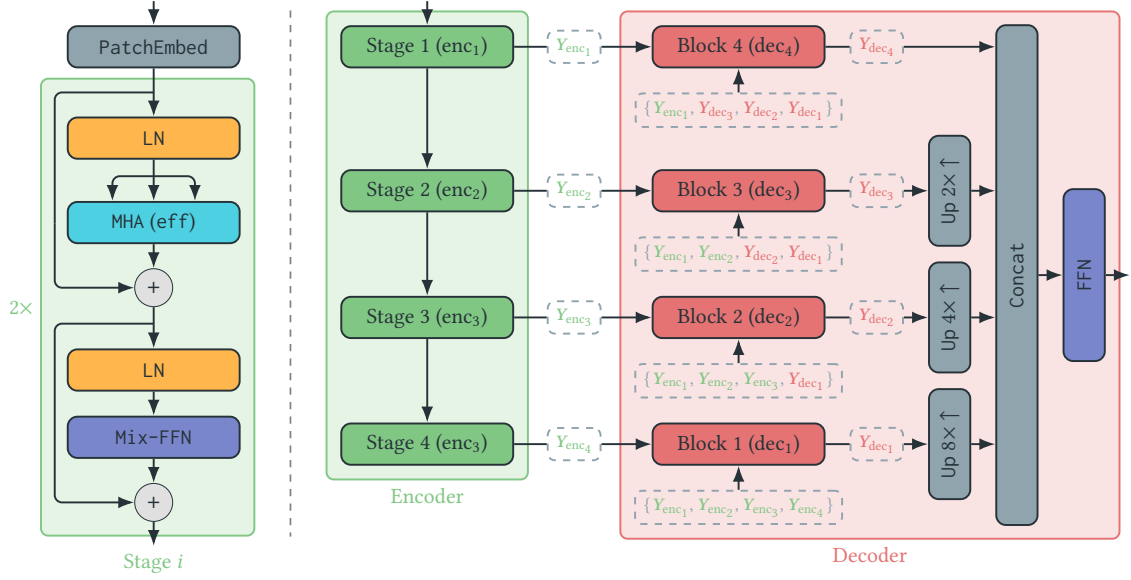

Our model architecture follows the encoder-decoder design, using an adapted version of the Mix Transformer~(MiT)~\cite{Xie2021SegFormer} as encoder and U-MixFormer~\cite{Yeom2025UMixFormer} as decoder. Figure~\ref{fig:model_architecture} shows the adapted architecture.
We first revisit the original architectures and then detail our adaptations and the reasoning behind them.

\subsection{Mix Transformer}\label{sec:mit}

The MiT was introduced as the encoder of the SegFormer~\cite{Xie2021SegFormer}, where it was combined with a lightweight MLP decoder, and ach\-ieved state-of-the-art performance for semantic segmentation in terms of efficiency and accuracy.
The MiT's design is inspired by the Pyramid Vision Transformer (PVT)~\cite{Wang2021PVT}, and thus also has four stages, which create multi-scale image features~\cite{Xie2021SegFormer}. It utilizes the PVT's spatial reduction attention~\cite{Wang2021PVT}, and performs a patch embedding at the beginning of every stage~\cite{Xie2021SegFormer}. 

The MiT architecture is based on two main innovations: the $\mixffn$ and an overlapped patch embedding ($\patchembed$)~\cite{Xie2021SegFormer}. The $\mixffn$ inserts a depth-wise convolution ($\dwconv$) layer with a kernel size of $3$ and padding and stride of $1$ in between the linear layers of the $\ffn$, which can be formalized as:
$$\mY_\mixffn = \linear(\gelu(\dwconv(\linear(\mX_\mixffn)))) + \mX_\mixffn$$
While this adaptation goes against the initial idea of the $\ffn$ to post-process every token individually, it has the merit that---as the tokens interact with their surrounding tokens---no positional encoding is required~\cite{Xie2021SegFormer}.
The $\patchembed$ layer has an increased kernel size to include information from the surrounding patches and enhance local continuity. For the initial patch embedding, a $\conv$ layer with a kernel size $7$, stride $4$, and padding $3$ is used, which divides the image into patches of size $4 \times 4$. For the subsequent ones, a $\conv$ layer with a kernel size $3$, stride $2$, and padding $1$ is used, halving the image size with every recurring embedding.

\subsection{U-MixFormer}

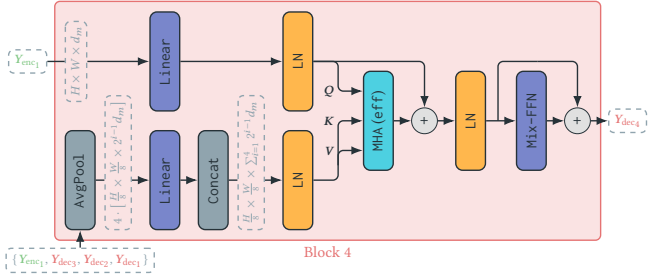
\begin{figure}[t]
    \centering
    \tikzset{block/.append style={minimum width=2cm}}
\tikzset{shape_annotation_box/.append style={fill=none}}
\resizebox{0.48\textwidth}{!}{%
\begin{tikzpicture}

	\node[block, fill=mha_color, rotate=90] at (0,0) (mix_mha) {$\mha$ ($\eff$)};
	\node[add, right=\ssp of mix_mha.south, anchor=north, rotate=90] (add_1) {\small $+$};
	\node[norm block, right=\ssp of add_1.south, anchor=north, rotate=90] (ln_3) {$\layernorm$};
	\node[ffn block, right=\msp of ln_3.south, anchor=north, rotate=90] (mix_ffn) {$\mixffn$};
	\node[add, right=\ssp of mix_ffn.south, anchor=north, rotate=90] (add_2) {\small $+$};
    \node[shape_annotation_box, right=\ssp of add_2.south] (dec_4_output) {$\textcolor{decoder_color}{\mY_{\text{dec}_4}}$};

    \draw[main arrow] (mix_mha.south) -- (add_1.north);
    \draw[main arrow] (add_1.south) -- (ln_3.north);
    \draw[main arrow] (ln_3.south) -- (mix_ffn.north);
    \draw[main arrow] (mix_ffn.south) -- (add_2.north);
    \draw[main arrow] (add_2.south) -- (dec_4_output.west);

	\node[norm block, left=\bsp of mix_mha.north, anchor=south, rotate=90, xshift=1.2cm] (ln_1) {$\layernorm$};

    \draw[main arrow] (ln_1.south) -| (add_1.east);

	\coordinate (dec_help_5) at ($(ln_3.south)!0.4!(mix_ffn.north)$);
    \coordinate (dec_help_6) at ($(add_2)!0.5!(dec_help_5)$);
    \coordinate (dec_help_7) at (dec_help_6 |- ln_1.south);
    \draw[main] (dec_help_5) |- (dec_help_7);
    \draw[main arrow] (dec_help_7) -| (add_2.east);

	\node[norm block, left=\bsp of mix_mha.north, anchor=south, rotate=90, xshift=-1.2cm] (ln_2) {$\layernorm$};
    \node[shape_annotation_box, rotate=90, left=\ssp of ln_2.north west, anchor=south west] (concat_shape) {$\frac \height 8 \times \frac \width 8 \times \sum_{i=1}^4 2^{i-1}\dmodel$};
	\node[block, left=0.5*\ssp of concat_shape.north west, anchor=south west, rotate=90] (concat) {$\concat$};
    \node[ffn block, left=\ssp of concat.north, anchor=south, rotate=90] (linear) {$\linear$};
    \node[shape_annotation_box, rotate=90, left=\ssp of linear.north west, anchor=south west] (avgpool_shape) {$4 \cdot \left[ \frac \height 8 \times \frac \width 8 \times 2^{i-1}\dmodel \right]$};
	\node[block, left=0.5*\ssp of avgpool_shape.north west, anchor=south west, rotate=90] (avgpool) {$\avgpool$};
    \path let \p1 = ($(avgpool.north)-(\ssp, 0)$), \p2 = (ln_1.north) in
        coordinate (dec_4_q_coord) at (\x1,\y2);
    \node[shape_annotation_box, anchor=east] (dec_4_q) at (dec_4_q_coord)
        {$\textcolor{encoder_color}{\mY_{\text{enc}_1}}$};
    \node[shape_annotation_box, rotate=90, right=\ssp of dec_4_q.east, anchor=north] (dec_4_q_shape) {$\height \times \width \times \dmodel$};

    \node[ffn block, rotate=90, anchor=center] (linear_q) at (ln_1.north -| linear.center) {$\linear$};

    \node[shape_annotation_box, below=\ssp of avgpool.west] (dec_4_kv) {$\{ \textcolor{encoder_color}{\mY_{\text{enc}_1}}, \textcolor{decoder_color}{\mY_{\text{dec}_3}}, \textcolor{decoder_color}{\mY_{\text{dec}_2}}, \textcolor{decoder_color}{\mY_{\text{dec}_1}} \} $};

    \draw[main] (dec_4_q.east) -- (dec_4_q_shape.north);
    \draw[main arrow] (dec_4_q_shape.south) -- (linear_q.north);
    \draw[main arrow] (linear_q.south) -- (ln_1.north);
    \draw[main arrow] (dec_4_kv.north) -- (avgpool.west);
    \draw[main] (avgpool.south) -- (avgpool_shape.north  |- avgpool.south);
    \draw[main arrow] (avgpool_shape.south  |- avgpool.south) -- (linear.north);
    \draw[main arrow] (linear.south) -- (concat.north);
    \draw[main] (concat.south) -- (concat_shape.north |- concat.south);
    \draw[main arrow] (concat_shape.south |- concat.south) -- (ln_2.north);

    \begin{scope}[on background layer]
        \node[bgbox, draw=decoder_color, fill=decoder_color!20, fit={(avgpool) (ln_1) (ln_2) (add_2)}, label=below:{\textcolor{decoder_color}{Block 4}}] (block_dec) {};
    \end{scope}

    \coordinate (mix_mha_target_q) at ([yshift=\mhaoffset cm]mix_mha.north);
    \coordinate (mix_mha_target_v) at ([yshift=-\mhaoffset cm]mix_mha.north);
    \coordinate (help_q) at ($(ln_1.south)!0.5!(mix_mha_target_q)$);
    \coordinate (help_kv) at ($(ln_2.south)!0.5!(mix_mha_target_v)$);

    \draw[main arrow] (ln_1.south) -| (help_q) |- (mix_mha_target_q);
    \draw[main arrow] (ln_2.south) -| (help_kv) |- (mix_mha_target_v);
    \draw[main arrow] (help_kv) |- (mix_mha.north);

    \node[anchor=east, inner sep=2pt, font=\footnotesize] (label_k) at ($(mix_mha.north)+(-0.5*\bsp, \mhaoffset)$) {$\mQ$};
    \node[anchor=east, inner sep=2pt, font=\footnotesize] (label_k) at ($(mix_mha.north)+(-0.5*\bsp, 0)$) {$\mK$};
    \node[anchor=east, inner sep=2pt, font=\footnotesize] (label_k) at ($(mix_mha.north)+(-0.5*\bsp, -\mhaoffset)$) {$\mV$};

\end{tikzpicture}
}
    \vspace{-\baselineskip}
    \caption{Architecture of the decoder's block 4 of a model with patch size $\patchsize = 1$. The other blocks have a similar structure, but operate on tensors of different shapes. The encoder feature is the query, while the encoder-decoder feature pyramid provides the keys and values.}
    \Description{The decoder block differs from standard, self-attention, Transformer blocks in the following way: Input for the keys and values are not the same (encoder feature of the corresponding layer), but the feature pyramid of available encoder-decoder features. This feature pyramid is projected to the keys and values by first applying average pooling, then a linear layer, and then concatenating them.}
    \label{fig:umix_block}
\end{figure}

The U-MixFormer decoder also consists of four stages, each of which comprises a single Transformer block~\cite{Yeom2025UMixFormer}.
Instead of adding or concatenating the skipped encoder features and decoder features, they introduce the so-called mix-attention module to fuse them, replacing the self-attention mechanism~\cite{Yeom2025UMixFormer}. Figure~\ref{fig:umix_block} shows the structure of a Transformer block with mix-attention.
Self-attention projects the queries, keys and values from a single input, while cross-attention projects the queries from one input and the keys and values from another one~\cite{Vaswani17}.
For mix-attention, a single-scale feature is the input for the queries, while a second, multi-scale feature map is the input for the keys and values~\cite{Yeom2025UMixFormer}.
Note that the distinction between self-, cross-, and mix-attention is independent of the attention mechanism used.

The query for the decoder block with index $4-i+1$ with $i \in \{1, \dots, 4\}$ is obtained by linearly projecting the output of the corresponding encoder block $i$, i.e., $\mQ_{\text{dec}_{4-i+1}} = \linear(\mY_{\text{enc}_i})$.
The keys and values for the same decoder block are derived from the hierarchical feature map constructed from all encoder outputs, where the encoder feature $\mY_{\text{enc}_i}$ is replaced by its decoder feature of same dimensions $\mY_{\text{dec}_{4-i+1}}$, if it is available.
Thus, for the first decoder block, $\mK_{\text{dec}_1}$ and $\mV_{\text{dec}_1}$ are derived from all the encoder features, $\{ \mY_{\text{enc}_1}, \dots, \mY_{\text{enc}_4} \}$, as no decoder features are available yet, while for the last decoder block, $\text{dec}_4$, the three lower-dimensional encoder feature maps have already been created and $\mK_{\text{dec}_4}$ and $\mV_{\text{dec}_4}$ are derived from $\{ \mY_{\text{enc}_1}, \mY_{\text{dec}_3}, \mY_{\text{dec}_2}, \mY_{\text{dec}_1} \}$.

The transformation of the four features (of different resolution) into the keys and values works as follows: An 
$\avgpool$ layer unifies the spatial dimensions by pooling all feature maps to the resolution of the lowest encoder block ($\frac \height {32} \times \frac \width {32}$ for the original setting of $\patchsize=4$ and $\frac \height 8 \times \frac \width 8$ when setting $\patchsize=1$). Then a $\linear$ layer postprocesses the channel dimension and a $\concat$ layer concatenates all feature maps. Subsequently, the spatial dimensions are flattened, and the channel dimension split into two (for the individual projections for keys and values)~\cite{Yeom2025UMixFormer}.
After computing the $\mha$, the tokens are postprocessed via a $\mixffn$, as introduced by the SegFormer's MiT (see Section~\ref{sec:mit})~\cite{Xie2021SegFormer,Yeom2025UMixFormer}.

\subsection{Architectural Adaptations}

We adapt the original implementations in two minor yet crucial ways.
In our experiments, we vary the patch size $\patchsize \in \{1, 2, 4, 8\}$ to study its impact. To achieve this, we implement the first overlapped patch embedding via a $\conv$ layer with a kernel size of $2\patchsize-1$, stride of $\patchsize$, and padding of $\patchsize-1$ (subsequent patch embedding layers stay unchanged). A bigger patch size would result in an output with a proportionally smaller spatial resolution. To counteract this effect, we restore the original resolution by adding \texttt{PatchExpand} layers in front of the final linear layer, which creates the prediction. 
The \texttt{PatchExpand} layer was introduced by \citet{Cao2022SwinUnet} to increase a tensor's spatial resolution. It first expands the embedding dimension by a factor of two, then reshapes the tensor to
twice the spatial dimension and half the original embedding dimension.
This approach of parametric upsampling is common practice, and superior to the non-parametric bilinear upsampling of the predictions~\cite{Cao2022SwinUnet}.
We further substitute all attention blocks across the model with efficient attention~\cite{Shen2021Efficient} by default and experiment with other attention variants described in Section~\ref{sec:results_attention}.

\section{Experiments}\label{sec:experiments}
Followingly, we empirically study pixel-level ViTs for canopy height prediction w.r.t. prediction quality and computational efficiency. We first describe two datasets used for training and evaluation, and then analyze the impact of patch size, model dimension, and attention mechanism. Finally, we benchmark the resulting model against established dense prediction architectures.

\subsection{Datasets}

We consider two datasets for experimental evaluation: The \texttt{Europe} dataset is based on~\citet{Pauls2025CapturingTemporalDynamics} and the \texttt{France} dataset contains labels from the OpenCanopy dataset by~\citet{Fogel2025OpenCanopy}.
Both datasets have Sentinel-2 images as input, which consist of twelve channels and have a shape of $256 \times 256$ pixels. We take median composites across multiple months to reduce the prevalence of clouds. All channels are clipped to a per-channel reflectance range before being min-max normalized to a range of zero to one~\cite{Pauls2025CapturingTemporalDynamics}.

The \texttt{Europe} dataset consists of about $395\,000$ patches, which were split into a training ($95\%$) and validation set ($5\%$).
It has a temporal coverage from 2019 to 2023.
The GEDI labels follow a right-skewed distribution with the $5\%$ quantile at \SI{2.4}{m},\footnote{Due to the measurement process and preprocessing to obtain the labels, GEDI labels indicate a height of approximately \SI{2.5}{m} for grassland.} the median at \SI{3.7}{m}, and the $95\%$ quantile at \SI{28.2}{m}. This dataset does not have high-resolution ALS labels.

The \texttt{France} dataset combines the ALS labels provided by the OpenCanopy dataset~\cite{Fogel2025OpenCanopy} with the available GEDI labels from the same year and region.
The ALS labels are available on a resolution of \SI{1.5}{m}, which we resampled to \SI{10}{m} by taking the maximum to match the resolution of our Sentinel-2 inputs.
Using the large tiles from OpenCanopy, we use a sliding window approach to create non-overlapping cutouts of size $256 \times 256$ pixels (to match the resolution of the \texttt{Europe} dataset).
The GEDI labels have the~$5\%$ quantile at~\SI{2.5}{m}, the median at~\SI{10.4}{m} and the~$95\%$ quantile at~\SI{35.7}{m}, while the quantiles of the ALS labels are at \SI{0.1}{m}, \SI{10.9}{m}, and~\SI{30.2}{m}.
We train and evaluate our models on the \texttt{Europe} dataset. For the final evaluation, we consider the \texttt{France} dataset, ensuring that none of its patches overlap with those used during training.
The geographical extent of the \texttt{Europe} dataset's training and validation patches, as well as the \texttt{France} dataset's evaluation patches, is shown in Figure~\ref{fig:dataset_extent} in the appendix.

\subsection{Evaluation Metrics}

Apart from perceptual (i.e. qualitative) evaluation, we quantitatively evaluate the models on validation datasets along two dimensions: predictive accuracy and computational cost. We obtain robust estimates by training each model configuration five times with different seeds. To assess predictive accuracy, we use two metrics. First, we compute the mean absolute error over all pixels with a label greater than $\SI{5}{m}$ ($MAE_{>5\text{m}} \in [0, \infty)$, lower is better). This threshold mitigates the influence of grassland pixels, which dominate the dataset, focusing on pixels containing trees. Second, we report the coefficient of determination ($R^2 \in (-\infty, 1]$, higher is better), which quantifies the variance reduction achieved by the model relative to a constant mean prediction. We also indicate the label type (GEDI or ALS) on which the metric is computed.
Computational cost is assessed by measuring the runtime of a single training step on a mini-batch containing one image (forward and backward pass), along with its peak VRAM usage.
For perceptual evaluation, we show predictions of the model with the median $MAE_{>5\text{m}}$ (ALS).

\subsection{Results}

We first investigate the impact of varying patch sizes and model dimensions, fixing the attention mechanism to efficient attention.
Then, we explore the impact of different attention mechanisms, fixing the patch size and model dimension to the optimal values determined by the first experiment.
Finally, we compare the best-performing configuration with established baseline architectures commonly used for canopy height prediction, depth estimation, and semantic segmentation more broadly. All models, except for the baselines, follow the architecture described in Section~\ref{sec:model_arch}. The number of attention heads doubles in every stage from $1$ to $8$. The decoder dimension is fixed to $96$.

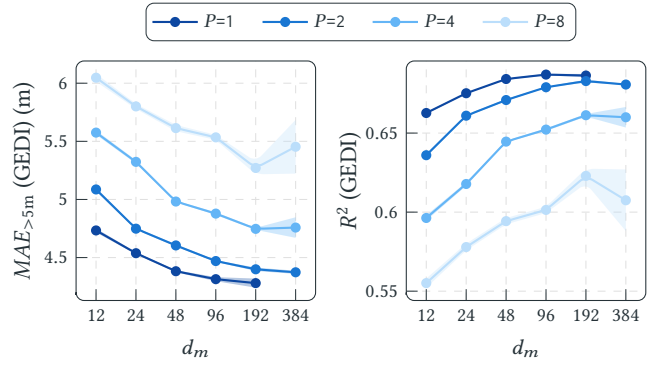
\begin{figure}[t]
    \centering
    \
\usepgfplotslibrary{fillbetween}
\usetikzlibrary{calc}
\begin{tikzpicture}
\begin{groupplot}[
    group style={group size=2 by 1, horizontal sep=1.2cm},
]
\nextgroupplot[
    width=4.75cm, height=4.75cm,
    xmode=log, log basis x=2, log ticks with fixed point,
    xtick={12,24,48,96,192,384},
    xlabel={$\dmodel$},
    ymin=4.1308,
    ymax=6.1911,
    ymajorgrids=true, grid style={line width=0.3pt, draw=gray!25},
    axis line style={bluegrey900}, tick style={bluegrey900},
    tick label style={font=\footnotesize},
    label style={font=\small},
    every axis plot/.append style={line width=0.8pt},
    ylabel={$MAE_{>5\text{m}}$ (GEDI) (m)},
    ylabel near ticks,
    ylabel style={yshift=-3pt},
    legend to name=generatepatchsizepredictionaccuracylegend, legend columns=-1,
    legend style={font=\footnotesize, draw=bluegrey900, line width=0.3pt, fill=white, inner xsep=2pt, inner ysep=2pt, column sep=0pt, /tikz/every even column/.append style={column sep=8pt}},
]
\addplot[draw=none, forget plot, name path=lop0s0] coordinates {(12,4.7160) (24,4.5289) (48,4.3735) (96,4.2968) (192,4.2411)};
\addplot[draw=none, forget plot, name path=hip0s0] coordinates {(12,4.7508) (24,4.5463) (48,4.3916) (96,4.3318) (192,4.3195)};
\addplot[fill=blue900, opacity=0.3, forget plot] fill between[of=lop0s0 and hip0s0];
\addplot[color=blue900, solid, mark=*, mark size=1.5pt, mark options={solid, fill=blue900, draw=blue900}] coordinates {(12,4.7334) (24,4.5376) (48,4.3826) (96,4.3143) (192,4.2803)};
\addlegendentry{\texttt{$\patchsize$=1}}
\addplot[draw=none, forget plot, name path=lop0s1] coordinates {(12,5.0702) (24,4.7402) (48,4.5947) (96,4.4629) (192,4.3847) (384,4.3597)};
\addplot[draw=none, forget plot, name path=hip0s1] coordinates {(12,5.1018) (24,4.7570) (48,4.6142) (96,4.4775) (192,4.4137) (384,4.3872)};
\addplot[fill=blue700, opacity=0.3, forget plot] fill between[of=lop0s1 and hip0s1];
\addplot[color=blue700, solid, mark=*, mark size=1.5pt, mark options={solid, fill=blue700, draw=blue700}] coordinates {(12,5.0860) (24,4.7486) (48,4.6044) (96,4.4702) (192,4.3992) (384,4.3735)};
\addlegendentry{\texttt{$\patchsize$=2}}
\addplot[draw=none, forget plot, name path=lop0s2] coordinates {(12,5.5565) (24,5.3088) (48,4.9751) (96,4.8665) (192,4.7382) (384,4.6697)};
\addplot[draw=none, forget plot, name path=hip0s2] coordinates {(12,5.5937) (24,5.3381) (48,4.9891) (96,4.8902) (192,4.7560) (384,4.8467)};
\addplot[fill=blue300, opacity=0.3, forget plot] fill between[of=lop0s2 and hip0s2];
\addplot[color=blue300, solid, mark=*, mark size=1.5pt, mark options={solid, fill=blue300, draw=blue300}] coordinates {(12,5.5751) (24,5.3234) (48,4.9821) (96,4.8784) (192,4.7471) (384,4.7582)};
\addlegendentry{\texttt{$\patchsize$=4}}
\addplot[draw=none, forget plot, name path=lop0s3] coordinates {(12,6.0138) (24,5.7786) (48,5.5868) (96,5.5125) (192,5.2162) (384,5.2221)};
\addplot[draw=none, forget plot, name path=hip0s3] coordinates {(12,6.0807) (24,5.8223) (48,5.6407) (96,5.5559) (192,5.3256) (384,5.6847)};
\addplot[fill=blue100, opacity=0.3, forget plot] fill between[of=lop0s3 and hip0s3];
\addplot[color=blue100, solid, mark=*, mark size=1.5pt, mark options={solid, fill=blue100, draw=blue100}] coordinates {(12,6.0473) (24,5.8004) (48,5.6138) (96,5.5342) (192,5.2709) (384,5.4534)};
\addlegendentry{\texttt{$\patchsize$=8}}

\nextgroupplot[
    width=4.75cm, height=4.75cm,
    xmode=log, log basis x=2, log ticks with fixed point,
    xtick={12,24,48,96,192,384},
    xlabel={$\dmodel$},
    ymin=0.5441,
    ymax=0.6956,
    ymajorgrids=true, grid style={line width=0.3pt, draw=gray!25},
    axis line style={bluegrey900}, tick style={bluegrey900},
    tick label style={font=\footnotesize},
    label style={font=\small},
    every axis plot/.append style={line width=0.8pt},
    ylabel={$R^2$ (GEDI)},
    ylabel near ticks,
    ylabel style={yshift=-5pt},
    legend to name=discardgeneratepatchsizepredictionaccuracylegend1,
]
\addplot[draw=none, forget plot, name path=lop1s0] coordinates {(12,0.6620) (24,0.6749) (48,0.6839) (96,0.6865) (192,0.6853)};
\addplot[draw=none, forget plot, name path=hip1s0] coordinates {(12,0.6634) (24,0.6754) (48,0.6845) (96,0.6875) (192,0.6874)};
\addplot[fill=blue900, opacity=0.3, forget plot] fill between[of=lop1s0 and hip1s0];
\addplot[color=blue900, solid, mark=*, mark size=1.5pt, mark options={solid, fill=blue900, draw=blue900}] coordinates {(12,0.6627) (24,0.6752) (48,0.6842) (96,0.6870) (192,0.6864)};
\addlegendentry{\texttt{$\patchsize$=1}}
\addplot[draw=none, forget plot, name path=lop1s1] coordinates {(12,0.6349) (24,0.6605) (48,0.6703) (96,0.6787) (192,0.6820) (384,0.6804)};
\addplot[draw=none, forget plot, name path=hip1s1] coordinates {(12,0.6370) (24,0.6616) (48,0.6715) (96,0.6793) (192,0.6838) (384,0.6811)};
\addplot[fill=blue700, opacity=0.3, forget plot] fill between[of=lop1s1 and hip1s1];
\addplot[color=blue700, solid, mark=*, mark size=1.5pt, mark options={solid, fill=blue700, draw=blue700}] coordinates {(12,0.6360) (24,0.6610) (48,0.6709) (96,0.6790) (192,0.6829) (384,0.6807)};
\addlegendentry{\texttt{$\patchsize$=2}}
\addplot[draw=none, forget plot, name path=lop1s2] coordinates {(12,0.5947) (24,0.6166) (48,0.6442) (96,0.6513) (192,0.6604) (384,0.6535)};
\addplot[draw=none, forget plot, name path=hip1s2] coordinates {(12,0.5980) (24,0.6189) (48,0.6451) (96,0.6530) (192,0.6621) (384,0.6665)};
\addplot[fill=blue300, opacity=0.3, forget plot] fill between[of=lop1s2 and hip1s2];
\addplot[color=blue300, solid, mark=*, mark size=1.5pt, mark options={solid, fill=blue300, draw=blue300}] coordinates {(12,0.5963) (24,0.6178) (48,0.6446) (96,0.6522) (192,0.6613) (384,0.6600)};
\addlegendentry{\texttt{$\patchsize$=4}}
\addplot[draw=none, forget plot, name path=lop1s3] coordinates {(12,0.5522) (24,0.5758) (48,0.5923) (96,0.5994) (192,0.6182) (384,0.5880)};
\addplot[draw=none, forget plot, name path=hip1s3] coordinates {(12,0.5580) (24,0.5801) (48,0.5963) (96,0.6034) (192,0.6275) (384,0.6270)};
\addplot[fill=blue100, opacity=0.3, forget plot] fill between[of=lop1s3 and hip1s3];
\addplot[color=blue100, solid, mark=*, mark size=1.5pt, mark options={solid, fill=blue100, draw=blue100}] coordinates {(12,0.5551) (24,0.5779) (48,0.5943) (96,0.6014) (192,0.6229) (384,0.6075)};
\addlegendentry{\texttt{$\patchsize$=8}}
\end{groupplot}
\node[anchor=south, yshift=0.15cm] at ($(group c1r1.north)!0.5!(group c2r1.north)$)
    {\pgfplotslegendfromname{generatepatchsizepredictionaccuracylegend}};
\end{tikzpicture}
    \vspace{-2\baselineskip}
    \caption{Performance comparison of models with varying patch size ($\patchsize$) and model dimension ($\dmodel$) on the validation subset of the \texttt{Europe} dataset (GEDI labels). Shown are the mean and standard deviation of five training runs per config.}%
    \Description{The undelying trend is that smaller patch sizes and larger model dimensions tend to be better.}%
    \label{fig:patchsize_performance}
\end{figure}

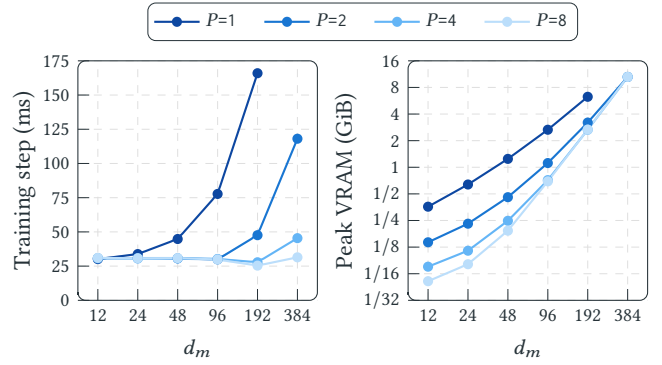
\begin{figure}[t]
    \centering
    \
\usepgfplotslibrary{fillbetween}
\usetikzlibrary{calc}
\begin{tikzpicture}
\begin{groupplot}[
    group style={group size=2 by 1, horizontal sep=1.2cm},
]
\nextgroupplot[
    width=4.75cm, height=4.75cm,
    xmode=log, log basis x=2, log ticks with fixed point,
    xtick={12,24,48,96,192,384},
    xlabel={$\dmodel$},
    enlarge y limits=false,
    ymin=0.0000,
    ymax=175.0000,
    ytick={0,25,50,75,100,125,150,175},
    minor y tick num=0,
    ymajorgrids=true, grid style={line width=0.3pt, draw=gray!25},
    axis line style={bluegrey900}, tick style={bluegrey900},
    tick label style={font=\footnotesize},
    label style={font=\small},
    every axis plot/.append style={line width=0.8pt},
    ylabel={Training step (ms)},
    ylabel near ticks,
    ylabel style={yshift=-4pt},
    legend to name=generatepatchsizeruntimememorylegend, legend columns=-1,
    legend style={font=\footnotesize, draw=bluegrey900, line width=0.3pt, fill=white, inner xsep=2pt, inner ysep=2pt, column sep=0pt, /tikz/every even column/.append style={column sep=8pt}},
]
\addplot[draw=none, forget plot, name path=lop0s0] coordinates {(12,29.2524) (24,33.4912) (48,44.7744) (96,77.5679) (192,165.7503)};
\addplot[draw=none, forget plot, name path=hip0s0] coordinates {(12,31.0230) (24,34.1217) (48,44.8489) (96,77.8995) (192,166.1401)};
\addplot[fill=blue900, opacity=0.3, forget plot] fill between[of=lop0s0 and hip0s0];
\addplot[color=blue900, solid, mark=*, mark size=1.5pt, mark options={solid, fill=blue900, draw=blue900}] coordinates {(12,30.1377) (24,33.8064) (48,44.8116) (96,77.7337) (192,165.9452)};
\addlegendentry{\texttt{$\patchsize$=1}}
\addplot[draw=none, forget plot, name path=lop0s1] coordinates {(12,30.5702) (24,30.7893) (48,29.6815) (96,28.9253) (192,47.4253) (384,117.9841)};
\addplot[draw=none, forget plot, name path=hip0s1] coordinates {(12,30.7794) (24,31.0960) (48,31.5791) (96,30.8899) (192,47.7783) (384,118.1798)};
\addplot[fill=blue700, opacity=0.3, forget plot] fill between[of=lop0s1 and hip0s1];
\addplot[color=blue700, solid, mark=*, mark size=1.5pt, mark options={solid, fill=blue700, draw=blue700}] coordinates {(12,30.6748) (24,30.9426) (48,30.6303) (96,29.9076) (192,47.6018) (384,118.0819)};
\addlegendentry{\texttt{$\patchsize$=2}}
\addplot[draw=none, forget plot, name path=lop0s2] coordinates {(12,30.4541) (24,29.9912) (48,29.9984) (96,29.3274) (192,27.7809) (384,45.2813)};
\addplot[draw=none, forget plot, name path=hip0s2] coordinates {(12,31.1619) (24,31.1088) (48,31.3524) (96,31.1144) (192,27.8518) (384,45.5872)};
\addplot[fill=blue300, opacity=0.3, forget plot] fill between[of=lop0s2 and hip0s2];
\addplot[color=blue300, solid, mark=*, mark size=1.5pt, mark options={solid, fill=blue300, draw=blue300}] coordinates {(12,30.8080) (24,30.5500) (48,30.6754) (96,30.2209) (192,27.8164) (384,45.4342)};
\addlegendentry{\texttt{$\patchsize$=4}}
\addplot[draw=none, forget plot, name path=lop0s3] coordinates {(12,30.4760) (24,29.7805) (48,29.8649) (96,28.6875) (192,25.1446) (384,31.1805)};
\addplot[draw=none, forget plot, name path=hip0s3] coordinates {(12,31.5783) (24,31.7813) (48,31.8406) (96,31.0260) (192,25.5971) (384,31.6620)};
\addplot[fill=blue100, opacity=0.3, forget plot] fill between[of=lop0s3 and hip0s3];
\addplot[color=blue100, solid, mark=*, mark size=1.5pt, mark options={solid, fill=blue100, draw=blue100}] coordinates {(12,31.0272) (24,30.7809) (48,30.8527) (96,29.8567) (192,25.3709) (384,31.4212)};
\addlegendentry{\texttt{$\patchsize$=8}}

\nextgroupplot[
    width=4.75cm, height=4.75cm,
    xmode=log, log basis x=2, log ticks with fixed point,
    xtick={12,24,48,96,192,384},
    xlabel={$\dmodel$},
    ymode=log, log basis y=2, log ticks with fixed point,
    enlarge y limits=false,
    ymin=32.0000,
    ymax=16384.0000,
    ytick={32,64,128,256,512,1024,2048,4096,8192,16384},
    yticklabels={{$1/32$},{$1/16$},{$1/8$},{$1/4$},{$1/2$},{$1$},{$2$},{$4$},{$8$},{$16$}},
    minor y tick num=0,
    ymajorgrids=true, grid style={line width=0.3pt, draw=gray!25},
    axis line style={bluegrey900}, tick style={bluegrey900},
    tick label style={font=\footnotesize},
    label style={font=\small},
    every axis plot/.append style={line width=0.8pt},
    ylabel={Peak VRAM (GiB)},
    ylabel near ticks,
    ylabel style={yshift=-4pt},
    legend to name=discardgeneratepatchsizeruntimememorylegend1,
]
\addplot[draw=none, forget plot, name path=lop1s0] coordinates {(12,366.7544) (24,653.6431) (48,1274.3726) (96,2721.9976) (192,6420.5010)};
\addplot[draw=none, forget plot, name path=hip1s0] coordinates {(12,366.7544) (24,653.6431) (48,1274.3726) (96,2721.9976) (192,6422.5449)};
\addplot[fill=blue900, opacity=0.3, forget plot] fill between[of=lop1s0 and hip1s0];
\addplot[color=blue900, solid, mark=*, mark size=1.5pt, mark options={solid, fill=blue900, draw=blue900}] coordinates {(12,366.7544) (24,653.6431) (48,1274.3726) (96,2721.9976) (192,6421.5229)};
\addlegendentry{\texttt{$\patchsize$=1}}
\addplot[draw=none, forget plot, name path=lop1s1] coordinates {(12,145.1606) (24,235.5220) (48,469.8247) (96,1143.0928) (192,3280.2279) (384,10767.7147)};
\addplot[draw=none, forget plot, name path=hip1s1] coordinates {(12,145.1606) (24,235.5220) (48,469.8247) (96,1143.0928) (192,3281.1485) (384,10767.8244)};
\addplot[fill=blue700, opacity=0.3, forget plot] fill between[of=lop1s1 and hip1s1];
\addplot[color=blue700, solid, mark=*, mark size=1.5pt, mark options={solid, fill=blue700, draw=blue700}] coordinates {(12,145.1606) (24,235.5220) (48,469.8247) (96,1143.0928) (192,3280.6882) (384,10767.7695)};
\addlegendentry{\texttt{$\patchsize$=2}}
\addplot[draw=none, forget plot, name path=lop1s2] coordinates {(12,76.8716) (24,116.5508) (48,254.1001) (96,732.6206) (192,2723.5425) (384,10765.6802)};
\addplot[draw=none, forget plot, name path=hip1s2] coordinates {(12,76.8716) (24,116.5508) (48,254.1001) (96,732.6206) (192,2723.5425) (384,10765.6802)};
\addplot[fill=blue300, opacity=0.3, forget plot] fill between[of=lop1s2 and hip1s2];
\addplot[color=blue300, solid, mark=*, mark size=1.5pt, mark options={solid, fill=blue300, draw=blue300}] coordinates {(12,76.8716) (24,116.5508) (48,254.1001) (96,732.6206) (192,2723.5425) (384,10765.6802)};
\addlegendentry{\texttt{$\patchsize$=4}}
\addplot[draw=none, forget plot, name path=lop1s3] coordinates {(12,52.6699) (24,81.9834) (48,196.4502) (96,709.4897) (192,2734.5010) (384,10781.1201)};
\addplot[draw=none, forget plot, name path=hip1s3] coordinates {(12,52.6699) (24,81.9834) (48,196.4502) (96,709.4897) (192,2734.5010) (384,10781.1201)};
\addplot[fill=blue100, opacity=0.3, forget plot] fill between[of=lop1s3 and hip1s3];
\addplot[color=blue100, solid, mark=*, mark size=1.5pt, mark options={solid, fill=blue100, draw=blue100}] coordinates {(12,52.6699) (24,81.9834) (48,196.4502) (96,709.4897) (192,2734.5010) (384,10781.1201)};
\addlegendentry{\texttt{$\patchsize$=8}}
\end{groupplot}
\node[anchor=south, yshift=0.15cm] at ($(group c1r1.north)!0.5!(group c2r1.north)$)
    {\pgfplotslegendfromname{generatepatchsizeruntimememorylegend}};
\end{tikzpicture}
    \vspace{-2\baselineskip}
    \caption{Efficiency comparison of models with varying patch size ($\patchsize$) and model dimension ($\dmodel$). Shown is the mean of~$1\,000$ training steps on single-image batches across five runs. The standard deviation is imperceptible, due to its small scale.}%
    \Description{The undelying trend is that larger patch sizes and lower model dimensions are better.}%
    \label{fig:patchsize_efficiency}
\end{figure}

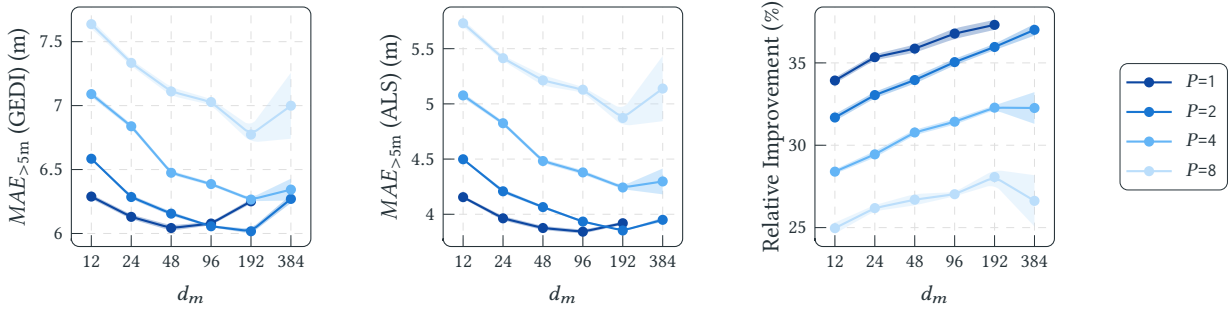
\begin{figure*}
    \centering
    \
\usepgfplotslibrary{fillbetween}
\usetikzlibrary{calc}
\begin{tikzpicture}
\begin{groupplot}[
    group style={group size=3 by 1, horizontal sep=1.75cm},
]
\nextgroupplot[
    width=4.75cm, height=4.75cm,
    xmode=log, log basis x=2, log ticks with fixed point,
    xtick={12,24,48,96,192,384},
    xlabel={$\dmodel$},
    ymin=5.8981,
    ymax=7.7709,
    ymajorgrids=true, grid style={line width=0.3pt, draw=gray!25},
    axis line style={bluegrey900}, tick style={bluegrey900},
    tick label style={font=\footnotesize},
    label style={font=\small},
    every axis plot/.append style={line width=0.8pt},
    ylabel={$MAE_{>5\text{m}}$ (GEDI) (m)},
    ylabel near ticks,
    ylabel style={yshift=-3pt},
    legend to name=generatepatchsizealspredictionaccuracylegend, legend columns=1,
    legend style={font=\footnotesize, draw=bluegrey900, line width=0.3pt, fill=white, inner xsep=2pt, inner ysep=2pt, column sep=3pt, row sep=1pt},
    legend cell align=left,
]
\addplot[draw=none, forget plot, name path=lop0s0] coordinates {(12,6.2708) (24,6.1117) (48,6.0217) (96,6.0639) (192,6.2330)};
\addplot[draw=none, forget plot, name path=hip0s0] coordinates {(12,6.3065) (24,6.1483) (48,6.0637) (96,6.0915) (192,6.2693)};
\addplot[fill=blue900, opacity=0.3, forget plot] fill between[of=lop0s0 and hip0s0];
\addplot[color=blue900, solid, mark=*, mark size=1.5pt, mark options={solid, fill=blue900, draw=blue900}] coordinates {(12,6.2887) (24,6.1300) (48,6.0427) (96,6.0777) (192,6.2511)};
\addlegendentry{\texttt{$\patchsize$=1}}
\addplot[draw=none, forget plot, name path=lop0s1] coordinates {(12,6.5782) (24,6.2697) (48,6.1387) (96,6.0489) (192,5.9984) (384,6.2385)};
\addplot[draw=none, forget plot, name path=hip0s1] coordinates {(12,6.5907) (24,6.3005) (48,6.1718) (96,6.0639) (192,6.0363) (384,6.3022)};
\addplot[fill=blue700, opacity=0.3, forget plot] fill between[of=lop0s1 and hip0s1];
\addplot[color=blue700, solid, mark=*, mark size=1.5pt, mark options={solid, fill=blue700, draw=blue700}] coordinates {(12,6.5845) (24,6.2851) (48,6.1552) (96,6.0564) (192,6.0173) (384,6.2704)};
\addlegendentry{\texttt{$\patchsize$=2}}
\addplot[draw=none, forget plot, name path=lop0s2] coordinates {(12,7.0682) (24,6.8205) (48,6.4591) (96,6.3730) (192,6.2523) (384,6.2578)};
\addplot[draw=none, forget plot, name path=hip0s2] coordinates {(12,7.1120) (24,6.8563) (48,6.4908) (96,6.3997) (192,6.2783) (384,6.4285)};
\addplot[fill=blue300, opacity=0.3, forget plot] fill between[of=lop0s2 and hip0s2];
\addplot[color=blue300, solid, mark=*, mark size=1.5pt, mark options={solid, fill=blue300, draw=blue300}] coordinates {(12,7.0901) (24,6.8384) (48,6.4749) (96,6.3864) (192,6.2653) (384,6.3432)};
\addlegendentry{\texttt{$\patchsize$=4}}
\addplot[draw=none, forget plot, name path=lop0s3] coordinates {(12,7.6031) (24,7.3099) (48,7.0772) (96,7.0041) (192,6.7115) (384,6.7410)};
\addplot[draw=none, forget plot, name path=hip0s3] coordinates {(12,7.6706) (24,7.3582) (48,7.1437) (96,7.0516) (192,6.8357) (384,7.2573)};
\addplot[fill=blue100, opacity=0.3, forget plot] fill between[of=lop0s3 and hip0s3];
\addplot[color=blue100, solid, mark=*, mark size=1.5pt, mark options={solid, fill=blue100, draw=blue100}] coordinates {(12,7.6368) (24,7.3341) (48,7.1105) (96,7.0278) (192,6.7736) (384,6.9991)};
\addlegendentry{\texttt{$\patchsize$=8}}

\nextgroupplot[
    width=4.75cm, height=4.75cm,
    xmode=log, log basis x=2, log ticks with fixed point,
    xtick={12,24,48,96,192,384},
    xlabel={$\dmodel$},
    ymin=3.7079,
    ymax=5.8774,
    ymajorgrids=true, grid style={line width=0.3pt, draw=gray!25},
    axis line style={bluegrey900}, tick style={bluegrey900},
    tick label style={font=\footnotesize},
    label style={font=\small},
    every axis plot/.append style={line width=0.8pt},
    ylabel={$MAE_{>5\text{m}}$ (ALS) (m)},
    ylabel near ticks,
    ylabel style={yshift=-3pt},
    legend to name=discardgeneratepatchsizealspredictionaccuracylegend1,
]
\addplot[draw=none, forget plot, name path=lop1s0] coordinates {(12,4.1413) (24,3.9434) (48,3.8569) (96,3.8241) (192,3.9050)};
\addplot[draw=none, forget plot, name path=hip1s0] coordinates {(12,4.1686) (24,3.9833) (48,3.8941) (96,3.8604) (192,3.9328)};
\addplot[fill=blue900, opacity=0.3, forget plot] fill between[of=lop1s0 and hip1s0];
\addplot[color=blue900, solid, mark=*, mark size=1.5pt, mark options={solid, fill=blue900, draw=blue900}] coordinates {(12,4.1550) (24,3.9634) (48,3.8755) (96,3.8423) (192,3.9189)};
\addlegendentry{\texttt{$\patchsize$=1}}
\addplot[draw=none, forget plot, name path=lop1s1] coordinates {(12,4.4882) (24,4.1901) (48,4.0526) (96,3.9240) (192,3.8498) (384,3.9304)};
\addplot[draw=none, forget plot, name path=hip1s1] coordinates {(12,4.5084) (24,4.2261) (48,4.0774) (96,3.9435) (192,3.8565) (384,3.9693)};
\addplot[fill=blue700, opacity=0.3, forget plot] fill between[of=lop1s1 and hip1s1];
\addplot[color=blue700, solid, mark=*, mark size=1.5pt, mark options={solid, fill=blue700, draw=blue700}] coordinates {(12,4.4983) (24,4.2081) (48,4.0650) (96,3.9338) (192,3.8531) (384,3.9498)};
\addlegendentry{\texttt{$\patchsize$=2}}
\addplot[draw=none, forget plot, name path=lop1s2] coordinates {(12,5.0534) (24,4.8068) (48,4.4632) (96,4.3602) (192,4.2266) (384,4.1804)};
\addplot[draw=none, forget plot, name path=hip1s2] coordinates {(12,5.1001) (24,4.8426) (48,4.5013) (96,4.3984) (192,4.2588) (384,4.4142)};
\addplot[fill=blue300, opacity=0.3, forget plot] fill between[of=lop1s2 and hip1s2];
\addplot[color=blue300, solid, mark=*, mark size=1.5pt, mark options={solid, fill=blue300, draw=blue300}] coordinates {(12,5.0767) (24,4.8247) (48,4.4823) (96,4.3793) (192,4.2427) (384,4.2973)};
\addlegendentry{\texttt{$\patchsize$=4}}
\addplot[draw=none, forget plot, name path=lop1s3] coordinates {(12,5.6990) (24,5.3901) (48,5.1649) (96,5.1033) (192,4.7986) (384,4.8431)};
\addplot[draw=none, forget plot, name path=hip1s3] coordinates {(12,5.7612) (24,5.4372) (48,5.2599) (96,5.1535) (192,4.9462) (384,5.4346)};
\addplot[fill=blue100, opacity=0.3, forget plot] fill between[of=lop1s3 and hip1s3];
\addplot[color=blue100, solid, mark=*, mark size=1.5pt, mark options={solid, fill=blue100, draw=blue100}] coordinates {(12,5.7301) (24,5.4137) (48,5.2124) (96,5.1284) (192,4.8724) (384,5.1388)};
\addlegendentry{\texttt{$\patchsize$=8}}

\nextgroupplot[
    width=4.75cm, height=4.75cm,
    xmode=log, log basis x=2, log ticks with fixed point,
    xtick={12,24,48,96,192,384},
    xlabel={$\dmodel$},
    ymin=23.8517,
    ymax=38.3926,
    ymajorgrids=true, grid style={line width=0.3pt, draw=gray!25},
    axis line style={bluegrey900}, tick style={bluegrey900},
    tick label style={font=\footnotesize},
    label style={font=\small},
    every axis plot/.append style={line width=0.8pt},
    ylabel={Relative Improvement (\%)},
    ylabel near ticks,
    ylabel style={yshift=-5pt},
    legend to name=discardgeneratepatchsizealspredictionaccuracylegend2,
]
\addplot[draw=none, forget plot, name path=lop2s0] coordinates {(12,33.7520) (24,35.1630) (48,35.6179) (96,36.4565) (192,37.0041)};
\addplot[draw=none, forget plot, name path=hip2s0] coordinates {(12,34.1065) (24,35.5273) (48,36.1129) (96,37.1040) (192,37.6137)};
\addplot[fill=blue900, opacity=0.3, forget plot] fill between[of=lop2s0 and hip2s0];
\addplot[color=blue900, solid, mark=*, mark size=1.5pt, mark options={solid, fill=blue900, draw=blue900}] coordinates {(12,33.9292) (24,35.3452) (48,35.8654) (96,36.7803) (192,37.3089)};
\addlegendentry{\texttt{$\patchsize$=1}}
\addplot[draw=none, forget plot, name path=lop2s1] coordinates {(12,31.4748) (24,32.8487) (48,33.7406) (96,34.8480) (192,35.7737) (384,36.6551)};
\addplot[draw=none, forget plot, name path=hip2s1] coordinates {(12,31.8925) (24,33.2444) (48,34.1760) (96,35.2478) (192,36.1584) (384,37.3585)};
\addplot[fill=blue700, opacity=0.3, forget plot] fill between[of=lop2s1 and hip2s1];
\addplot[color=blue700, solid, mark=*, mark size=1.5pt, mark options={solid, fill=blue700, draw=blue700}] coordinates {(12,31.6836) (24,33.0465) (48,33.9583) (96,35.0479) (192,35.9660) (384,37.0068)};
\addlegendentry{\texttt{$\patchsize$=2}}
\addplot[draw=none, forget plot, name path=lop2s2] coordinates {(12,28.2607) (24,29.2502) (48,30.6084) (96,31.2505) (192,32.1403) (384,31.2983)};
\addplot[draw=none, forget plot, name path=hip2s2] coordinates {(12,28.5335) (24,29.6434) (48,30.9429) (96,31.6056) (192,32.4247) (384,33.2259)};
\addplot[fill=blue300, opacity=0.3, forget plot] fill between[of=lop2s2 and hip2s2];
\addplot[color=blue300, solid, mark=*, mark size=1.5pt, mark options={solid, fill=blue300, draw=blue300}] coordinates {(12,28.3971) (24,29.4468) (48,30.7757) (96,31.4280) (192,32.2825) (384,32.2621)};
\addlegendentry{\texttt{$\patchsize$=4}}
\addplot[draw=none, forget plot, name path=lop2s3] coordinates {(12,24.6307) (24,25.9981) (48,26.3480) (96,26.8989) (192,27.6309) (384,25.0673)};
\addplot[draw=none, forget plot, name path=hip2s3] coordinates {(12,25.3037) (24,26.3708) (48,27.0422) (96,27.1567) (192,28.5110) (384,28.1808)};
\addplot[fill=blue100, opacity=0.3, forget plot] fill between[of=lop2s3 and hip2s3];
\addplot[color=blue100, solid, mark=*, mark size=1.5pt, mark options={solid, fill=blue100, draw=blue100}] coordinates {(12,24.9672) (24,26.1845) (48,26.6951) (96,27.0278) (192,28.0709) (384,26.6241)};
\addlegendentry{\texttt{$\patchsize$=8}}
\end{groupplot}
\node[anchor=west, xshift=0.75cm] at (group c3r1.east)
    {\pgfplotslegendfromname{generatepatchsizealspredictionaccuracylegend}};
\end{tikzpicture}
    \vspace{-\baselineskip}
    \caption{Comparison of $MAE_{>5\text{m}}$ computed on GEDI labels \emph{(left)} and ALS labels \emph{(center)} of the \texttt{France} dataset across models with varying patch size ($\patchsize$) and model dimension ($\dmodel$). \emph{(Right:)} Relative improvement when computing the $MAE_{>5\text{m}}$ on ALS labels instead of GEDI labels.}
    \Description{The undelying trend is that for both types of labels (ALS and GEDI), smaller patch sizes and larger model dimensions tend to be better. By comparing the relative improvement, we see that the models already performing better on GEDI labels have an even greater relative improvement of up to 37.4\%, compared to other models (25.1\% improvement).}
    \label{fig:patchsize_als_prediction_accuracy}
\end{figure*}

\subsubsection{Impact of Patch Size \& Model Dimension}

We independently vary the patch size $\patchsize \in \{1, 2, 4, 8\}$ and model dimension $\dmodel \in \{ 12, 24, 48, 96, 192, 384\}$\footnote{We start with $\dmodel$ equal to the number of input channels and double it in each subsequent configuration. Models with $\dmodel = 768$ went out-of-memory regardless of the patch size.}, resulting in 24 model configurations.\footnote{Training for the model with $\patchsize = 1$ and $\dmodel = 384$ took too long.}
All models were trained for six epochs, which was enough for the models to converge (note that one epoch is based on $395\,000$ instances), while not requiring too many resources.
No form of regularization was used, as no overfitting occurred for any of the models, presumably due to the sparsity of the supervision signal and the size and diversity of the dataset.
We employed gradient checkpointing after every stage to reduce the VRAM requirements of our models~\cite{Griewank2000Checkpointing,Chen2016Checkpointing}.
Models with larger patch sizes use $\patchexpand$ layers to increase the resolution to the original input resolution.\footnote{This approach produced better results across the board than using bilinear upsampling in the prediction space.}

\paragraph{Quantitative Evaluation:}
Figure~\ref{fig:patchsize_performance} shows the two regression metrics, $MAE_{>5\text{m}}$ and $R^2$, in dependence of the model dimension and patch size, averaged over five training runs.
Both metrics follow the same trend: For a fixed model dimension, a smaller patch size results in stronger performance, which is in line with the patchification scaling laws~\cite{Wang2025ScalingLawsInPatchification}. The larger the model dimension, the weaker the effect.
At the same time, for a fixed patch size, a larger model dimension generally improves performance, but with diminishing returns.

\paragraph{Qualitative Evaluation:}
Figures~\ref{fig:qualitative_patchsizes} and~\ref{fig:qualitative_patchsizes_2} (in the appendix) show predictions of models with varying patch size on exemplary validation patches of the \texttt{France} dataset.
In general, the differences in perceptual quality are most pronounced for image regions with fine details, like narrow streams or paths through a forest.
For models with a patch size $\patchsize > 1$, unpleasant grid-like artifacts arise, with the size of a grid ``cell'' equal to the patch size.
To quantify the grids' intensity, inspired by \citet{Wu1997GBIM} and \citet{Wang2002PerceptualQuality}, we compute the differences of two adjacent pixels' residuals (based on the ALS labels) within a cell of length $\patchsize$ (referred to as $A_\patchsize$) and across a cell's borders ($B_\patchsize$). Their ratio ($B_\patchsize/A_\patchsize$) is shown in Figure~\ref{fig:boundary_ratio} in the appendix, indicating that larger patch sizes lead to stronger grid artifacts. This is plausible, as more upsampling steps need to be performed in the decoder.

\paragraph{Runtime and Memory Analysis:}
However, the performance gains come at the price of reduced efficiency, as shown in Figure~\ref{fig:patchsize_efficiency}. Transformers are resource-intensive---even when using an attention variant with linear complexity, like efficient attention.
Training time for a single instance and required memory scale roughly by a factor of $1/\patchsize^2$, i.e., halving the patch size roughly quadruples the training time.
The training time as well as the models' number of parameters grow quadratically in the model dimension, due to the nature of fully connected linear layers. The number of parameters is independent of the patch size.

\begin{figure*}[t]
    \centering
    \resizebox{0.98\linewidth}{!}{\begin{tikzpicture}[rounded corners=0pt]

\begin{axis}[
    name=gedi,
    width=6cm, height=6cm,
    xmin=0, xmax=50.0,
    ymin=0, ymax=50.0,
    xtick={0,10,20,30,40,50}, ytick={0,10,20,30,40,50},
    enlargelimits=false, tick align=inside, axis on top,
    axis line style={bluegrey900}, tick style={bluegrey900},
    rounded corners=0pt,
    xlabel={GEDI Label (m)}, ylabel={Prediction (m)},
    ylabel style={yshift=-1mm},
    colormap name=ScatterMagma,
    point meta min=0, point meta max=4.9760,
]
\addplot[matrix plot*, point meta=explicit, opacity=0,
    mesh/rows=2, mesh/cols=2]
    table[x=x, y=y, meta=z, row sep=crcr] {%
        x y z \\
        0 0 0.000000 \\ 50.0 0 4.976043 \\
        0 50.0 0.000000 \\ 50.0 50.0 4.976043 \\
    };
\addplot graphics[xmin=0, xmax=50.0, ymin=0, ymax=50.0]
    {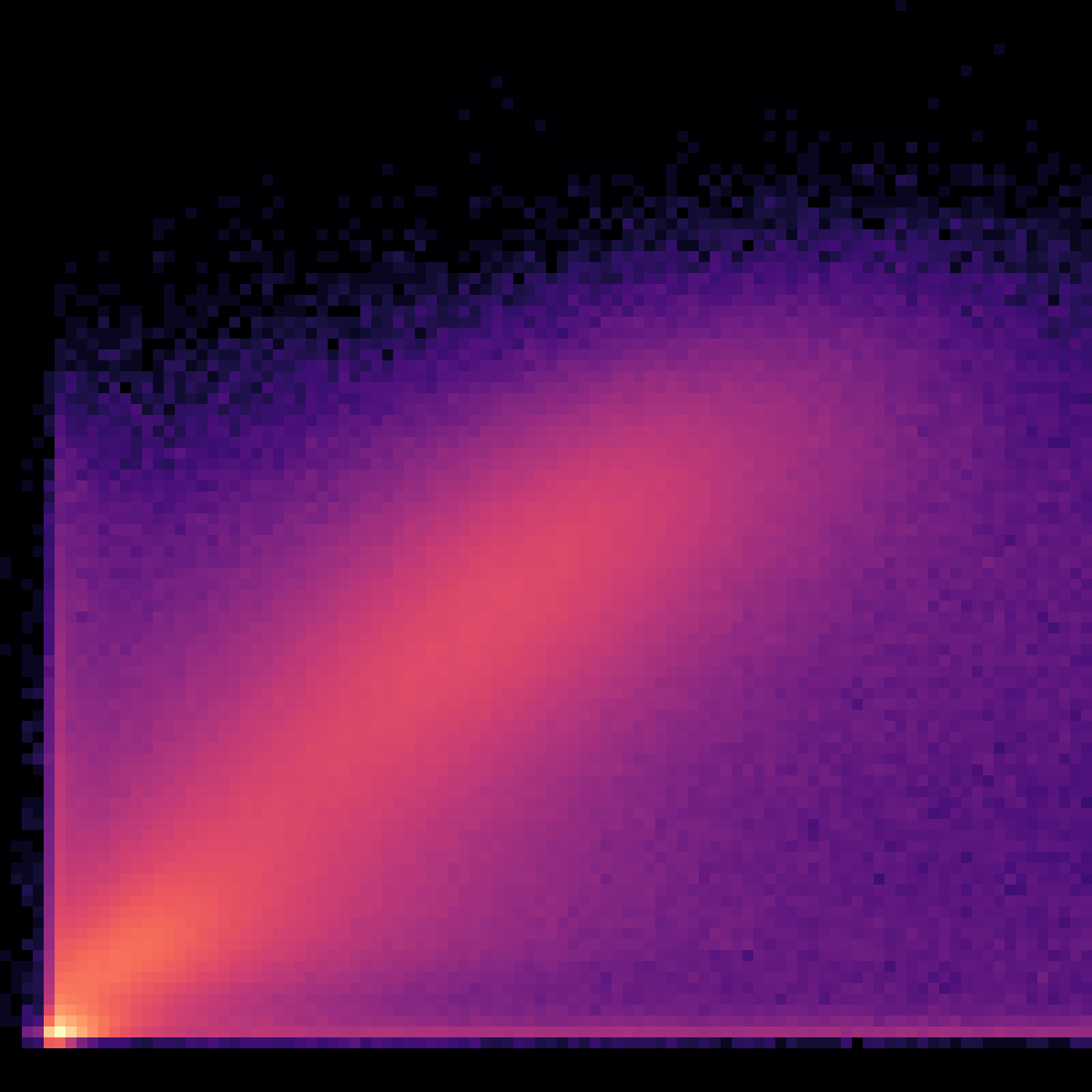};
\addplot[black, dashed, thick, domain=0:50.0] {x};
    \node[anchor=north west, fill=white, fill opacity=0.8,
         text opacity=1, rounded corners=0pt, inner sep=3pt,
         font=\small, align=left]
        at (axis cs:1.00,49.00)
        {$MAE_{>5\text{m}} = 6.06$\,m \\\ $R^2 = 0.426$};
\end{axis}

\begin{axis}[
    name=als,
    at={(gedi.outer south east)}, anchor=outer south west,
    xshift=-0.5mm,
    width=6cm, height=6cm,
    xmin=0, xmax=50.0,
    ymin=0, ymax=50.0,
    xtick={0,10,20,30,40,50}, ytick={0,10,20,30,40,50},
    enlargelimits=false, tick align=inside, axis on top,
    axis line style={bluegrey900}, tick style={bluegrey900},
    rounded corners=0pt,
    xlabel={ALS Label (m)}, yticklabels={},
    colormap name=ScatterMagma,
    colorbar,
    colorbar style={
        xshift=-0.5mm,
        width=4mm,
        axis line style={bluegrey900}, tick style={bluegrey900},
        rounded corners=0pt,
        ytick={0,1,2,3,4},
        yticklabels={$10^{0}$,$10^{1}$,$10^{2}$,$10^{3}$,$10^{4}$},
        yticklabel style={font=\scriptsize, anchor=west},
        tick align=inside,
        ylabel={Number of pixels},
        ylabel shift=-1mm,
    },
    point meta min=0, point meta max=4.9760,
]
\addplot[matrix plot*, point meta=explicit, opacity=0,
    mesh/rows=2, mesh/cols=2]
    table[x=x, y=y, meta=z, row sep=crcr] {%
        x y z \\
        0 0 0.000000 \\ 50.0 0 4.976043 \\
        0 50.0 0.000000 \\ 50.0 50.0 4.976043 \\
    };
\addplot graphics[xmin=0, xmax=50.0, ymin=0, ymax=50.0]
    {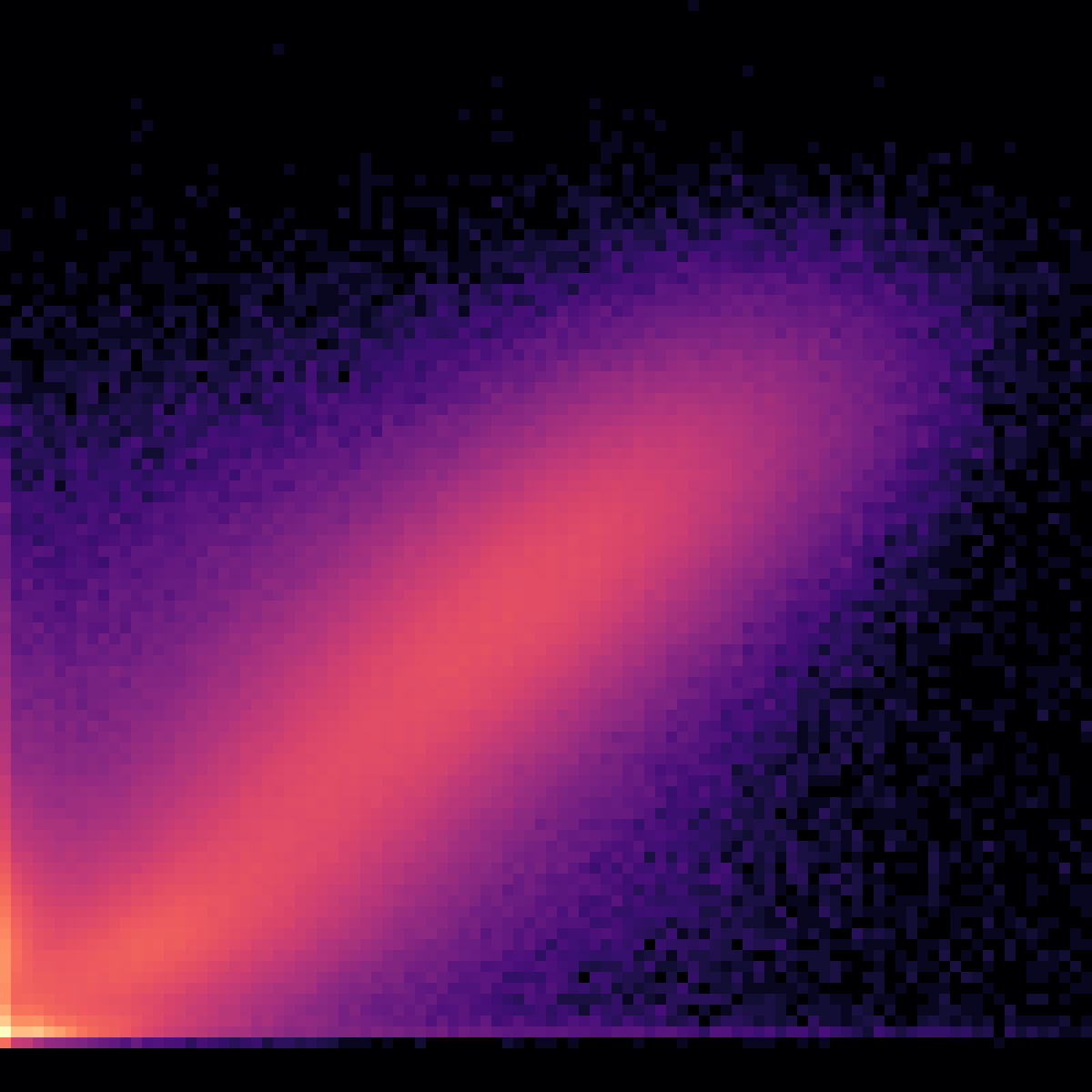};
\addplot[black, dashed, thick, domain=0:50.0] {x};
    \node[anchor=north west, fill=white, fill opacity=0.8,
         text opacity=1, rounded corners=0pt, inner sep=3pt,
         font=\small, align=left]
        at (axis cs:1.00,49.00)
        {$MAE_{>5\text{m}} = 3.84$\,m \\\ $R^2 = 0.751$};
\end{axis}

\begin{axis}[
    name=diff,
    at={(gedi.outer south east -| als.outer east)}, anchor=outer south west,
    xshift=20mm,
    width=6cm, height=6cm,
    xmin=0, xmax=50.0,
    ymin=0, ymax=50.0,
    xtick={0,10,20,30,40,50}, ytick={0,10,20,30,40,50},
    enlargelimits=false, tick align=inside, axis on top,
    axis line style={bluegrey900}, tick style={bluegrey900},
    rounded corners=0pt,
    xlabel={ALS Label (m)}, ylabel={Prediction (m)},
    ylabel style={yshift=-1mm},
    colormap name=ScatterDiverging,
    colorbar,
    colorbar style={
        xshift=-0.5mm,
        width=4mm,
        axis line style={bluegrey900}, tick style={bluegrey900},
        rounded corners=0pt,
        ytick={-200,-150,-100,-50,0,50,100,150,200},
        yticklabels={$\leq-200$,$-150$,$-100$,$-50$,$0$,$50$,$100$,$150$,$\geq200$},
        yticklabel style={font=\scriptsize, anchor=west},
        tick align=inside,
        ylabel={Difference: ALS$-$GEDI},
        ylabel shift=-2mm,
    },
    point meta min=-200.00000000, point meta max=200.00000000,
]
\addplot[matrix plot*, point meta=explicit, opacity=0,
    mesh/rows=2, mesh/cols=2]
    table[x=x, y=y, meta=z, row sep=crcr] {%
        x y z \\
        0 0 -200.000000 \\ 50.0 0 200.000000 \\
        0 50.0 -200.000000 \\ 50.0 50.0 200.000000 \\
    };
\addplot graphics[xmin=0, xmax=50.0, ymin=0, ymax=50.0]
    {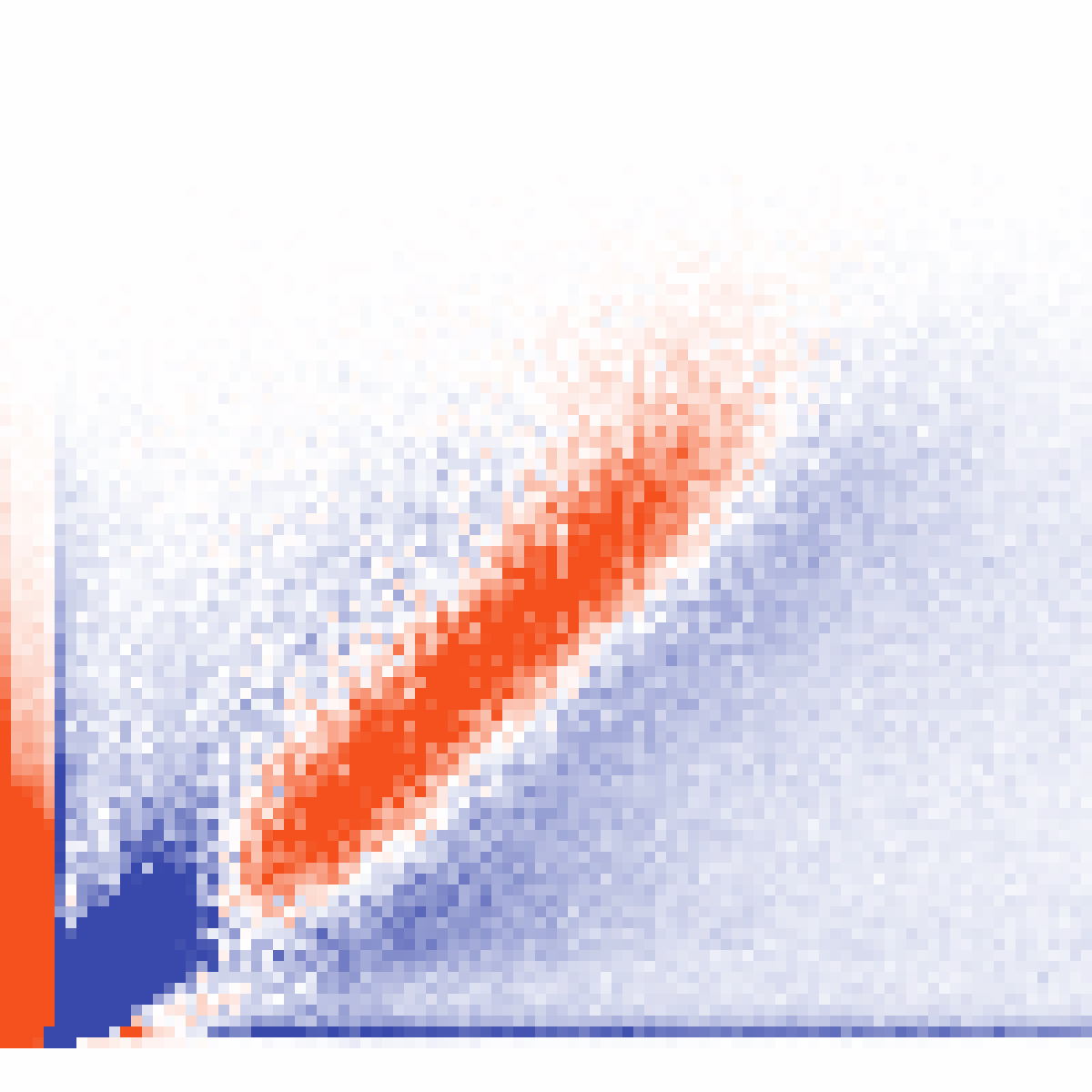};
\addplot[black, dashed, thick, domain=0:50.0] {x};
\end{axis}

\end{tikzpicture}}
    \vspace{-\baselineskip}
    \caption{\emph{(Left:)} Reference canopy height (GEDI/ALS label) vs. predictions by the model with $\patchsize=1,\dmodel=96$. Both heatmaps are computed only for pixels where both labels exist. \emph{(Right:)} Difference of the heatmaps, where red color indicates a higher density of the ALS heatmap and blue color a higher density of the GEDI heatmap.}
    \Description{The scatter heatmaps indicate that the number of predictions close to the perfect-fit line are much higher when comparing them to ALS labels than to GEDI labels.}
    \label{fig:scatter_heatmap_p1_dm96}
\end{figure*}
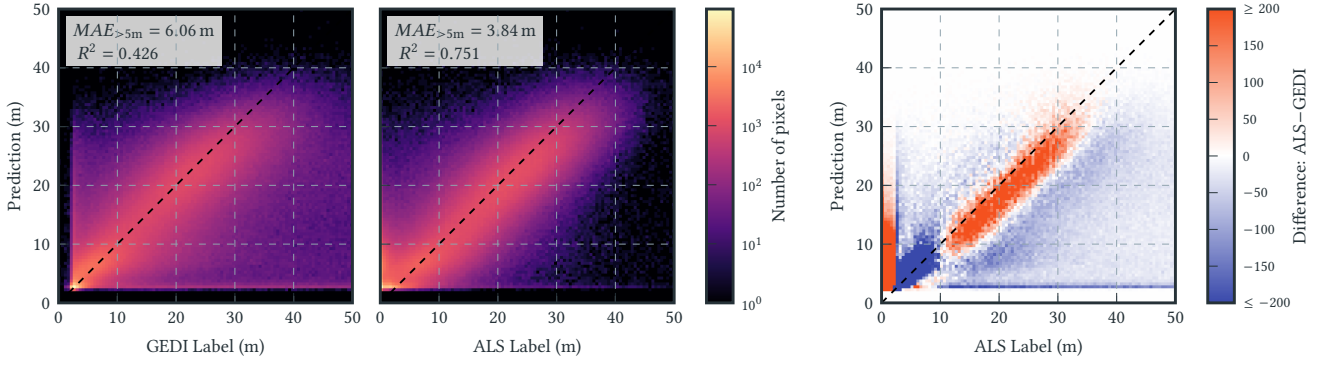

\paragraph{GEDI vs. ALS Ground Truth:}
As the GEDI labels are inherently noisy, we also evaluated our models on the ALS labels given in the \texttt{France} dataset. The results of this evaluation are shown in Figure~\ref{fig:patchsize_als_prediction_accuracy}.
Note that here the $MAE_{>5\text{m}}$ computed on the GEDI labels is substantially worse than the one computed on the \texttt{Europe} dataset, as the model has not seen any inputs close to the \texttt{France} dataset, which was only used for testing, not for training.
Across all models, the $MAE_{>5\text{m}}$ is significantly better when computed on ALS labels instead of GEDI labels. The relative improvement ranges from $25.1\%$ to $37.4\%$.
Interestingly, the models which performed better initially (i.e., models with a smaller patch size or larger model dimension) have an even bigger relative improvement when evaluated on the higher-quality ALS labels.
This implies that models with larger patch sizes get evaluated favorably by GEDI data, misrepresenting the actual performance difference.
An explanation is that due to the label noise present with GEDI labels, models producing smoother predictions have an advantage.
It can also be observed that for smaller patch sizes, models with comparably smaller model dimensions can perform better. The best models have the configurations ($\patchsize=1,\dmodel=96$), ($\patchsize=2,\dmodel=192$), ($\patchsize=4,\dmodel=192$), and ($\patchsize=8,\dmodel=192$). That the optimal model dimension is smaller, when choosing a smaller patch size, can be explained by viewing patchification as a compression step~\cite{Wang2025ScalingLawsInPatchification}: Smaller patch sizes mean less information has to be compressed into a single token, so the dimensionality of each token can be reduced.

Analyzing the error for various bins of label heights (see Figure~\ref{fig:binned_errors} in the appendix), we observe that smaller patch sizes result in a lower variance of errors on ALS labels for smaller trees, and a smaller absolute error for taller trees.

Figure~\ref{fig:scatter_heatmap_p1_dm96} shows the joint distribution of the labels (one with GEDI, the other with ALS labels) and the model predictions as heatmaps, and the difference between the two heatmaps. The difference for labels smaller than \SI{10}{m} arises due to a particularity of GEDI labels: Structures shorter than \SI{2.5}{m} are generally not reliably detected, resulting in labels of around \SI{2.5}{m}, even when there is no vegetation. For labels greater than \SI{10}{m}, it can be clearly seen that the density around the perfect-fit line (prediction = label) is consistently higher for the ALS-vs-prediction heatmap. Further away from the perfect-fit line, the GEDI-vs-prediction heatmap has a greater density. This shows that the model's predictions match the ALS labels better than the GEDI labels (which it was trained on) across almost all tree heights. The observed effect is less pronounced for models with larger patch sizes (see Figure~\ref{fig:scatter_heatmap_p8_dm192} in the appendix). Figure~\ref{fig:als_gedi_label_correlation} in the appendix shows that GEDI and ALS labels have a surprisingly small correlation. 

\subsubsection{Impact of Attention Mechanism}
\label{sec:results_attention}

\begin{figure}[t]
    \centering
    \begin{tikzpicture}
  \begin{axis}[
      width=6.5cm, height=5.5cm, scale only axis,
      xmin=68, xmax=112,
      ymin=3.75, ymax=4.6,
      xtick={70, 80, 90, 100, 110},
      xticklabels={{70}, {80}, {90}, {170}, {180}},
      ytick={3.8, 3.9, 4, 4.1, 4.2, 4.3, 4.4, 4.5, 4.6},
      xlabel={Training step (ms)},
      ylabel={$MAE_{>5\text{m}}$ (ALS) (m)},
      xmajorgrids=true, ymajorgrids=true,
      grid style={line width=0.3pt, draw=gray!25},
      axis line style={bluegrey900}, tick style={bluegrey900},
      line width=0.5pt,
      tick label style={font=\footnotesize},
      label style={font=\small},
      xlabel near ticks,
      ylabel near ticks,
      tick align=outside,
      enlargelimits=false,
      legend style={at={(0.5,1)}, anchor=south, yshift=0.15cm, draw=bluegrey900, line width=0.3pt, fill=white, fill opacity=1, text opacity=1, font=\footnotesize\ttfamily, inner xsep=2pt, inner ysep=2pt, column sep=0pt, /tikz/every even column/.append style={column sep=8pt}, row sep=2pt},
      legend cell align=left,
      legend columns=4,
      legend image code/.code={\draw[#1] plot coordinates {(0.15cm,0cm)};},
  ]
      \fill[white] ([yshift=0.4pt]axis cs:94.0000,3.75)
          rectangle ([yshift=-0.4pt]axis cs:96.0000,4.6);
      \addlegendimage{empty legend}
      \addlegendentry{\textnormal{Encoder}}
      \addlegendimage{legend image code/.code={\draw[amber500, line width=1.6pt] (0cm,0cm) -- (0.3cm,0cm);}}
      \addlegendentry{eff}
      \addlegendimage{legend image code/.code={\draw[green500, line width=1.6pt] (0cm,0cm) -- (0.3cm,0cm);}}
      \addlegendentry{flash}
      \addlegendimage{legend image code/.code={\draw[pink500, line width=1.6pt] (0cm,0cm) -- (0.3cm,0cm);}}
      \addlegendentry{swin}
      \addlegendimage{empty legend}
      \addlegendentry{\textnormal{Decoder}}
      \addlegendimage{only marks, mark=*, mark size=2.2pt, color=gray!55, mark options={solid, fill=gray!55, draw=gray!55}}
      \addlegendentry{eff}
      \addlegendimage{only marks, mark=triangle*, mark size=2.65pt, color=gray!55, mark options={solid, fill=gray!55, draw=gray!55}}
      \addlegendentry{flash}
      \addlegendimage{only marks, mark=square*, mark size=1.9pt, color=gray!55, mark options={solid, fill=gray!55, draw=gray!55}}
      \addlegendentry{swin}
      \addplot[forget plot, only marks, mark=*, mark size=2.2pt, color=amber500, mark options={solid, fill=amber500, draw=amber500, fill opacity=0.8}, error bars/.cd, y dir=both, y explicit, error bar style={solid, draw=amber500, line width=0.6pt, draw opacity=0.85}, error mark=-, error mark options={mark size=1.3pt, solid, draw=amber500, line width=0.6pt, draw opacity=0.85}]
      coordinates {(77.7337,3.8423) +- (0,0.0181)};
      \addplot[forget plot, only marks, mark=triangle*, mark size=2.65pt, color=amber500, mark options={solid, fill=amber500, draw=amber500, fill opacity=0.8}, error bars/.cd, y dir=both, y explicit, error bar style={solid, draw=amber500, line width=0.6pt, draw opacity=0.85}, error mark=-, error mark options={mark size=1.3pt, solid, draw=amber500, line width=0.6pt, draw opacity=0.85}]
      coordinates {(83.8104,3.8708) +- (0,0.0424)};
      \addplot[forget plot, only marks, mark=square*, mark size=1.9pt, color=amber500, mark options={solid, fill=amber500, draw=amber500, fill opacity=0.8}, error bars/.cd, y dir=both, y explicit, error bar style={solid, draw=amber500, line width=0.6pt, draw opacity=0.85}, error mark=-, error mark options={mark size=1.3pt, solid, draw=amber500, line width=0.6pt, draw opacity=0.85}]
      coordinates {(79.6912,3.9412) +- (0,0.0210)};
      \addplot[forget plot, only marks, mark=*, mark size=2.2pt, color=green500, mark options={solid, fill=green500, draw=green500, fill opacity=0.8}, error bars/.cd, y dir=both, y explicit, error bar style={solid, draw=green500, line width=0.6pt, draw opacity=0.85}, error mark=-, error mark options={mark size=1.3pt, solid, draw=green500, line width=0.6pt, draw opacity=0.85}]
      coordinates {(100.5162,3.8948) +- (0,0.0223)};
      \addplot[forget plot, only marks, mark=triangle*, mark size=2.65pt, color=green500, mark options={solid, fill=green500, draw=green500, fill opacity=0.8}, error bars/.cd, y dir=both, y explicit, error bar style={solid, draw=green500, line width=0.6pt, draw opacity=0.85}, error mark=-, error mark options={mark size=1.3pt, solid, draw=green500, line width=0.6pt, draw opacity=0.85}]
      coordinates {(106.6332,3.9515) +- (0,0.0493)};
      \addplot[forget plot, only marks, mark=square*, mark size=1.9pt, color=green500, mark options={solid, fill=green500, draw=green500, fill opacity=0.8}, error bars/.cd, y dir=both, y explicit, error bar style={solid, draw=green500, line width=0.6pt, draw opacity=0.85}, error mark=-, error mark options={mark size=1.3pt, solid, draw=green500, line width=0.6pt, draw opacity=0.85}]
      coordinates {(102.4623,4.1943) +- (0,0.1771)};
      \addplot[forget plot, only marks, mark=*, mark size=2.2pt, color=pink500, mark options={solid, fill=pink500, draw=pink500, fill opacity=0.8}, error bars/.cd, y dir=both, y explicit, error bar style={solid, draw=pink500, line width=0.6pt, draw opacity=0.85}, error mark=-, error mark options={mark size=1.3pt, solid, draw=pink500, line width=0.6pt, draw opacity=0.85}]
      coordinates {(71.9520,4.0045) +- (0,0.0338)};
      \addplot[forget plot, only marks, mark=triangle*, mark size=2.65pt, color=pink500, mark options={solid, fill=pink500, draw=pink500, fill opacity=0.8}, error bars/.cd, y dir=both, y explicit, error bar style={solid, draw=pink500, line width=0.6pt, draw opacity=0.85}, error mark=-, error mark options={mark size=1.3pt, solid, draw=pink500, line width=0.6pt, draw opacity=0.85}]
      coordinates {(78.0410,4.0426) +- (0,0.0770)};
      \addplot[forget plot, only marks, mark=square*, mark size=1.9pt, color=pink500, mark options={solid, fill=pink500, draw=pink500, fill opacity=0.8}, error bars/.cd, y dir=both, y explicit, error bar style={solid, draw=pink500, line width=0.6pt, draw opacity=0.85}, error mark=-, error mark options={mark size=1.3pt, solid, draw=pink500, line width=0.6pt, draw opacity=0.85}]
      coordinates {(74.0178,4.4628) +- (0,0.0671)};
      \coordinate (xbreakbottom) at (axis cs:95.0000,3.75);
      \coordinate (xbreaktop) at (axis cs:95.0000,4.6);
  \end{axis}
  \draw[white, line width=1.0pt] ([xshift=-1.95pt]xbreakbottom) -- ([xshift=1.95pt]xbreakbottom);
  \draw[white, line width=1.0pt] ([xshift=-1.95pt]xbreaktop) -- ([xshift=1.95pt]xbreaktop);
  \draw[bluegrey900, line width=0.5pt, decorate,
        decoration={zigzag, amplitude=1.6pt, segment length=7.5pt}, rounded corners=0pt]
        ([xshift=-1.60pt,yshift=-0.35pt]xbreakbottom) -- ([xshift=-1.60pt,yshift=0.35pt]xbreaktop);
  \draw[bluegrey900, line width=0.5pt, decorate,
        decoration={zigzag, amplitude=1.6pt, segment length=7.5pt}, rounded corners=0pt]
        ([xshift=1.60pt,yshift=-0.35pt]xbreakbottom) -- ([xshift=1.60pt,yshift=0.35pt]xbreaktop);
\end{tikzpicture}
    \vspace{-\baselineskip}
    \caption{Trade-off between performance ($MAE_{>5\text{m}}$ computed on ALS labels) and efficiency (training time for a single-image batch). Compared are nine models with varying attention mechanisms in the encoder (indicated by marker color) and decoder (indicated by marker shape): efficient (\texttt{eff}), flash (\texttt{flash}), and shifted window (\texttt{swin}). Shown are the mean and standard deviation of five training runs per configuration.}
    \Description{}
    \label{fig:attention_performance_efficiency_tradeoff}
\end{figure}
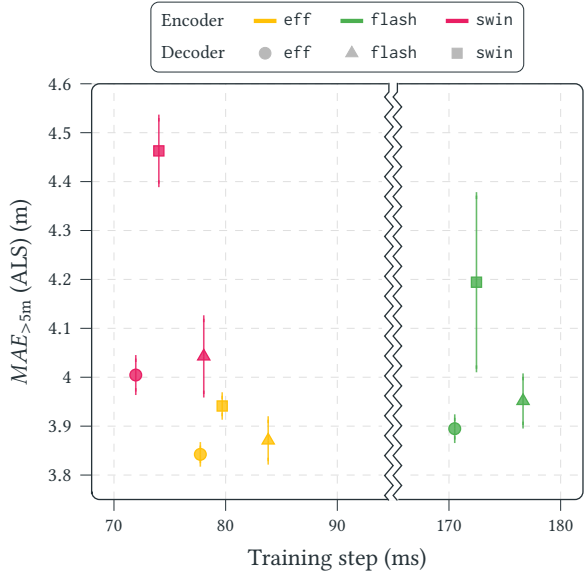

Next, we fix the hyperparameters to $\patchsize=1$ and $\dmodel=96$ to investigate the impact of three well-known attention mechanisms in isolation: efficient attention ($\eff$), flash attention ($\flash$), and shifted window attention ($\shiftedwindow$), which we vary individually for the encoder and decoder.
The window size of $\shiftedwindow$ attention is set to $7 \times 7$ tokens. The other two variants do not require setting any hyperparameters other than the ones shared by all attention mechanisms, like the number of attention heads and attention dropout.

Figure~\ref{fig:attention_performance_efficiency_tradeoff} shows the performance-efficiency trade-off for models with varying attention mechanisms, where better models are positioned towards the bottom left corner.
We observe significant performance gains when using global attention (\texttt{eff} or \texttt{flash}) instead of local attention (\texttt{swin}) in the decoder and/or encoder.
The model with the best performance, while still being comparatively efficient, uses efficient attention in the encoder and decoder---as was also used in the previous experiment---and has an average $MAE_{>5\text{m}}$ (computed on ALS labels) of \SI{3.84}{m} and a runtime of about \SI{78}{ms}.

The training time of models with \texttt{swin} and \texttt{eff} attention is very similar, while \texttt{flash} takes more than twice as long. In general, the impact of the encoder's attention mechanism is significantly bigger than the decoder's, as the keys and values are pooled to the lowest resolution in the decoder block (see Figure~\ref{fig:umix_block}). All architectures have a similar peak VRAM: Training the model with a batch size of $1$ requires about \SI{2.66}{GiB} for the architectures using \texttt{eff} and \texttt{flash} attention and about \SI{2.81}{GiB} for the architecture using \texttt{swin} attention throughout the model. In contrast, vanilla attention already produces an out-of-memory error for inputs of this size.\footnote{Training was performed on NVIDIA's RTX 4090 GPUs, which have \SI{24}{GiB} of VRAM.}

For a qualitative evaluation, predictions on an exemplary validation patch can be seen in Figure~\ref{fig:qualitative_attention} in the appendix. 

\subsubsection{Comparison with Baseline Models}

\begin{figure*}[t]
    \centering
    \resizebox{\linewidth}{!}{
        \input{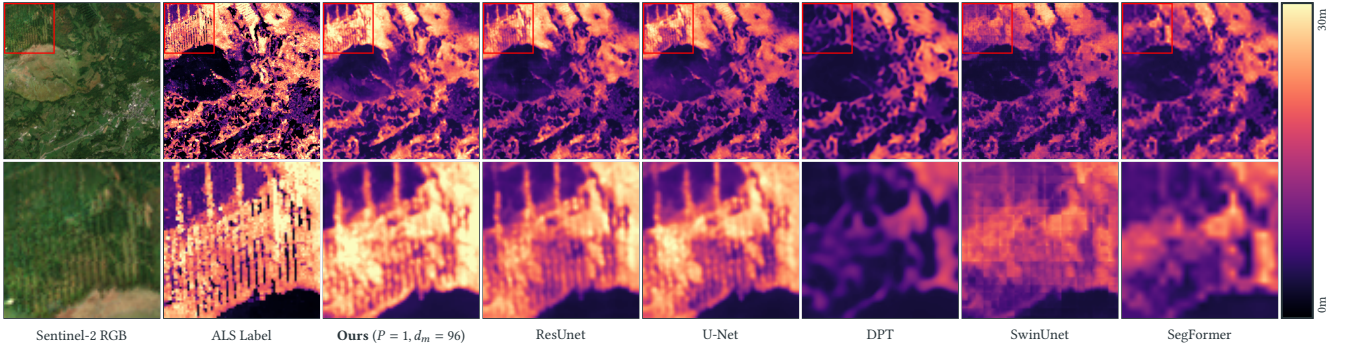}
    }
    \vspace{-\baselineskip}
    \caption{Qualitative comparison of predictions made by various models.
    The UNet and ResUnet are strong CNN baselines, while ViT-based architectures with a high patch size produce predictions which are blurry or show a grid pattern.}
    \Description{This example patch features trees of varying heights, sometimes with fine details, which are well identified by our proposed model as well as the CNN-based models (U-Net, ResUnet). Reference Vision Transformer-based architectures with larger patch sizes struggle to generate high-quality predictions.}
    \label{fig:qualitative_baselines}
\end{figure*}

We compare the performance of the best configuration of the previous experiments\footnote{I.e., when setting patch size $\patchsize=1$ and model dimension $\dmodel=96$.} with several models commonly used for dense prediction tasks.
The U-Net architecture~\cite{Ronneberger2015Unet} consists of a symmetric convolutional encoder and decoder and has been used to generate a global canopy height map~\cite{Pauls2024EstimatingCanopyHeight} and predict canopy height and above-ground biomass across France~\cite{Schwartz2023FORMS}.
The Residual U-Net (ResUnet) has been used as a baseline by~\citet{Tolan2024VeryHighRes}, and their final model is based on the Dense Prediction Transformer, which has a default patch size of $\patchsize=16$~\cite{Ranftl2021DPT}.
The SegFormer consists of a Mix Transformer encoder with patch size $\patchsize=4$ and a lightweight Multilayer Perceptron decoder~\cite{Xie2021SegFormer}.
Based on the Swin Transformer~\cite{Liu2021Swin}, the Swin-Unet is a symmetric encoder-decoder architecture, and was adapted by \citet{Pauls2026ECHOSAT} to produce a multi-year global canopy height map.

\begin{table}[t]
    \centering
    \caption{Comparison of well-known dense prediction baselines using three predictive metrics, as well as the runtime per training step (ts) in ms and peak VRAM usage in MiB. The model \texttt{Ours} uses $\patchsize = 1$, $\dmodel = 96$ and is averaged over its five seed repeats; the baselines were trained once.}
    \resizebox{\columnwidth}{!}{
    \begin{tabular}{lccccc}
    \toprule
    Model & $MAE_{>5\text{m}}$ $\downarrow$ & $R^2$ $\uparrow$ & $MAE_{>5\text{m}}$ $\downarrow$ & ts $\downarrow$ & VRAM $\downarrow$ \\
          & (GEDI) & (GEDI) & (ALS) & & \\
    \midrule
    SegFormer & 5.02 & 0.641 & 4.62 & 52.2 & \phantom{0}525 \\
    SwinUnet & 5.17 & 0.629 & 4.66 & 45.3 & \phantom{0}551 \\
    DPT & 5.35 & 0.615 & 5.04 & 18.1 & 2069 \\
    U-Net & 4.59 & 0.672 & 4.14 & \phantom{0}8.2 & \phantom{0}851 \\
    ResUnet & 4.56 & 0.674 & 4.19 & \phantom{0}\textbf{7.4} & \phantom{0}\textbf{517} \\
    \textbf{Ours} & \textbf{4.31} & \textbf{0.687} & \textbf{3.84} & 77.7 & 2722 \\
    \bottomrule
    \end{tabular}
    }
    \label{tab:baseline_metrics}
\end{table}

Table~\ref{tab:baseline_metrics} shows the resulting metrics and Figure~\ref{fig:qualitative_baselines} the predictions on an example patch. While our pixel-level Transformer model achieves the strongest predictive performance (measured by $MAE_{>5\text{m}}$ and $R^2$), the results also show that U-Net-based architectures remain highly competitive: They offer substantially faster and more resource-efficient training while achieving performance superior to Transformer models with larger patch sizes. Meanwhile, our experiments also demonstrate that carefully configured Transformer architectures can surpass other models, suggesting that the additional computational cost can translate into meaningful gains in prediction quality.

\subsubsection{Discussion}

Our results should be interpreted as relative comparisons within a controlled experimental setting. As our focus lies on comparing architectural choices rather than maximizing canopy height estimation performance, the reported metrics should not be directly compared with those of existing canopy height products.

The architectural choices considered in this work only represent a small subset of the design space of Transformer-based models. While we focused on patch size, model dimension, and the attention mechanism, other factors such as model depth and decoder design may further improve performance.
In particular, a stronger decoder might be able to reconstruct the spatial dimensions lost in the patchification step.

If the training is highly resource-constrained and a small performance loss is acceptable, Transformer-based models with a patch size of $\patchsize=2$ or CNN-based models remain strong alternatives.

\section{Conclusion}

We conducted experiments on adapting important hyperparameters of Vision Transformer-based models: patch size, model dimension and the attention mechanism used.
The results indicate that, while pixel-level Transformers are remarkably resource-hungry, choosing an appropriate model dimension and using an efficient attention approximation gives rise to models with a very strong performance.
Architectures operating at a pixel level have the further benefit of being intuitive (as the image is not split into arbitrary chunks) and simpler (as the model predictions inherently have the same resolution as the input).
For the studied task of canopy height prediction, even though the models are trained on noisy labels derived from GEDI data, they indeed learn fine details, which can be evaluated using higher-quality ALS labels that are only available for selected regions.
Only evaluating the models on GEDI labels paints an incomplete picture of model performance. Note that, while the benchmark comparison is based on the task of canopy height prediction, we believe that similar observations can be made for other remote sensing tasks such as biomass estimation, soil moisture mapping, or crop yield forecasting. 
Through our experiments, we hope to motivate other researchers to explore efficient pixel-level Transformers for dense prediction tasks, especially when given medium- or low-resolution satellite imagery as input.

\begin{acks}
This work was supported via the AI4Forest project, which is funded by the German Federal Ministry of Research, Technology and Space (BMFTR; grant number 01IS23025A) and the French National Research Agency (ANR; grant number ANR-22-FAI1-0002).
Calculations for this publication were performed on the HPC cluster PALMA II of the University of M\"unster, subsidized by the DFG (INST 211/667-1).
\end{acks}

\bibliographystyle{ACM-Reference-Format}
\bibliography{library}

\clearpage

\appendix

\begin{figure*}[tp]
    \section{Supporting Figures}
    \centering
    \includegraphics[width=0.85\textwidth]{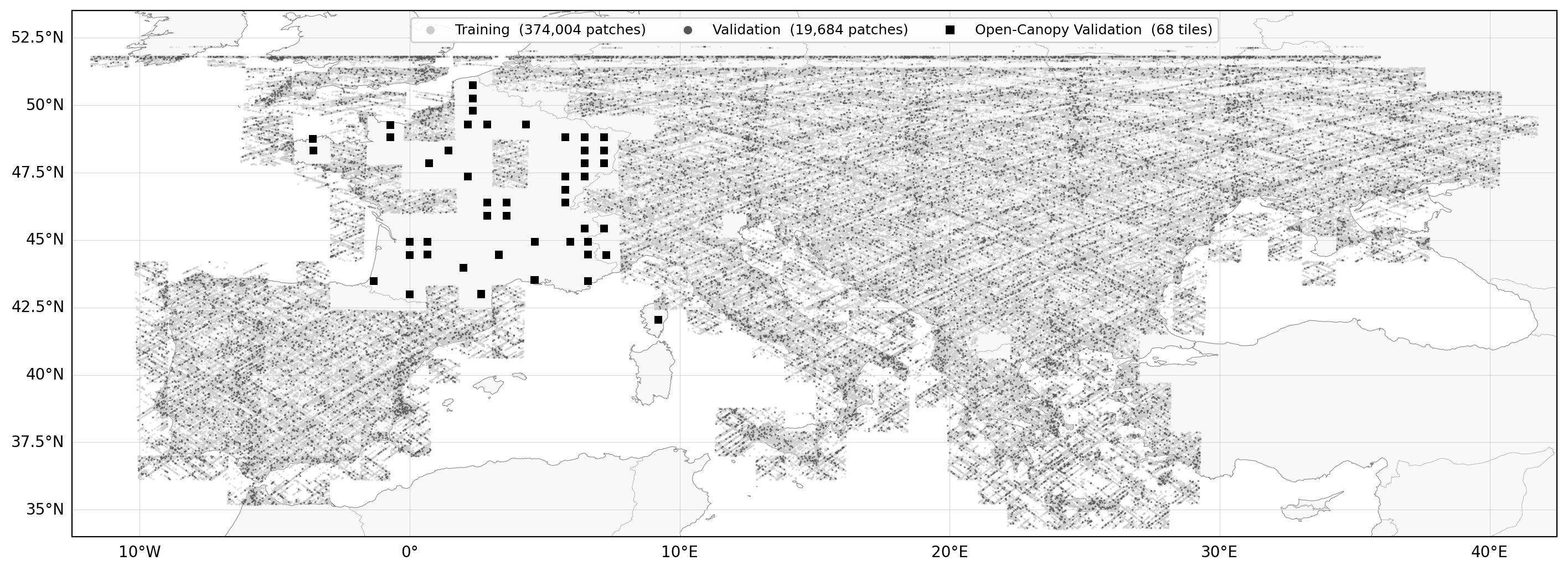}
    \vspace{-\baselineskip}
    \caption{Geographical distribution of training patches (light grey) and validation patches (dark grey) of the \texttt{Europe} dataset, and validation patches of the \texttt{France} dataset (black).}
    \Description{A map showing Europe. Parts of France are validation patches from the France dataset and thus excluded from the training and validation set, which covers the rest of Europe, from about 35°N to 52°N and 10°W to 42°E.}
    \label{fig:dataset_extent}
\end{figure*}

\begin{figure*}[tp]
    \centering
    \input{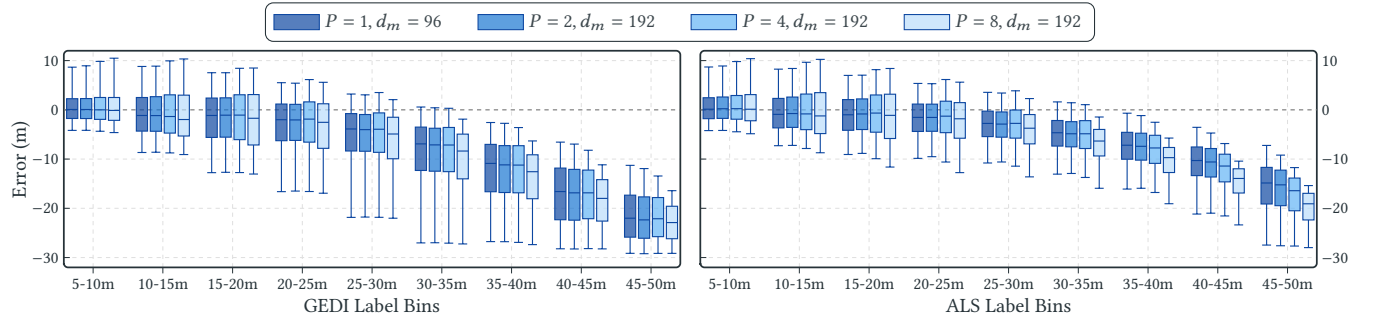}
    \vspace{-\baselineskip}
    \caption{Comparison of errors across labels of different height for GEDI labels on the \texttt{Europe} dataset \emph{(left)} and ALS labels on the \texttt{France} dataset \emph{(right)}.
    All models underestimate taller trees' height, but models with smaller patch sizes less severely.}
    \Description{The models have lower errors for the ALS labels of the France dataset, than for the GEDI labels of the Europe dataset, especially for taller trees.}
    \label{fig:binned_errors}
\end{figure*}

\begin{figure*}[tp]
    \centering
    \begin{minipage}[t]{0.67\textwidth}
        \centering
        \
\usepgfplotslibrary{fillbetween}
\begin{tikzpicture}
\begin{axis}[
    width=7cm, height=4cm,
    xmode=log, log basis x=2, log ticks with fixed point,
    xtick={12,24,48,96,192,384},
    xlabel={$\dmodel$},
    ymin=0.9893, ymax=1.1894,
    ymajorgrids=true, grid style={line width=0.3pt, draw=gray!25},
    axis line style={bluegrey900}, tick style={bluegrey900},
    tick label style={font=\footnotesize},
    label style={font=\small},
    every axis plot/.append style={line width=0.8pt},
    ylabel={Ratio $\frac{B_\patchsize}{A_\patchsize}$},
    ylabel near ticks,
    ytick={1.00, 1.05, 1.10, 1.15},
    yticklabel style={/pgf/number format/.cd, fixed, fixed zerofill, precision=2},
    ylabel style={yshift=-2pt},
    legend style={at={(1,0.5)}, anchor=west, xshift=0.75cm, legend columns=1, font=\footnotesize, draw=bluegrey900, line width=0.3pt, fill=white, inner xsep=2pt, inner ysep=2pt, column sep=3pt, row sep=1pt},
    legend cell align=left,
]
\addplot[gray!60, dashed, line width=0.6pt, forget plot] coordinates {(12,1.0000) (384,1.0000)};
\addplot[draw=none, forget plot, name path=lop0s0] coordinates {(12,1.0004) (24,1.0005) (48,1.0003) (96,1.0023) (192,1.0040)};
\addplot[draw=none, forget plot, name path=hip0s0] coordinates {(12,1.0016) (24,1.0009) (48,1.0007) (96,1.0029) (192,1.0045)};
\addplot[fill=blue900, opacity=0.3, forget plot] fill between[of=lop0s0 and hip0s0];
\addplot[color=blue900, solid, mark=*, mark size=1.5pt, mark options={solid, fill=blue900, draw=blue900}] coordinates {(12,1.0010) (24,1.0007) (48,1.0005) (96,1.0026) (192,1.0043)};
\addlegendentry{\texttt{$\patchsize$=1}}
\addplot[draw=none, forget plot, name path=lop0s1] coordinates {(12,1.0245) (24,1.0240) (48,1.0214) (96,1.0247) (192,1.0230) (384,1.0077)};
\addplot[draw=none, forget plot, name path=hip0s1] coordinates {(12,1.0331) (24,1.0264) (48,1.0229) (96,1.0254) (192,1.0281) (384,1.0258)};
\addplot[fill=blue700, opacity=0.3, forget plot] fill between[of=lop0s1 and hip0s1];
\addplot[color=blue700, solid, mark=*, mark size=1.5pt, mark options={solid, fill=blue700, draw=blue700}] coordinates {(12,1.0288) (24,1.0252) (48,1.0221) (96,1.0250) (192,1.0256) (384,1.0168)};
\addlegendentry{\texttt{$\patchsize$=2}}
\addplot[draw=none, forget plot, name path=lop0s2] coordinates {(12,1.0651) (24,1.0579) (48,1.0502) (96,1.0415) (192,1.0445) (384,1.0309)};
\addplot[draw=none, forget plot, name path=hip0s2] coordinates {(12,1.0772) (24,1.0658) (48,1.0545) (96,1.0470) (192,1.0488) (384,1.0415)};
\addplot[fill=blue300, opacity=0.3, forget plot] fill between[of=lop0s2 and hip0s2];
\addplot[color=blue300, solid, mark=*, mark size=1.5pt, mark options={solid, fill=blue300, draw=blue300}] coordinates {(12,1.0712) (24,1.0619) (48,1.0523) (96,1.0443) (192,1.0466) (384,1.0362)};
\addlegendentry{\texttt{$\patchsize$=4}}
\addplot[draw=none, forget plot, name path=lop0s3] coordinates {(12,1.1630) (24,1.1400) (48,1.1065) (96,1.0942) (192,1.0866) (384,1.0576)};
\addplot[draw=none, forget plot, name path=hip0s3] coordinates {(12,1.1787) (24,1.1518) (48,1.1351) (96,1.1021) (192,1.1039) (384,1.1006)};
\addplot[fill=blue100, opacity=0.3, forget plot] fill between[of=lop0s3 and hip0s3];
\addplot[color=blue100, solid, mark=*, mark size=1.5pt, mark options={solid, fill=blue100, draw=blue100}] coordinates {(12,1.1708) (24,1.1459) (48,1.1208) (96,1.0982) (192,1.0953) (384,1.0791)};
\addlegendentry{\texttt{$\patchsize$=8}}
\end{axis}
\end{tikzpicture}
        \vspace{-\baselineskip}
        \caption{Ratio $\frac {B_\patchsize}{A_\patchsize}$ of how much larger the differences of residuals (computed on the \texttt{France} dataset's ALS labels) are across borders of a grid cell of size $\patchsize$ ($B_\patchsize$) compared to within a grid cell ($A_\patchsize$). Deviating from this, we compute $\frac {B_2}{A_2}$ for the model with $\patchsize = 1$.}
        \label{fig:boundary_ratio}
    \end{minipage}\hfill
    \begin{minipage}[t]{0.305\textwidth}
        \centering
        \resizebox{!}{95px}{%
            \begin{tikzpicture}[rounded corners=0pt]

\begin{axis}[
    name=corr,
    width=6cm, height=6cm,
    xmin=0, xmax=50.0,
    ymin=0, ymax=50.0,
    xtick={0,10,20,30,40,50}, ytick={0,10,20,30,40,50},
    enlargelimits=false, tick align=inside, axis on top,
    axis line style={bluegrey900}, tick style={bluegrey900},
    rounded corners=0pt,
    xlabel={ALS Label (m)}, ylabel={GEDI Label (m)},
    ylabel style={yshift=-1mm},
    colormap name=ScatterMagma,
    colorbar,
    colorbar style={
        xshift=-0.5mm,
        width=4mm,
        axis line style={bluegrey900}, tick style={bluegrey900},
        rounded corners=0pt,
        ytick={0,1,2,3,4},
        yticklabels={$10^{0}$,$10^{1}$,$10^{2}$,$10^{3}$,$10^{4}$},
        yticklabel style={font=\scriptsize, anchor=west},
        tick align=inside,
        ylabel={Number of pixels},
        ylabel shift=-1mm,
    },
    point meta min=0, point meta max=4.8315,
]
\addplot[matrix plot*, point meta=explicit, opacity=0,
    mesh/rows=2, mesh/cols=2]
    table[x=x, y=y, meta=z, row sep=crcr] {%
        x y z \\
        0 0 0.000000 \\ 50.0 0 4.831511 \\
        0 50.0 0.000000 \\ 50.0 50.0 4.831511 \\
    };
\addplot graphics[xmin=0, xmax=50.0, ymin=0, ymax=50.0]
    {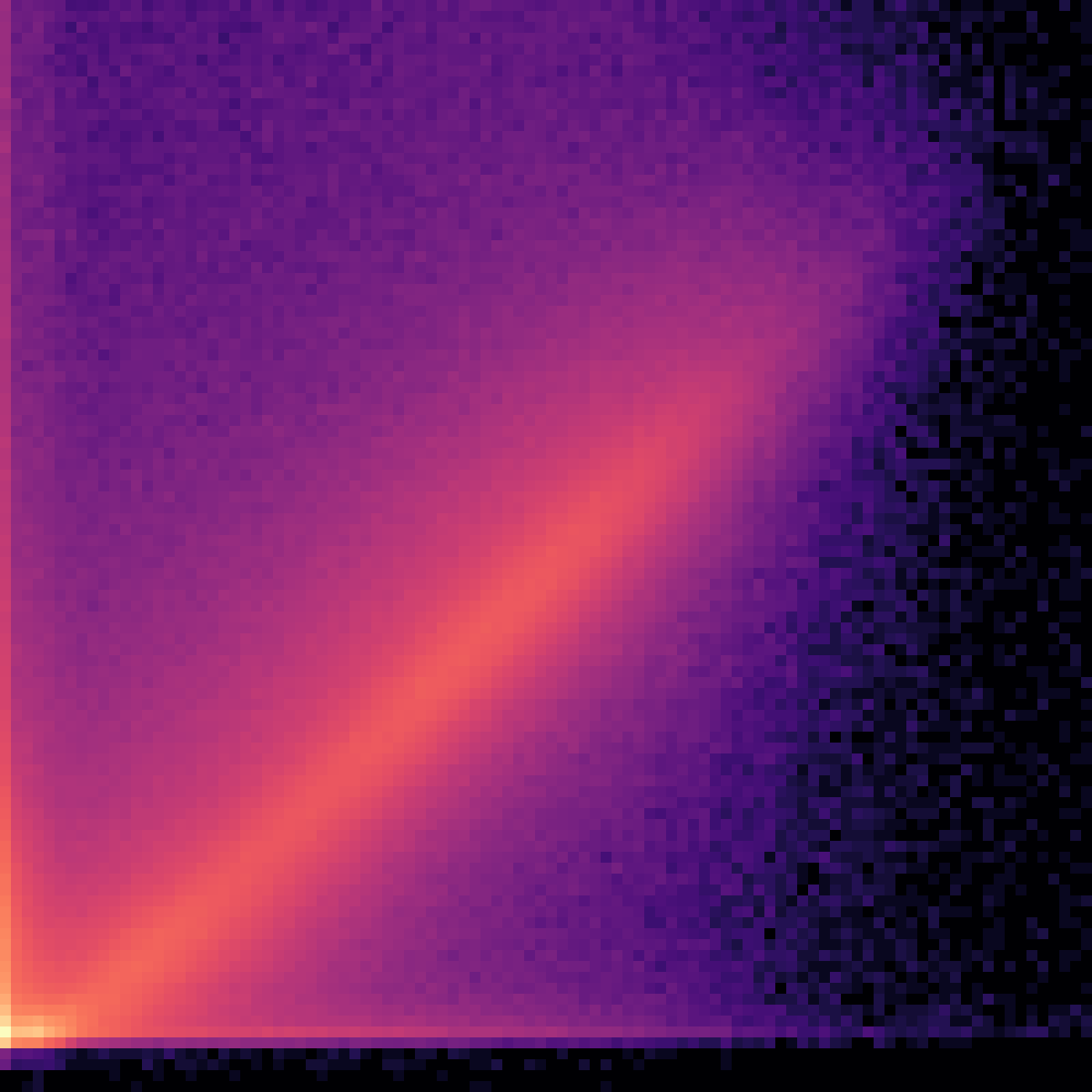};
\addplot[black, dashed, thick, domain=0:50.0] {x};
    \node[anchor=north west, fill=white, fill opacity=0.8,
         text opacity=1, rounded corners=0pt, inner sep=3pt,
         font=\small, align=left]
        at (axis cs:1.00,49.00)
        {$R^2 = 0.436$};
\end{axis}

\end{tikzpicture}
        }
        \vspace{-\baselineskip}
        \caption{Correlation of ALS and GEDI labels on the \texttt{France} dataset (computed where both labels are available).}
        \Description{The scatter heatmap clearly shows how noisy GEDI labels are: Taking the ALS labels as ground truth, GEDI labels are a significantly worse prediction than our model (which was trained on GEDI labels).}
        \label{fig:als_gedi_label_correlation}
    \end{minipage}
\end{figure*}

\begin{table*}
    \centering
    \caption{Height $\height$ (equal to the width $\width$) of the feature maps for various patch sizes and stages, given by $\height / (\patchsize \cdot 2^{i-1}) \times \width / (\patchsize \cdot 2^{i-1})$, for an input of $\height = \width = 256$. The channels double after every stage.}
    \begin{tabular}{lcccc}
    \toprule
    & \multicolumn{4}{c}{Patch size $\patchsize$} \\
    \cmidrule(lr){2-5}
    Stage $i$ & $1$ & $2$ & $4$ & $8$ \\
    \midrule
    1  & $256$ & $128$ & $64$ & $32$ \\
    2 & $128$ & $\phantom{0}64$ & $32$ & $16$ \\
    3 & $\phantom{0}64$ & $\phantom{0}32$ & $16$ & $\phantom{0}8$ \\
    4 & $\phantom{0}32$ & $\phantom{0}16$ & $\phantom{0}8$ & $\phantom{0}4$ \\
    \bottomrule
    \end{tabular}
    \label{tab:stage_resolutions}
\end{table*}

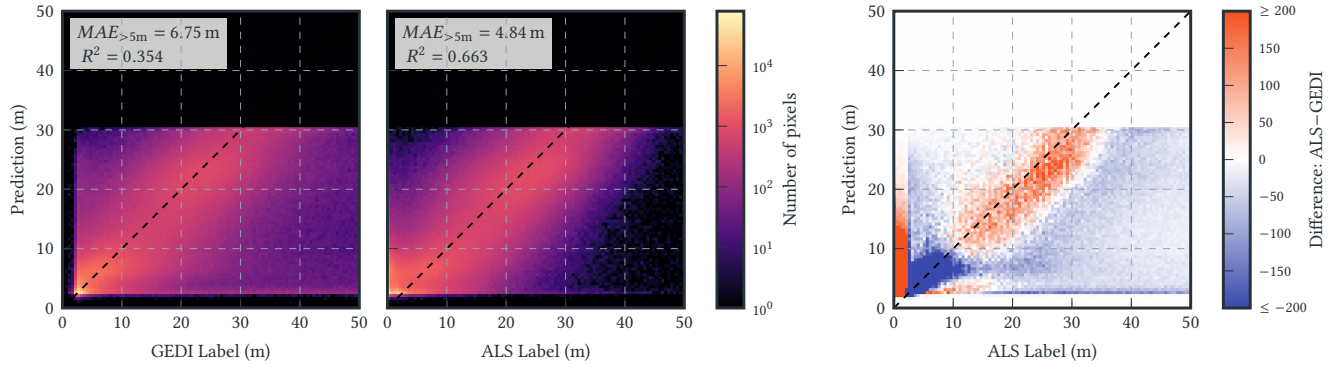
\begin{figure*}[tp]
    \centering
    \resizebox{0.99\linewidth}{!}{\begin{tikzpicture}[rounded corners=0pt]

\begin{axis}[
    name=gedi,
    width=6cm, height=6cm,
    xmin=0, xmax=50.0,
    ymin=0, ymax=50.0,
    xtick={0,10,20,30,40,50}, ytick={0,10,20,30,40,50},
    enlargelimits=false, tick align=inside, axis on top,
    axis line style={bluegrey900}, tick style={bluegrey900},
    rounded corners=0pt,
    xlabel={GEDI Label (m)}, ylabel={Prediction (m)},
    ylabel style={yshift=-1mm},
    colormap name=ScatterMagma,
    point meta min=0, point meta max=4.8990,
]
\addplot[matrix plot*, point meta=explicit, opacity=0,
    mesh/rows=2, mesh/cols=2]
    table[x=x, y=y, meta=z, row sep=crcr] {%
        x y z \\
        0 0 0.000000 \\ 50.0 0 4.899005 \\
        0 50.0 0.000000 \\ 50.0 50.0 4.899005 \\
    };
\addplot graphics[xmin=0, xmax=50.0, ymin=0, ymax=50.0]
    {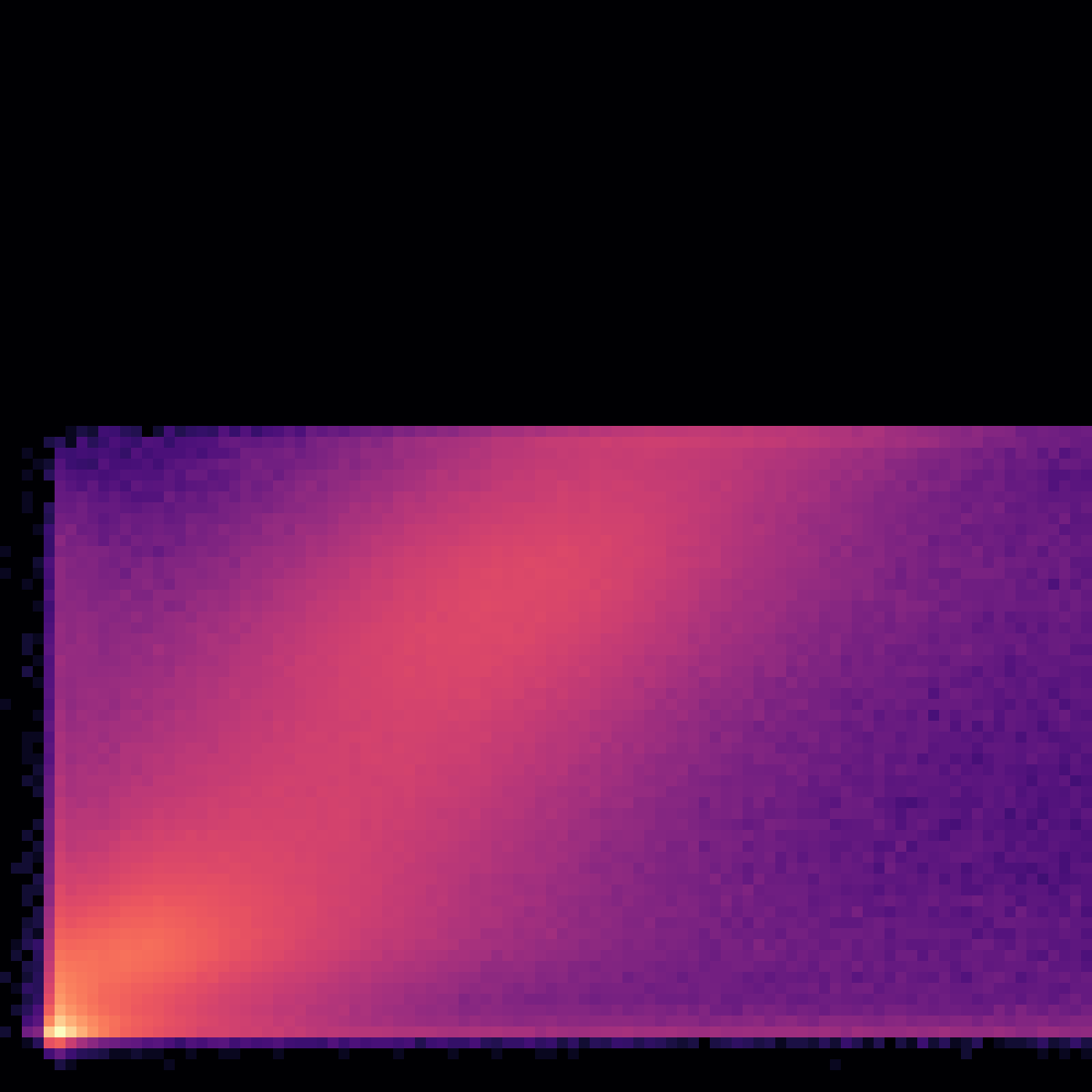};
\addplot[black, dashed, thick, domain=0:50.0] {x};
    \node[anchor=north west, fill=white, fill opacity=0.8,
         text opacity=1, rounded corners=0pt, inner sep=3pt,
         font=\small, align=left]
        at (axis cs:1.00,49.00)
        {$MAE_{>5\text{m}} = 6.75$\,m \\\ $R^2 = 0.354$};
\end{axis}

\begin{axis}[
    name=als,
    at={(gedi.outer south east)}, anchor=outer south west,
    xshift=-0.5mm,
    width=6cm, height=6cm,
    xmin=0, xmax=50.0,
    ymin=0, ymax=50.0,
    xtick={0,10,20,30,40,50}, ytick={0,10,20,30,40,50},
    enlargelimits=false, tick align=inside, axis on top,
    axis line style={bluegrey900}, tick style={bluegrey900},
    rounded corners=0pt,
    xlabel={ALS Label (m)}, yticklabels={},
    colormap name=ScatterMagma,
    colorbar,
    colorbar style={
        xshift=-0.5mm,
        width=4mm,
        axis line style={bluegrey900}, tick style={bluegrey900},
        rounded corners=0pt,
        ytick={0,1,2,3,4},
        yticklabels={$10^{0}$,$10^{1}$,$10^{2}$,$10^{3}$,$10^{4}$},
        yticklabel style={font=\scriptsize, anchor=west},
        tick align=inside,
        ylabel={Number of pixels},
        ylabel shift=-1mm,
    },
    point meta min=0, point meta max=4.8990,
]
\addplot[matrix plot*, point meta=explicit, opacity=0,
    mesh/rows=2, mesh/cols=2]
    table[x=x, y=y, meta=z, row sep=crcr] {%
        x y z \\
        0 0 0.000000 \\ 50.0 0 4.899005 \\
        0 50.0 0.000000 \\ 50.0 50.0 4.899005 \\
    };
\addplot graphics[xmin=0, xmax=50.0, ymin=0, ymax=50.0]
    {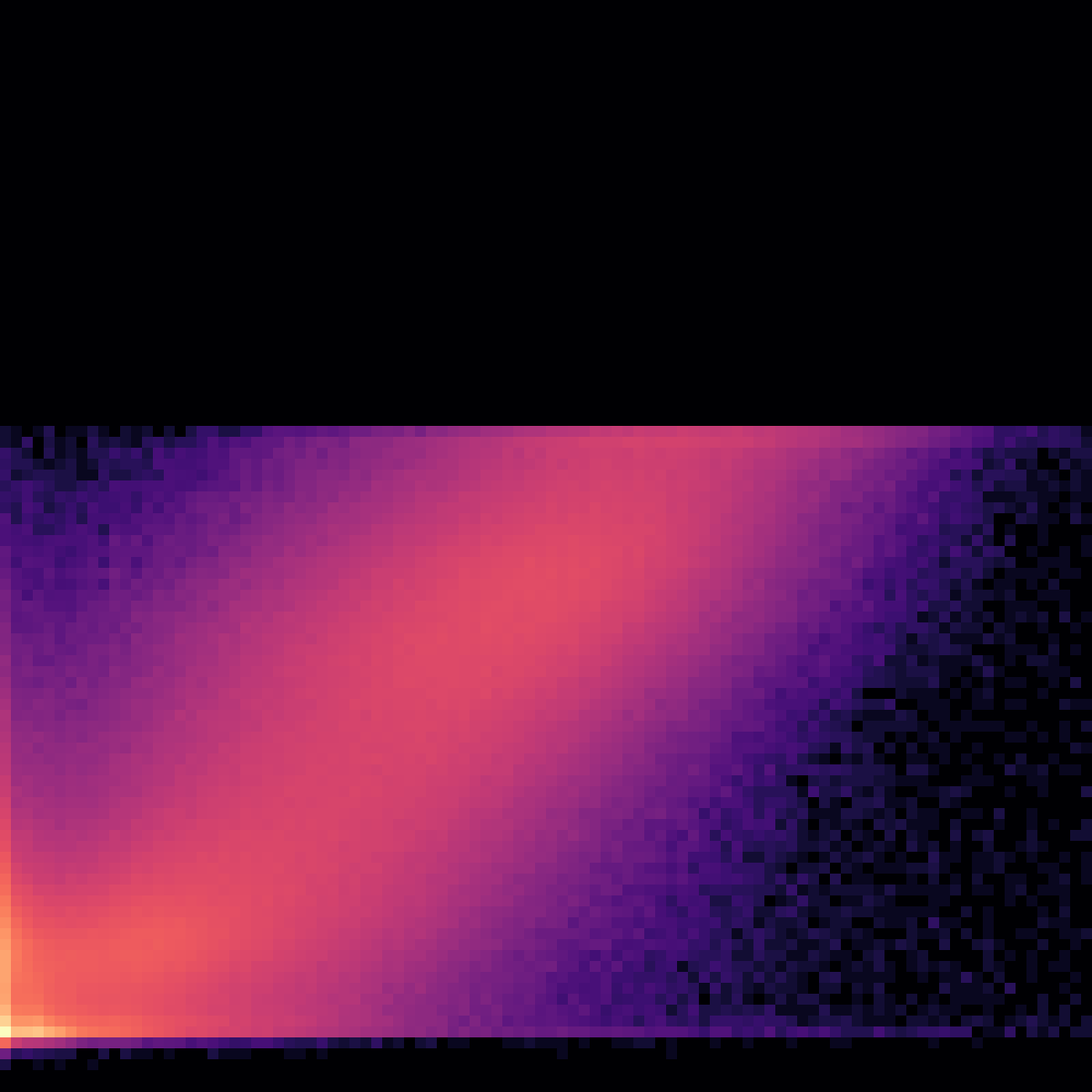};
\addplot[black, dashed, thick, domain=0:50.0] {x};
    \node[anchor=north west, fill=white, fill opacity=0.8,
         text opacity=1, rounded corners=0pt, inner sep=3pt,
         font=\small, align=left]
        at (axis cs:1.00,49.00)
        {$MAE_{>5\text{m}} = 4.84$\,m \\\ $R^2 = 0.663$};
\end{axis}

\begin{axis}[
    name=diff,
    at={(gedi.outer south east -| als.outer east)}, anchor=outer south west,
    xshift=20mm,
    width=6cm, height=6cm,
    xmin=0, xmax=50.0,
    ymin=0, ymax=50.0,
    xtick={0,10,20,30,40,50}, ytick={0,10,20,30,40,50},
    enlargelimits=false, tick align=inside, axis on top,
    axis line style={bluegrey900}, tick style={bluegrey900},
    rounded corners=0pt,
    xlabel={ALS Label (m)}, ylabel={Prediction (m)},
    ylabel style={yshift=-1mm},
    colormap name=ScatterDiverging,
    colorbar,
    colorbar style={
        xshift=-0.5mm,
        width=4mm,
        axis line style={bluegrey900}, tick style={bluegrey900},
        rounded corners=0pt,
        ytick={-200,-150,-100,-50,0,50,100,150,200},
        yticklabels={$\leq-200$,$-150$,$-100$,$-50$,$0$,$50$,$100$,$150$,$\geq200$},
        yticklabel style={font=\scriptsize, anchor=west},
        tick align=inside,
        ylabel={Difference: ALS$-$GEDI},
        ylabel shift=-2mm,
    },
    point meta min=-200.00000000, point meta max=200.00000000,
]
\addplot[matrix plot*, point meta=explicit, opacity=0,
    mesh/rows=2, mesh/cols=2]
    table[x=x, y=y, meta=z, row sep=crcr] {%
        x y z \\
        0 0 -200.000000 \\ 50.0 0 200.000000 \\
        0 50.0 -200.000000 \\ 50.0 50.0 200.000000 \\
    };
\addplot graphics[xmin=0, xmax=50.0, ymin=0, ymax=50.0]
    {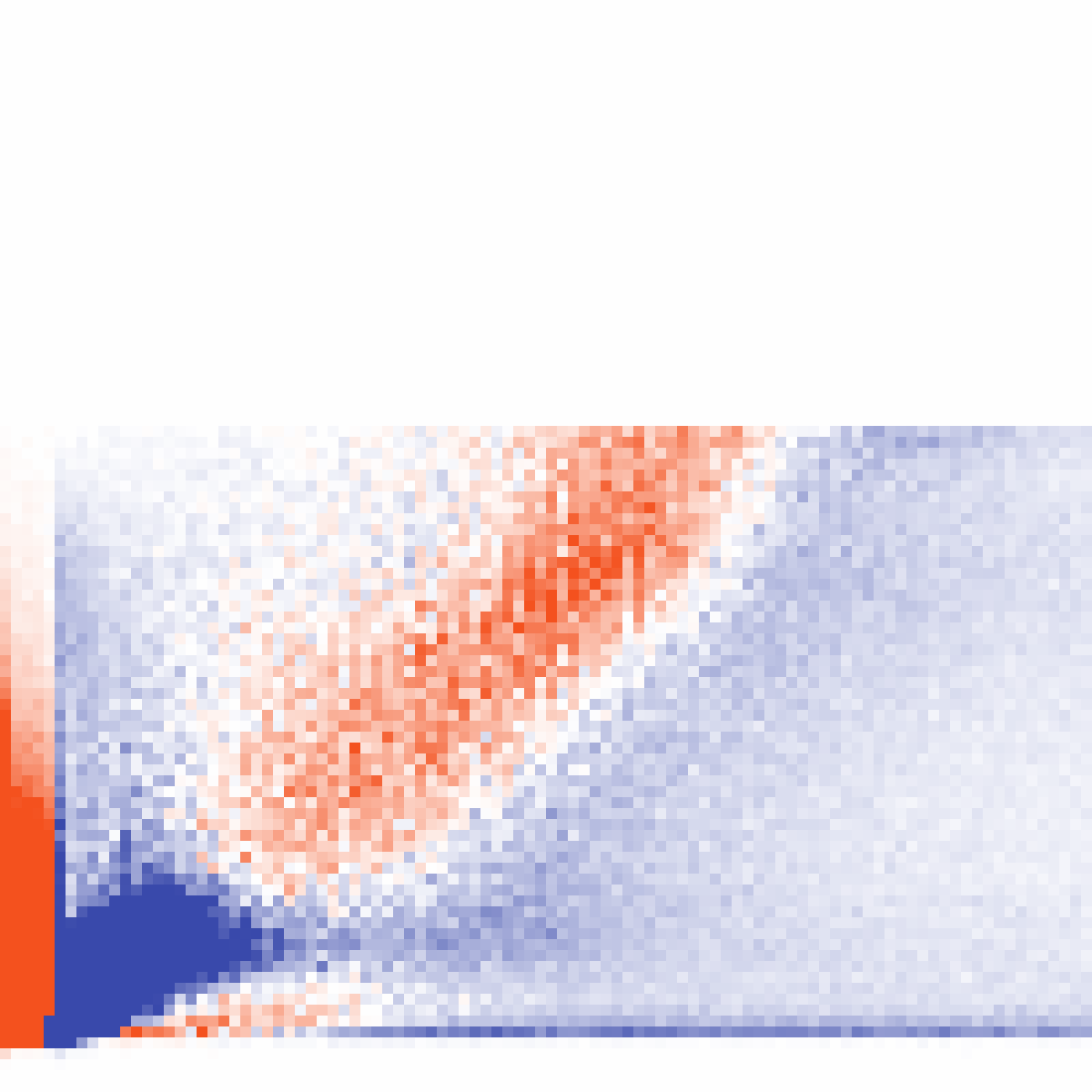};
\addplot[black, dashed, thick, domain=0:50.0] {x};
\end{axis}

\end{tikzpicture}}
    \vspace{-\baselineskip}
    \caption{Predicted vs.\ reference canopy height for the
    $\patchsize=8,\dmodel=192$ model, over pixels where both labels exist. Evidently, the model has not (yet) learned to predict trees taller than \SI{30}{m}.}
    \Description{The model does not predict trees higher than around 32 meters. Comparing the results of this model against the model with patch size of 1 and model dimension of 96, we see a larger deviation from the perfect fit line, and a smaller improvement between GEDI and ALS labels.}
    \label{fig:scatter_heatmap_p8_dm192}
\end{figure*}

\begin{figure*}[tp]
    \centering
    \resizebox{0.99\linewidth}{!}{\input{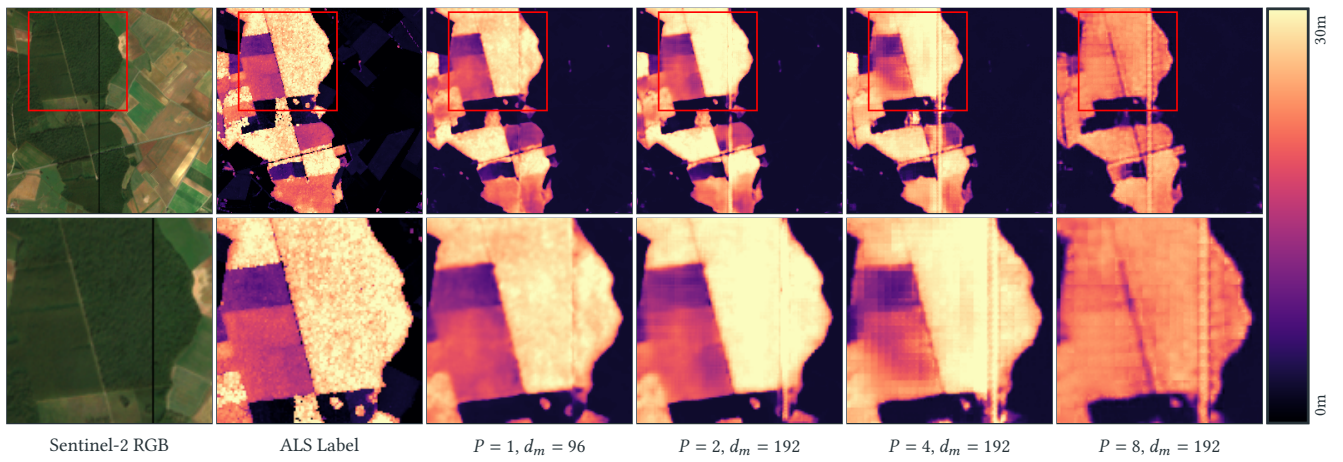}}
    \vspace{-\baselineskip}
    \caption{Qualitative comparison of models with different patch sizes ($\patchsize$) and model dimensions ($\dmodel$). Models with smaller patch sizes handle the vertical line of missing input data better and produce more accurate predictions.}
    \Description{The input has a vertical line of width one where all pixels are black. This problem in data quality disturbs the predictions of models with larger patch sizes more significanly than the ones of models with smaller patch sizes.}
    \label{fig:qualitative_patchsizes_2}
\end{figure*}

\begin{figure*}[tp]
    \centering
    \resizebox{0.99\linewidth}{!}{%
        \input{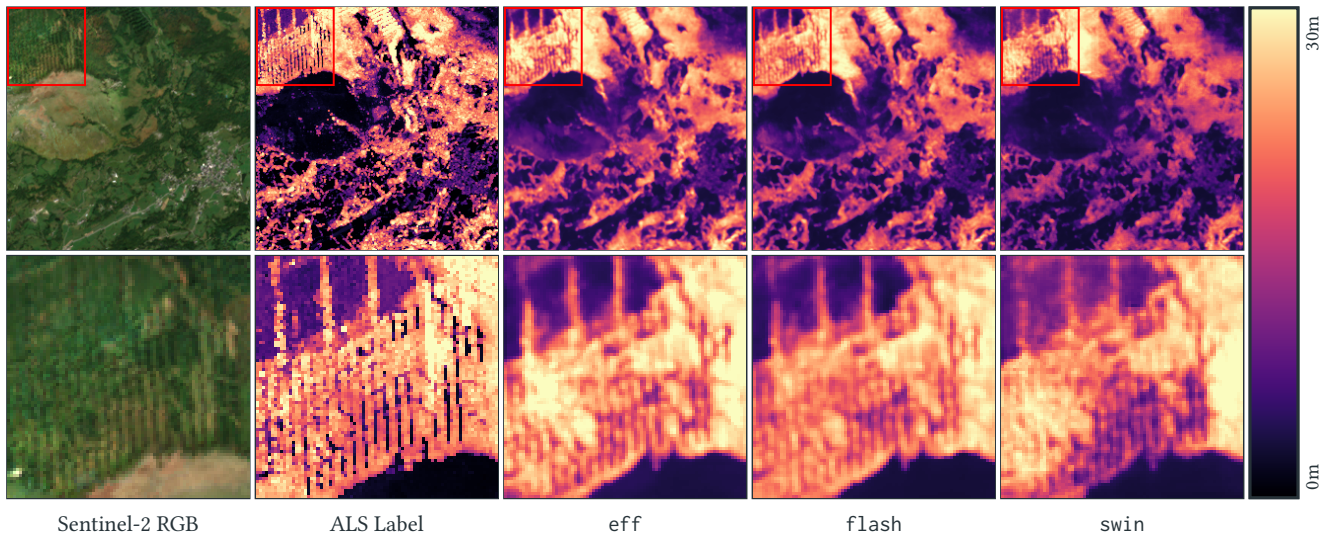}
    }
    \vspace{-\baselineskip}
    \caption{Qualitative comparison of models with different attention mechanisms (same in encoder and decoder).}
    \Description{Using shifted window attention in both encoder and decoder results in predictions of noticeably worse perceptual quality.}
    \label{fig:qualitative_attention}
\end{figure*}

\end{document}